\documentclass[10pt]{article}

\usepackage[top=2.5cm, bottom=2.5cm, left=2.5cm, right=2.5cm]{geometry}
\usepackage{amsmath,amssymb}
\usepackage{graphicx}
\usepackage{authblk}
\usepackage{multicol}
\usepackage[colorlinks=true,linkcolor=blue,citecolor=blue,urlcolor=blue]{hyperref}
\usepackage[numbers]{natbib}

\usepackage{amsfonts}
\usepackage{textcomp}
\usepackage{comment}
\usepackage{array}
\usepackage{subcaption}
\usepackage{pgffor}
\usepackage{makecell}
\usepackage{enumerate}
\usepackage{tabularx}

\usepackage{siunitx}
\usepackage{subcaption}
\usepackage{mathtools}
\usepackage{algorithm}
\usepackage{algorithmic}
\usepackage{booktabs}
\usepackage{todonotes}

\newcommand{\regularizer}{\mathcal{R}}

\DeclarePairedDelimiter{\flexpar}{(}{)}
\DeclarePairedDelimiter{\flexquad}{[}{]}
\DeclarePairedDelimiter{\flexbrace}{\lbrace}{\rbrace}

\newcommand{\xd}{x_d}
\newcommand{\xu}{x_u}
\newcommand{\nit}{n_{\mathrm{it}}}
\newcommand{\vectorization}{\mathrm{vec}}

\newcommand{\norm}[1]{\left\lVert #1 \right\rVert}

\newcommand{\scalar}[2]{\left\langle #1\, ,#2 \right\rangle}

\usepackage{pifont}

\usepackage{comment}

\newcommand{\our}{MPISuperRes-PnP}
\newcommand{\ddp}{\textit{deep denoiser prior}}

\title{MPISuperRes-PnP: A Super-Resolution Zero-Shot Plug-and-Play Reconstruction Algorithm for Magnetic Particle Imaging}

\author[1]{Vladyslav Gapyak \thanks{Emails: vladyslav.gapyak@dlr.de, thomas.maerz@h-da.de, andreas.weinmann@thws.de}}
\author[2,3]{Thomas März}
\author[4]{Andreas Weinmann}
\affil[1]{Institute for the Protection of Terrestrial Infrastructures, German Aerospace Center (DLR), Sankt Augustin, Germany}
\affil[2]{Hochschule Darmstadt, Sch\"{o}fferstr. 3, 64295, Darmastdt, Germany}
\affil[3]{Data Science Institute, European University of Technology, European Union}
\affil[4]{Algorithms for Computer Vision, Imaging and Data Analysis Lab, Technische Hochschule Würzburg-Schweinfurt, Ignaz-Schön-Straße 11, 97421 Schweinfurt, Germany}

\date{\today}

\begin{document}
	\maketitle
	
\begin{abstract}
	\emph{Objective.}
	Magnetic Particle Imaging (MPI) is a promising, emerging medical imaging modality.
	MPI is based on the non-linear response of magnetic nanoparticles to an applied magnetic field 
	and does not expose the specimen to ionizing radiation. 
	The measured signal is the voltage induced in receive coils by the particles' response. Reconstructing the particle concentration from the signal constitutes the imaging task.
	Even using state-of-the-art measurement-based reconstruction, the associated spatial grid is very coarse,
	hence super-resolution techniques are important.
	In this work, we propose an approach for super-resolution in MPI inspired by energy minimization. 
	
	\emph{Approach.}
	Different methods have been proposed for super-resolution in MPI, ranging from 
	upscaling of the associated system matrix to interpolation of the reconstruction.
	Here we incorporate super-resolution into the reconstruction task via an energy minimization formulation.
	Following the plug-and-play approach to energy minimization
	we derive a splitting scheme where the arising Gaussian denoising task is treated with a pre-trained learned Gaussian denoiser.
	
	\emph{Main results.}
	We derive a super-resolution method for MPI based on a plug-and-play approach
	using a pre-trained denoiser in zero-shot fashion.
	This way, we incorporate benefits of deep learning without training and avoid the need of training data. 
	Further, we provide a quantitative and qualitative evaluation of the proposed method.
	Hyper-parameter are selected via an extended parameter search.
	The found parameters are applied for reconstruction on real data. 
	We show the applicability of our method on synthetic and on real data
	(MPIData:~EquilibriumModelWithAnisotropy and 2D-OpenMPI Data).   
	
	\emph{Significance.}
	The proposed method employs a deep-learning denoiser without training 
	-- thus it does not require presently scarcely available MPI training data.
	The denoiser behaves conservatively, i.e., no hallucination artifacts were observed.
	The super-resolution approach is generic such that it can be applied in future MPI contexts
	involving different regularizers or different imaging tasks.
\end{abstract}
	\vspace{1em}
	

\section{Introduction}

Magnetic Particle Imaging (MPI) is a medical imaging modality introduced 
in 2005 by Gleich and Weizenecker~\cite{gleich2005original}. It is a tracer-based 
modality, which uses superparamagnetic nanoparticles as a contrast agent. 
MPI aims at reconstructing the distribution of particles injected 
in the target specimen by exploiting the particles' non-linear response 
to dynamic magnetic fields. As of today, multiple medical applications for 
MPI have been proposed. Among such applications we mention detection and 
cancer imaging~\cite{Yu2017,Song2018,Du2019,Tay2021}, tracing of 
stem cells~\cite{connell2015advancedcellTherapies,JUNG2018139,Lemaster2018}, 
blood flow and cardiovascular imaging~\cite{Franke2020BloodFlow,Bakenecker2018MPIvascular,Tong2021,Vaalma2017},
safety measurements of medical implants~\cite{Wegner2025cadaver}. 
MPI offers a series of benefits when compared with imaging modalities 
such as CT~\cite{Buzug2008CTbook}, MRI~\cite{schamelsafety2015}, PET~\cite{PET} and SPECT~\cite{SPECT}; 
for example, it does not employ ionizing radiation nor radioactive tracers, 
and offers shorter acquisition time as well as high spatial resolution~\cite{BuzugKnopp2012}. 
For a thorough comparison between MPI and other imaging modalities we refer 
the interested reader to~\cite{billingsMPIapplications,Yang2022}.

An MPI scan is usually performed by first applying a static selection field 
that magnetically saturates the particles everywhere in the specimen
with the exception of a field-free region (FFR).
Then, an additional dynamic drive field is superimposed to the static field 
to steer the FFR within the field of view (FoV) in the scanner. 
The magnetic nanoparticles exposed to the dynamic magnetic field 
induce a voltage in receive coils. The induced voltage 
constitutes the scan signal from which the particle concentration is to 
be reconstructed. 

There are currently two main classes of approaches for the reconstruction of
particle distributions: measurement-based approaches and model-based approaches. 
Beyond addressing theoretical understanding, model-based reconstruction techniques 
aim at significantly reducing calibration procedures;
for details, cf. for instance~\cite{Rahemeretal2009,GoodwillConolly2011,marz2016model, bringout2020new,maass2024equilibriumanysotropy, gapyak2023ffl3d}. 
In this work, we focus on the measurement-based approach.
Measurement-based approaches typically acquire a system matrix obtained 
through the following calibration procedure: 
for each column of the matrix, one performs scanning of a delta probe located 
in a corresponding pixel or voxel of a chosen 2D pixel or 3D voxel 
grid~\cite{knopp2011prediction,WeizeneckerBorgertGleich2007,Rahmer_etal2012,Lampe_etal2012}. 
More precisely, a probe with a reference concentration of tracer is iteratively 
positioned and scanned at each cell (pixel / voxel) of the grid. 
Actually, several scans are performed and averaged for each position of a delta probe.
This way the response of the scanning system to discrete delta impulses at each 
voxel position is collected and stored in the system matrix $A$. 
Assuming linearity, the acquired signal $f$ of the scan of a specimen is then 
obtained as a superpositon of the scans of the delta impulses.
For reconstruction, one has to solve a corresponding system of equations 
$A u = f$, where the symbol $u$ denotes the desired particle distribution; 
loosely speaking, one has to invert the system matrix for the measured signal $f$. Due to the presence of noise and the ill-conditioning of the system matrix (e.g., \cite{Knopp_etal2010ec,storath2016edge}), regularization techniques are needed for the inversion.
We observe that -- in a typical measurement-based reconstruction setup -- 
the grid associated with the system matrix determines a priori the later spatial 
resolution of the reconstructions. The simple observation that doubling the 
resolution of the reconstruction would require performing new scans at 4 times 
the considered delta probes (2D), or 8 times the considered delta probes (3D), 
and that the calibration can take hours 
(e.g.,\cite{Schrank2022superresolution, gungor2022tranSMS}), 
explains why super-resolution techniques are desirable in measure-based MPI 
reconstruction. 

\paragraph{Related Work.}
For super-resolution in measurement-based MPI reconstruction, 
most approaches proposed in the literature derive a higher resolved system 
matrix from a lower resolution measured version. Then they solve 
(a potentially regularized variant) of the corresponding linear system of 
equations (as in standard reconstruction). 
The solution yields a higher spatially resolved reconstruction. 
Classical methods, e.g., \cite{kluth2020joint},  as well as deep learning 
based methods have been proposed 
\cite{baltruschat20203d, Schrank2022superresolution}. 
A transformer-based approach is the TranSMS~\cite{gungor2022tranSMS}.
There is also research which employs the measured system matrix and works 
on a super-resolved image domain. We refer in particular to the papers 
\cite{omer2015simultaneous,timmermeyer2013super} which employ classical 
(non-learning based) iterative schemes    
An overview on super-resolution in MPI can be also found in the recent 
survey article \cite{zhang2024current}.

A particular class of deep learning-based reconstruction approaches 
are plug-and-play (PnP) approaches~\cite{venkatakrishnan2013pnp}.
PnP approaches have been used for a variety of imaging modalitites, 
e.g.,~\cite{Ahmad2020pnpMRI}, as well as for computer vision tasks, e.g., \cite{Zhang2022pnp}. 
For MPI reconstruction, PnP approaches have been proposed in~\cite{askin2022pnp} 
which is based on an ADMM scheme,
as well as in \cite{gapyak2025ell1pnp,gapyak2025trajectoryindep} which uses a half-quadratic splitting 
and allows for a model-guided hyper-parameter choice strategy 
as well as an additional $\ell^1$-prior.
PnP approaches typically result in iterative schemes which alternate 
between solving a classically regularized inversion problems such as 
classical Tikhonov reconstruction and applying Gaussian denoising.
Gaussian denoising in turn is a standard computer vision task
and many good deep learning denoising models are available
which can be ``plugged in'' to complete the iterative scheme. 

\paragraph{Contributions.} 
In this paper, we derive an energy-minimization inspired PnP-type method 
for super-resolution in MPI, and show its potential by application to real 
and synthetic data. More precisely, 
\begin{enumerate}[i)]
	\item We introduce \our, an algorithm for the super-resolved 
	reconstruction in MPI leveraging a plug-and-play approach 
	and the denoising capabilities of a zero-shot denoiser, 
	the \ddp~\cite{Zhang2022pnp}, akin to~\cite{gapyak2025ell1pnp}. 
	The incorporation of super-resolution in plug-and-play approaches 
	is derived as a splitting scheme that minimizes a specifically defined 
	cost function, providing mathematical motivation for the algorithm.
	\item We apply the derived method to various public MPI datasets. 
	In particular, we show reconstruction results on the MPI-MNIST 
	dataset~\cite{iske2025mpimnist} published 
	with \cite{iske2025learneddiscrepancy}, on 
	the ``MPIData: EquilibriumModelWithAnisotropy''~\cite{knopp2024equilibriumdata} 
	published with~\cite{maass2024equilibriumanysotropy}, 
	and on the 2D OpenMPIData~\cite{knopp2020openmpidata}.
\end{enumerate}
An implementation of  \our\ can be found on GitHub at: \href{https://github.com/}{https://github.com/} (the full link will be provided upon publication.)

\paragraph{Outline.} 
Section~\ref{sec:method} starts with a description of the proposed method.
More precisely, in Section~\ref{subsec:sm:reco:problem}, we review system-matrix 
based reconstruction in MPI, in section~\ref{subsec:data:preproc} 
we deal with data preprocessing, in section~\ref{subsec:super:pnp} 
we derive the proposed algorithm: we start out considering a suitable 
energy functional, discuss the choice of parameters, and provide a pseudocode.  
In section~\ref{sec:experiments} we present the conducted experiments 
together with the results obtained. More precisely, in Section~\ref{subsec:datasets} we describe the datasets used and in Section~\ref{subsec:measures} the image quality measures employed; then, in Sections~\ref{subsec:exp:mpimnist}, \ref{subsec:exp:mpichar}, \ref{subsec:exp:openmpi} and \ref{subsec:exp:emwa} we present the reconstructions on the four datasets described in Section \ref{subsec:datasets}.
Our conclusions follow in Section~\ref{sec:conclusion}.

\section{Methods}\label{sec:method}

In this section we introduce the MPI reconstruction problem based on a system matrix (Section~\ref{subsec:sm:reco:problem}).
We recall the main data preprocessing steps typically employed (Section~\ref{subsec:data:preproc}) and, finally, introduce the \our\ algorithm in Section~\ref{subsec:super:pnp}. 

\subsection{System-Matrix-Based Reconstruction in MPI}\label{subsec:sm:reco:problem}

In MPI the target concentration of nanoparticles $\rho_{\mathrm{GT}}$ is to be reconstructed from the measured signal $f$.
In the system-matrix-based approach, the problem is considered in a discrete (spatial) scenario.  
Consequently, in a two-dimensional setup, the field-of-view (FoV) $\Omega$ is discretized by considering 
an $N_x \times N_y$ grid. We denote with $N = N_x\cdot N_y$ the total number of grid cells. 
With this discretization, the target concentration $\rho_{\mathrm{GT}}$ is approximated by a 
vector $x\in\mathbb{R}^{N}$ such that $\rho_{\mathrm{GT}}\approx \tilde{x}\in\mathbb{R}^{N_x\times N_y}$ 
where $\tilde{\cdot}$ is the reshaping operator; additionally, the relation with the signal $f$ 
is modeled via the relation
\begin{equation}\label{eq:Ax:f}
	f = Ax + \eta
\end{equation}
where the  symbol $A\in\mathbb{R}^{M\times N}$ denotes the system matrix and $\eta$ represents the inherent noise in the measurement~\cite{Rahemeretal2009}. In the measurement-based approach, the system matrix $A$ is acquired via a calibration procedure performed by iteratively scanning and repositioning a known concentration of nanoparticles (the $\delta$-concentration) across each cell of the $N_x\times N_y$ grid. We point out that $A$ is the of size $M\times N$, where $N = N_x\times N_y$ as defined above, and $M$ is the number of rows, which depends on the number of channels employed and on the preprocessing applied to the data before solving the linear system in~\eqref{eq:Ax:f} (cf.~section~\ref{subsec:data:preproc}). As an example, in the 2D OpenMPI~\cite{knopp2020openmpidata} the calibration is performed on a $19\times 19$ grid, yielding $N=19^2=361$, and each scan is acquired at $1632$ time points along two channels (with coils positioned along the $x$- and $y$-axis) and Fourier transformed, giving $817$ complex-valued frequencies per channel; upon stacking of the real and imaginary parts per channel to obtain a real-valued matrix, the total amount of rows is $M = 2\cdot 2\cdot 817 = 3268$. Finally, given a scan $f$ and a calibration matrix $A$, the objective is to retrieve $x$ solving~\eqref{eq:Ax:f}. Because both $f$ and $A$ are affected by real noise, the inversion of~\eqref{eq:Ax:f} requires regularization techniques~\cite{bertero2021introduction,kirsch2011introduction}.

\subsection{Data Preprocessing}\label{subsec:data:preproc}

Usual preprocessing steps performed on MPI data affect the number of rows $M$ of the linear 
system in~\eqref{eq:Ax:f}. The reason is that these preprocessing steps are usually aimed at 
either improving the conditioning of the system in~\eqref{eq:Ax:f} or at discarding 
unreliable frequencies, that are deemed as such according to chosen criteria. 
We recall the main preprocessing steps as, for instance, outlined in~\cite{kluth2019enhancedrec}.

\begin{enumerate}[i)]
	\item \textit{Background Correction}: according to the models, the measured data can be decomposed 
	as $f = f_{\mathrm{cl.}}+b_f +\eta_f$ and coherently the system matrix as $A = A_{\mathrm{cl.}}+b_A +\eta_A$, 
	where $f_{\mathrm{cl.}}$, $S_{\mathrm{cl.}}$ are the clean signals, $b_f$, $b_A$ are the background signals 
	and $\eta_f$, $\eta_A$ are the arrays representing the noise. To exclude the background signals, 
	which bear no information about the underlying particle concentrations, empty scans without any 
	particles in the scanner are performed at the time of calibration as well as before or after the 
	scan of the target specimen. These background scans can be subtracted form $A$ and $f$ to correct 
	for the background terms.
	\item \textit{Bandpass Filtering}: some scanners (like the Bruker scanner) utilize an analog 
	filter to suppress excitation crosstalk -- at the price of decreasing the Signal-to-Noise ratio (SNR) --
	of frequencies below $80\ \si{\kilo\hertz}$~\cite{Rahmer_etal2012}. For this reason, a high-pass 
	filter at $80\ \si{\kilo\hertz}$ is usually necessary.
	\item \textit{SNR-Thresholding}: provided an estimation of the SNR of each frequency is available, 
	SNR-based thresholding is usually performed~\cite{kluth2019enhancedrec} by discarding those frequencies 
	whose SNR is below a certain threshold $\Theta$, e.g., below $\Theta = 1\ \si{\deci\bel}$.
	\item \textit{Whitening}: in~\cite{dittmer2020deep} the authors have proposed to apply a whitening 
	transformation, i.e., considering the whitened system $WAx=Wf$ where the matrix $W\in\mathbb{R}^{M\times M}$ 
	is the diagonal covariance matrix obtained from background scans. More specifically, the diagonal entries 
	of $W$ are the terms $1/\sigma_b$ and $\sigma_b^2$ is the variance of the background scan.
	\item \textit{Low Rank Approximation}: following~\cite{kluth2019enhancedrec}, a target rank 
	$K\leq\min\lbrace M,n\rbrace$ can be selected and a randomized singular value decomposition 
	(rSVD)~\cite{halko2011rSVD} is performed. The output of the rSVD is a triple of matrix $(U_K ,\Sigma_K ,V_K )$ 
	such that $A \approx U_K \Sigma_K V_K^*$. By employing $U_K$, the system with $K$ rows, $U^*_K A x= U_K^* f$, is then considered.
\end{enumerate}

\subsection{\our : an Algorithm for Super-Resolved Regularized Inversion with Zero-Shot Learned-Denoiser}\label{subsec:super:pnp}

In this section we introduce \our , the proposed algorithm to incorporate super-resolution within the ZeroShot-PnP 
framework in~\cite{gapyak2025ell1pnp}. We show that the proposed scheme arises from an energy 
minimization scheme which addresses~\eqref{eq:Ax:f} and incorporates regularization: 
\begin{equation}\label{eq:tikhonov}
	\hat{x} = \arg\min_{x} \flexbrace*{\frac{1}{2}\norm{Ax-f}_2^2 + \lambda\regularizer (x)} ,
\end{equation}
where $A\in\mathbb{R}^{M\times N}$, for $N=N_x\cdot N_y$ and $M$ which depends on the chosen 
preprocessing applied to the system (cf. section~\ref{subsec:data:preproc}). 
In equation~\eqref{eq:tikhonov} $\regularizer$ is a chosen regularizer and $\lambda >0$ 
is the regularization parameter regulating the strength of the regularization.

For the purpose of superresolution we need an upscaling operator $\mathcal{U}_s\colon\mathbb{R}^{N_x \times N_y }\to \mathbb{R}^{sN_x \times sN_y}$.
The upscale factor $s \in\mathbb{N}$ is a parameter of the upscaling operator $\mathcal{U}_s$
which we define and implement via bilinear interpolation. 
Because we treat particle concentrations $x$ both as 1-dimensional vectors in the solution of~\eqref{eq:scheme:tik} 
as well as 2-dimensional images, we define the reshaping operator $\tilde{\cdot}\colon \mathbb{R}^{N_x\cdot N_y}\to\mathbb{R}^{N_x\times N_y}$ 
(turning 1d vectors into 2d images)
and its inverse, the vectorization operator $\vectorization\colon \mathbb{R}^{N_x\times N_y}\to\mathbb{R}^{N_x\cdot N_y}$
(turning 2d images into 1d vectors). 
The operator $U_s\colon\mathbb{R}^{N_x \cdot N_y }\to \mathbb{R}^{sN_x \cdot sN_y}$ 
is the equivalent of the upscaling operator $\mathcal{U}_s$, but operates on 1d vectors while $\mathcal{U}_s$ operates on 2d images.
The two operators are related by the condition $\vectorization\left (\mathcal{U}_s \tilde{x}\right ) = U_s x$. 
The adjoint operator of $U_s$ is the operator $U_s^* \colon \mathbb{R}^{sN_x \cdot sN_y}\to \mathbb{R}^{N_x \cdot N_y }$ defined 
by the property $\scalar{U_s x}{y} = \scalar{x}{U_s^* y}$ for all $x\in \mathbb{R}^{N_x\cdot N_y}$ 
and $y\in\mathbb{R}^{sN_x\cdot sN_y}$. 

Having chosen the upscaling operator $U_s$, we propose to consider the following variation of~\eqref{eq:tikhonov}
\begin{align}\label{eq:tik:consensus}
	\min_{\xd ,\xu} \flexbrace*{\frac{1}{2}\norm{A\xd - f}_2^2 + \lambda\regularizer (\xu )} \quad \text{s.t.} \quad \xu - U_s\xd = 0 \, ,
\end{align}
where we have decoupled the data fidelity (residual/least squares) term 
and the regularizer such that they operate on the variable $x_d$ defined 
on the space $\mathbb{R}^{N_x\cdot N_y}$ and on $x_u$ defined on the higher resolved 
space $\mathbb{R}^{sN_x\cdot sN_y}$, respectively. 

We form the Lagrangian for the Half-Quadratic Splitting of~\eqref{eq:tik:consensus} which reads
\begin{equation}
	\mathcal{L}_{\mu} (\xd ,\xu ) = \frac{1}{2} \norm{A\xd - f}_2^2 +\lambda\regularizer (\xu ) + \frac{\mu}{2}\norm{\xu - U_s\xd}_2^2
\end{equation}
with a (new) parameter $\mu > 0$. The minimization of the Lagrangian is performed by iterative alternating minimization w.r.t. $x_d,x_u$ which yields the following iteration w.r.t. $k$:
\begin{align}
	\xd^{k+1} & = \arg\min_{\xd} \flexbrace*{\frac{1}{2}\norm{A\xd - f}_2^2 + \frac{\mu_k}{2}\norm{U_s\xd - \xu^{k}}_2^2} \label{eq:scheme:tik}\\
	\xu^{k+1} & = \arg\min_{\xu} \flexbrace*{\lambda\regularizer (\xu )+\frac{\mu_k}{2}\norm{\xu - U_s\xd^{k+1}}_2^2} . \label{eq:scheme:gauss:classic}
\end{align}
We observe that the problem in~\eqref{eq:scheme:tik} is a Tikhonov-type problem whose Euler-Lagrange equations are 
\begin{equation}
	\flexpar*{A^* A + \mu_k U_s^* U_s}\xd = A^* f + \mu_k U_s^* \xu^k \, .
\end{equation}
Because the operator $A^* A + \mu_k U_s^* U_s$ is symmetric positive definite (SPD), we may use the conjugate gradient method (CG) to solve the corresponding linear system.

Another observation is that the problem in~\eqref{eq:scheme:gauss:classic} can be rewritten as
\begin{equation}
	\xu^{k+1} = \arg\min_{\xu} \flexbrace*{\frac{1}{2(\sqrt{\lambda /\mu_k})^2}\norm{\xu - U_s\xd^{k+1}}_2^2 + \regularizer (\xu )}
\end{equation}
which describes a Gaussian denoising of $U_s\xd^{k+1}$ with noise level $\sigma_{k+1} = \sqrt{\lambda / \mu_k}$. In the spirit of Plug-and-Play algorithms~\cite{venkatakrishnan2013pnp}, we substitute the classical Gaussian denoising step resulting from~\eqref{eq:scheme:gauss:classic} with a machine-learning-based Gaussian denoiser to leverage its denoising capabilities.
Finally, we arrive at the splitting scheme underlying the \our\ algorithm:
\begin{align}
	\xd^{k+1} & = \mathrm{ConjGrad}\flexpar*{A^* A + \mu_k U_s^* U_s\, ; A^* f + \mu_k U_s^* \xu^{k}} \label{eq:scheme:cg}\\
	\xu^{k+1} & = \mathrm{GaussDenoiser}\flexpar*{\mathcal{U}_s \tilde{x}_d^{k+1}\, ; \sqrt{\frac{\lambda}{\mu_k}}} . \label{eq:scheme:gaussden}
\end{align}

The obtained scheme has two important points of 
similarity with the ZeroShot-PnP algorithm proposed 
in~\cite{gapyak2025ell1pnp}: 
(i) the performance of the scheme will depends on the 
choice of the Gussian Denoiser chosen in~\eqref{eq:scheme:gaussden}; 
(ii) the noise level $\sigma_{k+1} = \sqrt{\lambda / \mu_k}$ of the 
iterate $\mathcal{U}_s \tilde{x}_d^{k+1}$ is coupled with the Tikhonov 
parameter $\mu_k$ in~\eqref{eq:scheme:cg}. In view of these similarities, 
we employ the benchmark \ddp~\cite{Zhang2022pnp} as denoiser 
in~\eqref{eq:scheme:gaussden}, because the \ddp\ can take noise level 
maps as additional input. We adapt also the automatic update strategy 
proposed in~\cite{gapyak2025ell1pnp} for $\mu_k$.

We now provide more details to elucidate these statements.
The backbone architecture of the \ddp~\cite{Zhang2022pnp} is the DRUNet, a deep CNN architecture 
which combines a U-Net~\cite{Ronneberger2015unet} with the ResNet~\cite{He2016resnet}. 
The \ddp\ has been trained on a combination of various dataset such as the Waterloo 
Exploration Database~\cite{Ma2017Waterloo}, BSD~\cite{Chen2017bsd}, Flick2K~\cite{Lim2017Flick2k} 
and DIV2K~\cite{Agustsson2017ntire}. The denoiser is publicly available 
at \href{https://github.com/cszn/DPIR}{https://github.com/cszn/DPIR} and 
was trained as follows: the authors of~\cite{Zhang2022pnp} randomly cropped out 16 patches 
of size $128\times 128$, selected a random noise level $\sigma$ chosen from $[0,50]$ and added 
additive Gaussian noise with level $\sigma$. Additionally, a map filled uniformly with the value $\sigma$ 
and of the same size as the image has been provided as noise level map. The range $[0,50]$ has been chosen 
to account for large variations of the noise level. For completeness we also mention that the DRUNet 
consists of 32,638,656 trainable parameters. We remark that the \ddp\ is used in \our\ 
in a zero-shot fashion, that means it is used without further training nor fine-tuning on MPI-specific nor MPI-related data.

As mentioned above, the parameters in~\eqref{eq:scheme:cg} and~\eqref{eq:scheme:gaussden} are coupled and 
consequently, it is reasonable to take this into account when choosing the denoiser. 
We have seen that the \ddp\ has been designed in such a way that noise level maps can be set 
as inputs in the denoising, de facto allowing to perform denoising with a prescribed 
parameter in~\eqref{eq:scheme:gaussden}. In particular, inspired by the strategy 
proposed in~\cite{gapyak2025ell1pnp}, we leverage this feature of the \ddp\ to devise an 
automatic parameter update during the iterations. More specifically, at the beginning 
we set $\mu_0$ as a starting (hyper-) parameter; then, for each iterate $\tilde{x}_d^{k+1}$ 
obtained via~\eqref{eq:scheme:cg} with $\mu_k$, we estimate the noise level $\sigma_{k+1}^2$
of $\mathcal{U}_s \tilde{x}_d^{k+1}$ 
with the following estimation (which offers an upper bound on the real noise level):
\begin{equation}
	\hat{\sigma}_{k+1}^2 \coloneq \mathrm{Var} \flexpar*{\mathcal{U}_s \tilde{x}_d^{k+1}} = \mathbb{E}\flexquad*{\flexpar*{\mathcal{U}_s \tilde{x}_d^{k+1} - \overline{\mathcal{U}_s \tilde{x}_d^{k+1}}}^2}
\end{equation}
where $\overline{\mathcal{U}_s \tilde{x}_d^{k+1}}$ denotes the pixel average of 
$\mathcal{U}_s \tilde{x}_d^{k+1}$. We point out that we will use the hat sign $\hat{q}$ 
to denote an estimate of any given quantity $q$. In the first iteration ($k=0$) we obtain 
the estimated noise level  $\hat{\sigma}_{1}^2$ and consequently, can estimate the parameter 
$\lambda$, which mediates the noise level and the Tikhonov parameter, 
as $\hat{\lambda} =\mu_0 \cdot \hat{\sigma}_{1}^2$. 

Finally, the proposed \our\ method is summarized as pseudocode 
in Algorithm~\ref{alg:pnp}. Hyper-parameters of the method are the starting value $\mu_0$ 
and the number of iterations $n_{\mathrm{it}}$.

\begin{algorithm}
	\caption{Pseudocode of the \our\ algorithm.}\label{alg:pnp}
	\textbf{Input}: data $f$, system matrix $A$, upscale factor $s$, interpolator $U_s$, $n_{\mathrm{it}}$, $\mu_0$.\\
	\textbf{Output}: reconstructed $\tilde{x}_\mathrm{rec}$.\\
	\begin{algorithmic}[1]
		\STATE $\xd^0 ,\xu^0\gets 0 $;
		\STATE $k \gets 0$;
		\WHILE{$k \leq n_{\mathrm{it}}$}
		\STATE $\xd^{k+1}\gets\mathrm{ConjGrad}\flexpar*{A^* A + \mu_k U_s^* U_s\, ; A^* f + \mu_k U_s^* \xu^{k}}$;
		\STATE $\hat{\sigma}_{k+1} \gets $Noise-Estimator$(\xd^{k+1})$;
		\IF {$k=0$}
		\STATE $\hat{\lambda} \gets\mu_0 \cdot\hat{\sigma}_1^2$
		\ENDIF
		\STATE $\tilde{x}_u^{k+1}\gets$Denoiser$\left (\mathcal{U}_s \tilde{x}_d^{k+1}\, ,\hat{\sigma}_{k+1}\right ) $; \hfill\COMMENT{ZeroShot-Denoiser}
		\STATE $\xu^{k+1}\gets \vectorization (\tilde{x}_u^{k+1})$
		\STATE $\mu_{k+1}\gets \hat{\lambda} /\hat{\sigma}_{k+1}^2$;
		\STATE $k\gets k+1$;
		\ENDWHILE
		\RETURN $\tilde{x}_u^{k+1}$
	\end{algorithmic}
\end{algorithm}

\section{Experiments and Results}\label{sec:experiments}

In this section we show the results obtained using the \our\ algorithm on a variety of MPI datasets, both simulated and real.
In particular, in section \ref{subsec:datasets} we describe the datasets employed: 
the MPI-MNIST dataset \cite{iske2025mpimnist,iske2025learneddiscrepancy}, 
our own MPI-CHAR dataset, the OpenMPI dataset \cite{knopp2020openmpidata}, and the ``Equilibrium Model with Anisotropy" dataset \cite{knopp2024equilibriumdata}. In section \ref{subsec:measures} we describe the image quality measures employed to quantitatively evaluate the quality of the reconstructions. The reconstructions performed on the four datasets are described in the respective sections \ref{subsec:exp:mpimnist}, \ref{subsec:exp:mpichar}, \ref{subsec:exp:openmpi}, and \ref{subsec:exp:emwa}. 

The \our\ algorithm and the proprocessing of the data have been 
implemented in Python 3.9, using Numpy and PyTorch. 
The reconstructions were performed on a workstation with 
13th Gen Intel(R) Core(TM) i9-13900KS, 128 GB of RAM, 
an NVIDIA RTX A6000 GPU and Windows 11 Pro.

\subsection{Datasets}\label{subsec:datasets}

In this work we test the \our\ algorithm on a variety of simulated and real dataset, which we here describe:
\begin{enumerate}[1.]
	\item The MPI-MNIST \cite{iske2025mpimnist,iske2025learneddiscrepancy} 
	is a recently published dataset of simulated MPI measurements obtained 
	using state-of-the-art model-based system matrices. In addition, 
	the dataset employs real noise measurements obtained from a real MPI 
	scanner. In particular, the dataset contains simulated system matrices 
	$A_{N_x \times N_y}$ for three different resolutions, namely, 
	for $(N_x ,N_y )$ either $(15,17)$, $(45, 51)$ or $(75, 85)$. 
	The ground truths of this dataset have been generated in the following 
	manner: the $28\times 28$ images of the hand written digits in the 
	MNIST dataset are first downsampled to a $11\times 11$ pixel grid with 
	the nearest neighbor interpolation scheme and padded with zeros to get 
	to the size of $15\times 17$ pixels. The resulting $15\times 17$-sized 
	images are upscaled to the $75\times 85$ grid using nearest neighbor 
	interpolation. Finally, the scan signal $y$ is obtained for each ground 
	truth $x$ as
	\begin{equation}\label{eq:mnist:forward}
		y = A_{75\times 85}x + \eta
	\end{equation}
	where $\eta$ is an instance of real noise collected with a real scanner.
	\item We observe that in the MPI-MNIST dataset the ground truths are 
	natively on a $15 \times 17$ grid and upscaled to the $75\times 85$ 
	grid using nearest neighbors interpolation. As a consequence, the 
	features of the ground truth live on the $15 \times 17$ and no more fine-grained features 
	are to be gained from reconstructing the phantom on a $75\times 85$ grid. 
	It follows that the MPI-MNIST data set is limited for testing 
	super-resolution algorithms since it does not carry fine level details.
	For this reason, we have created the 
	MPI-CHAR dataset. The MPI-CHAR dataset is a simulated dataset created 
	using the data-generation methodology of the MPI-MNIST dataset, but 
	using a set of ground truths with higher native resolution. More 
	specifically, characters from an openly available 
	font\footnote{Liberation Sans Font Regular is licensed under 
		GNU general public license (GPL) and available at 
		\href{https://www.1001fonts.com/liberation-sans-font.html}{https://www.1001fonts.com/liberation-sans-font.html}.} 
	have been used to create ground truths on a $75\times 85$ grid; these ground 
	truths have been used to generate signals following the methodology of the 
	MPI-MNIST dataset, i.e., applying the forward operator $A_{75\times 85}$ and 
	adding real noise as in \eqref{eq:mnist:forward}. The matrix $A_{75\times 85}$ 
	and the noise instances are the one published in the MPI-MNIST dataset.
	\item The OpenMPI dataset \cite{knopp2020openmpidata} is a benchmark dataset, 
	which contains real 1D, 2D and 3D scan data of 3 phantoms obtained 
	with a Bruker scanner. Here, we use the 2D scan data and 2D system matrices.
	The dataset contains system matrices in two different resolutions: 
	a $19\times 19$ lower resolution and a $37\times 37$ higher resolution 
	system matrix. The presence of a higher resolution system matrix allows us 
	to compare higher resolution reconstructions of real phantoms with the ones 
	obtained with the \our\ method.
	\item The Equilibrium Model with Anisotropy dataset (EMWA dataset) 
	of \cite{knopp2024equilibriumdata} contains real 2D scan data of 6 phantoms 
	obtained with a Bruker scanner. In this dataset, the system matrix provided 
	has been calibrated on a $15\times 17$ grid. The dataset does not offer a 
	higher resolution system matrix. Nevertheless, we provide reconstructions 
	on this dataset to further test the \our\ algorithm on real MPI data.
\end{enumerate}

The Open-MPI and the EMWA datasets contain real MPI scans and are used 
in this paper to provide a qualitative evaluation of the method on real data. 
The MPI-MNIST and the MPI-CHAR datasets are simulated (with real noise) 
and contain ground truths. These are used to compute quality metrics 
and provide a quantitative evaluation of the method. In the following 
section we describe the metrics used. 

\subsection{Image Quality Measures}\label{subsec:measures}

To quantitatively asses the quality of the reconstructions whenever the ground truths are available, we use peak signal-to-noise-ratio (PSNR) and the structural similarity index measure (SSIM)~\cite{ssim}. Given two arrays $f$ and $g$ of size $N_x\times N_y$, the PSNR is defined as
\begin{equation}
	\mathrm{PSNR}(f,g) = 10\cdot \log_{10}\flexpar*{\frac{R^2}{\mathrm{MSE}(f,g)}}
\end{equation}
where $R=\max\lbrace f\rbrace$ and MSE is the mean square error
\begin{equation}
	\mathrm{MSE}(f,g) = \frac{1}{N_x  N_y}\sum_{i=1}^{N_x}\sum_{j=1}^{N_y} (f_{ij}-g_{ij})^2 ,
\end{equation}
and $f_{ij}$ and $g_{ij}$ are the $(i,j)$-th pixel of $f$ and $g$, respectively. The  SSIM is defined as
\begin{equation}
	\mathrm{SSIM}(f,g) = [l(f,g)]^{\alpha}\cdot [c(f,g)]^{\beta}\cdot [s(f,g)]^{\gamma} 
\end{equation}
where $l$ is the luminance, $c$ the contrast and $s$ the structure functions defined as
\begin{equation}
	l(f,g) = \frac{2\mu_f \mu_g +C_1}{\mu_f^2 + \mu_g^2 + C_1} , \quad c(f,g) = \frac{2\sigma_f \sigma_g +C_2}{\sigma_f^2 + \sigma_g^2 + C_2} ,\quad s(f,g) = \frac{2\sigma_{fg} + C_3}{\sigma_f^2 \sigma_g^2 + C_3}
\end{equation}
in terms of the mean values $\mu_f$, $\mu_g$ of $f$ and $g$, the respective standard deviations $\sigma_f$ and $\sigma_g$ and their covariance $\sigma_{fg}$; similarly to~\cite{dittmer2020deep}, we have set $\alpha = \beta = \gamma = 1$, and $C_1 = (0.01\cdot R)^2$, $C_2 = (0.03\cdot D)^2$ and $C_3 = 0.5\cdot C_2$ where $D$ is the range of the ground truth.

To account for the variation in range of the reconstructions, we compute PSNR and SSIM upon affine rescaling of the reconstructed images. More specifically, if $x_{\mathrm{GT}}$ is the ground truth and $x$ is the reconstructed image, then we  compute the quantity
\begin{equation}
	\mathrm{PSNR}_{\mathrm{aff}} (x_{\mathrm{GT}},x)= \max_{a,b}\flexbrace*{\mathrm{PSNR}(x_{\mathrm{GT}},ax+b)} 
\end{equation}
which can be computed simply as the $\mathrm{PSNR}$ between $x_{\mathrm{GT}}$ and $ax+b$ for the optimal parameters
\begin{equation}\label{eq:a:b}
	a = \frac{\mathrm{Cov}(x_{\mathrm{GT}}\, ,x)}{\mathrm{Var}(x_{\mathrm{GT}})}, \qquad b = \overline{x} - a\cdot \overline{x}_{\mathrm{GT}}
\end{equation}
and $\overline{x}$ (resp. $\overline{x}_{\mathrm{GT}}$) is the mean 
of $x$ (resp. $x_{\mathrm{GT}}$). Analogously, when computing the SSIM 
we will be in fact computing $\mathrm{SSIM}(x_{\mathrm{GT}}, ax+b)$ 
with $a$ and $b$ taken from~\eqref{eq:a:b}.

\subsection{Reconstructions on the MPI-MNIST Dataset}\label{subsec:exp:mpimnist}

\def\size{14em}
\newcolumntype{C}{>{\centering\arraybackslash}m{\size}}
\newcolumntype{L}{>{\raggedright\arraybackslash}X} 
\begin{table}[!t]
	\centering
	\setlength{\tabcolsep}{0pt}
	\begin{tabularx}{\linewidth}{L*3{C}@{}}
		\toprule
		& PSNR & SSIM & Rel. $L^2$-norm \\
		\midrule
		\makecell[l]{\rotatebox[origin=c]{90}{$\mathrm{Super}_2 (A_{45\times51} )$}} & \includegraphics[width=\size]{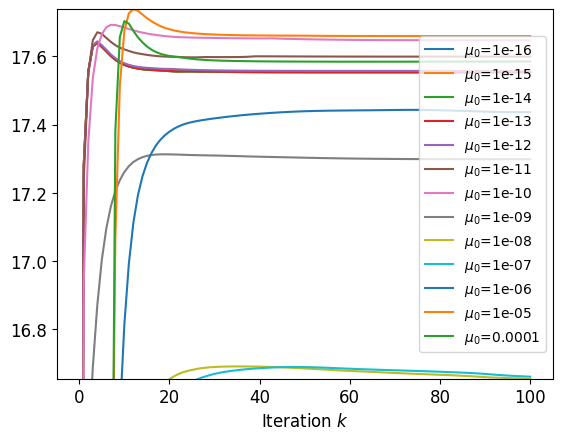} & \includegraphics[width=\size]{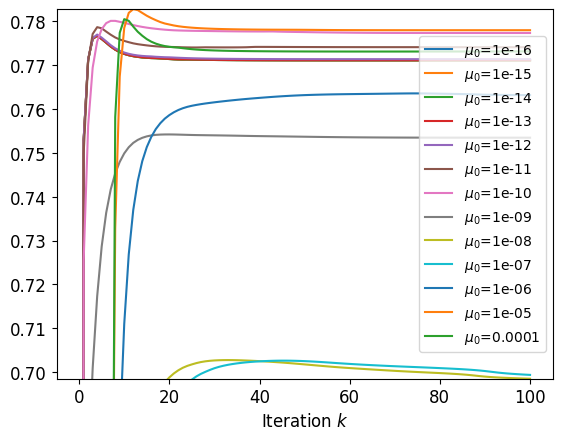} & \includegraphics[width=\size]{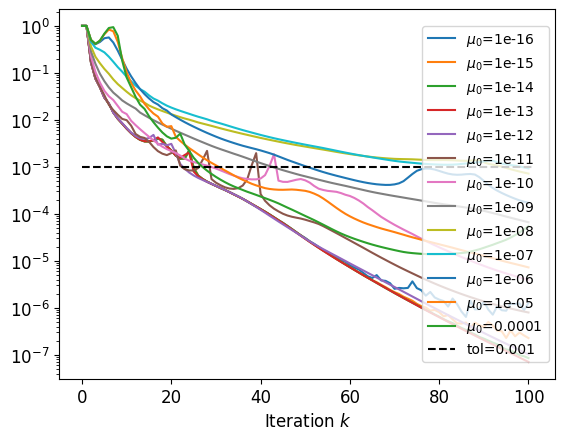} \\
		\midrule
		\makecell[l]{\rotatebox[origin=c]{90}{$U_2 (A_{45\times51})$}} & \includegraphics[width=\size]{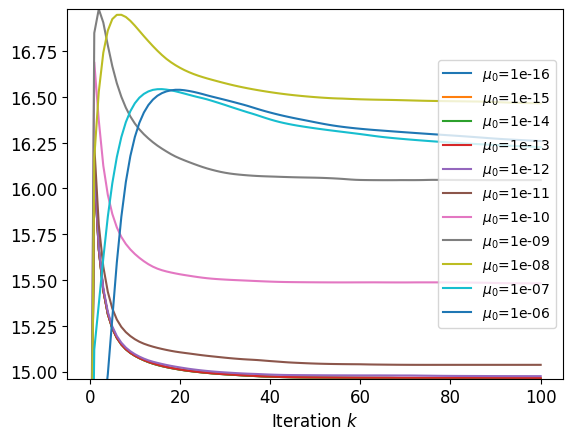} & \includegraphics[width=\size]{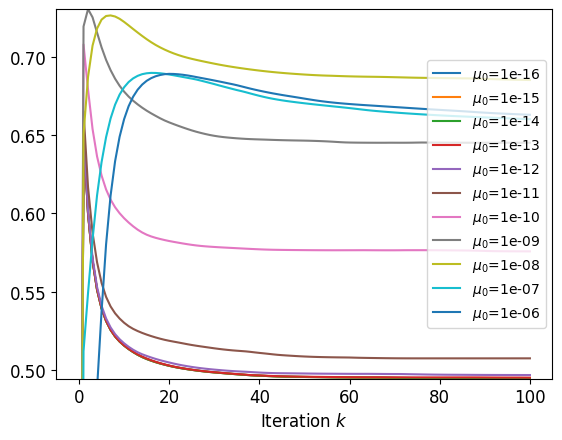} & \includegraphics[width=\size]{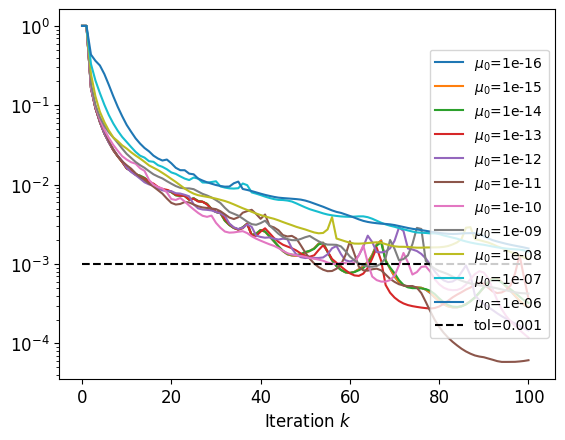} \\
		\bottomrule
	\end{tabularx}
	\caption{Average PSNR, SSIM and relative $L^2$-norm curves over the selected phantoms in the MPI-MNIST dataset for a variety of $\mu_0$ parameters for the three methods. We observe that in most cases the maximum PSNR value is reached before the tolerance is reached by the $L^2$-norm criterion.}
	\label{tab:curves:mnist}
\end{table}

\begin{table}[!thb]
	\centering
	\begin{tabular}{ccccccc}
		\toprule
		& $s$ & Reco. grid     & $\mu_0$      & $\nit$ & PSNR ($\uparrow$) & SSIM ($\uparrow$)\\
		\midrule
		$\mathrm{Super}_2(A_{45\times51})$ & 2   & $90\times 102$ & $\num{e-5}$  & 12     & $17.74 \pm 1.19 $ & $0.7827 \pm 0.0311  $\\
		$U_2(A_{45\times51})$              & 2   & $90\times 102$ & $\num{e-9}$  & 2      & $16.98\pm 1.18$   & $0.7305 \pm 0.0356$\\
		\bottomrule
	\end{tabular}
	\caption{Results of the validation on the MPI-MNIST dataset displayed in table \ref{tab:curves:mnist}. }
	\label{tab:mnist}
\end{table}

\def\size{6em}
\newcolumntype{C}{>{\centering\arraybackslash}m{6em}}
\newcolumntype{L}{>{\raggedright\arraybackslash}m{8em}}  
\begin{table}[!t]
	\centering
	\setlength{\tabcolsep}{0pt}
	\begin{tabular}{L*6{C}@{}}
		\toprule
		GT $15\times 17$ & \includegraphics[width=\size]{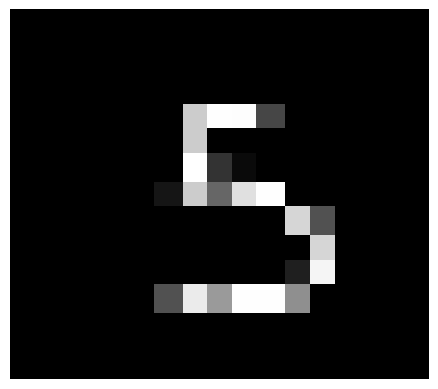} & \includegraphics[width=\size]{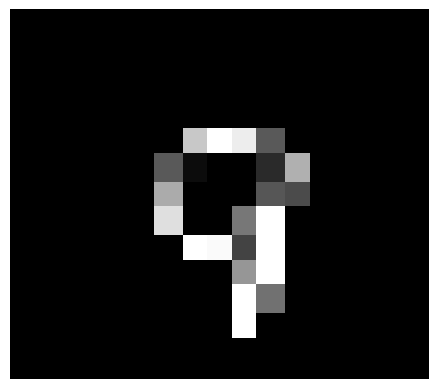} & \includegraphics[width=\size]{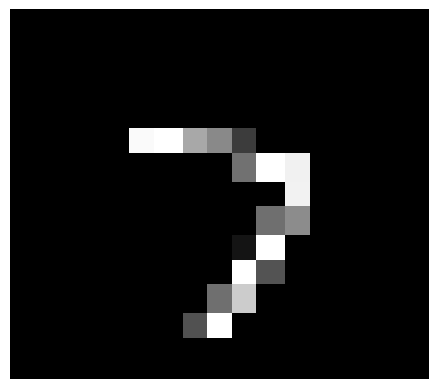} & \includegraphics[width=\size]{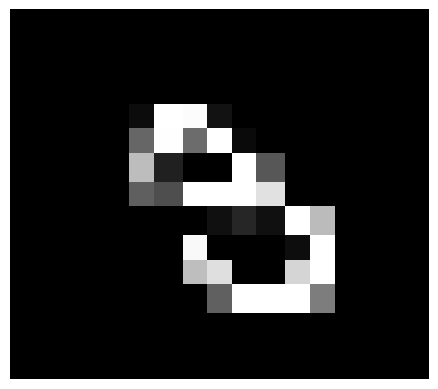} & \includegraphics[width=\size]{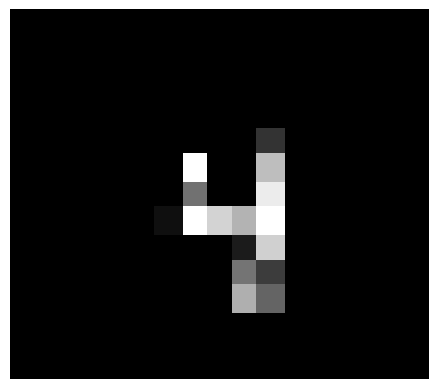}\\
		\midrule
		$\mathrm{Super}_2 (A_{45\times51} )$& \includegraphics[width=\size]{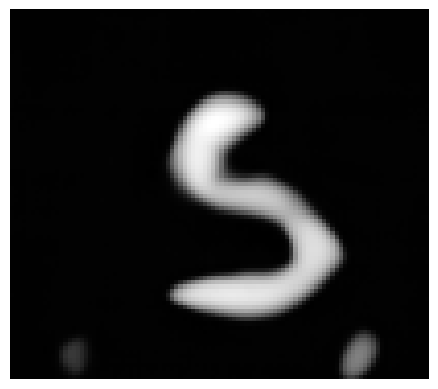} & \includegraphics[width=\size]{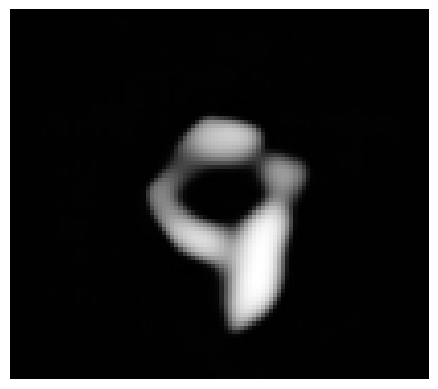} & \includegraphics[width=\size]{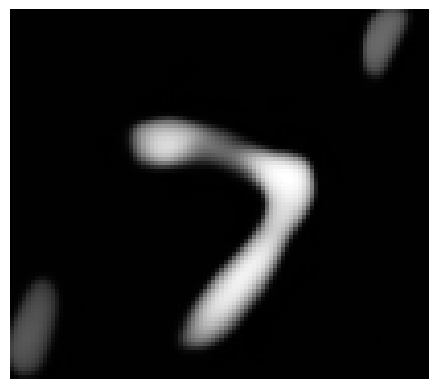} & \includegraphics[width=\size]{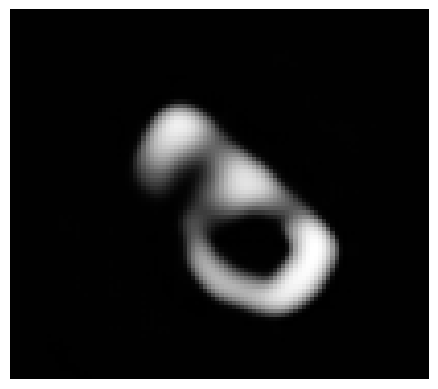} & \includegraphics[width=\size]{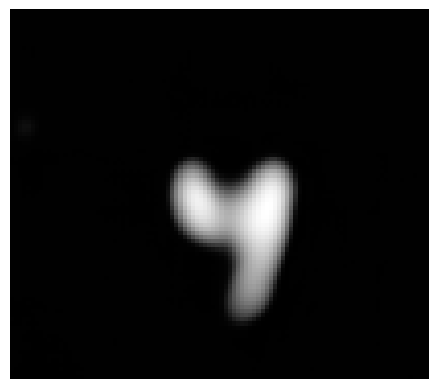}\\
		$U_2 (A_{45\times51} )$& \includegraphics[width=\size]{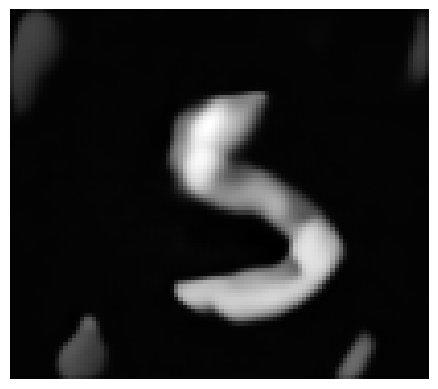} & \includegraphics[width=\size]{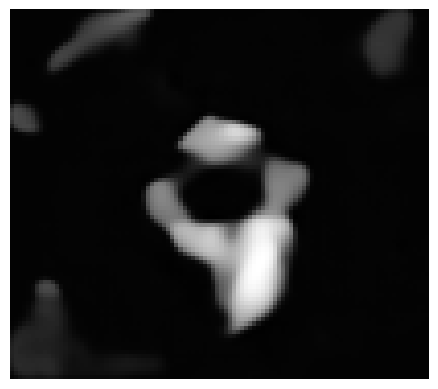} & \includegraphics[width=\size]{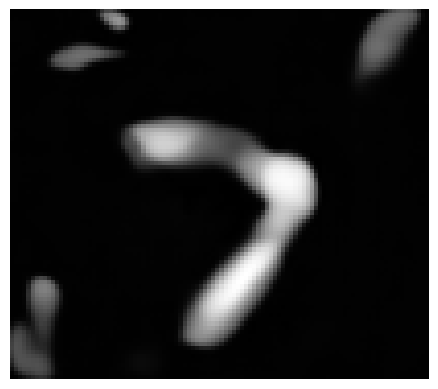} & \includegraphics[width=\size]{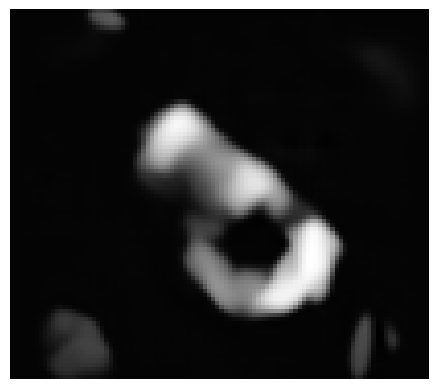} & \includegraphics[width=\size]{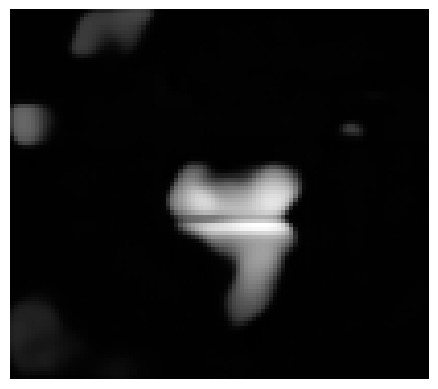}\\
		\bottomrule
	\end{tabular}
	\caption{Final reconstructions examples on five selected phantoms in the MPI-MNIST dataset. In the first row, we have the $15\times 17$ ground truths. In the second row, we display the reconstruction results obtained with \our\ algorithm with a super-resolution factor $s=2$. In the third row, we display the reconstructions obtained with the ZeroShot-PnP algorithm using the interpolated $A_{s 45\times s 51}$ system matrix. 
		Comparing the second with the third row, we observe that the results with \our\ algorithm present less artifacts than the results obtained by super-resolving the system matrix.}
	\label{tab:pics:mnist}
\end{table}

The first reconstructions we present have been performed on the first 20 
test phantoms of the MPI-MNIST dataset. The dataset offers a system 
matrix $A_{45\times 51}$, calibrated on a $45\times 51$ grid, which we 
use for reconstruction. We test the \our\ algorithm with a super-resolution 
factor of $s=2$ (cf. section \ref{subsec:super:pnp}) and 
we compare it with a baseline method. The baseline 
method~\cite{Gngr2020Superresolving} for comparison can be described 
in the following way: the reconstruction is performed using a matrix 
$A_{sN_x\times sN_y}$ which is obtained from the $A_{N_x\times N_y}$ matrix 
by interpolating its columns by a factor of $s=2$, using bilinear interpolation; 
the super-resolved system matrix $A_{sN_x\times sN_y}$ is used as forward 
operator in conjunction with a solver of linear systems. In this experiment 
we have chosen to use the ZeroShot-PnP algorithm as a regularized linear system solver 
for the system matrix $A_{sN_x\times sN_y}$. To differentiate between the 
reconstruction algorithms, we denote them with the following notation: 
we write $\text{Super}_s (A_{N_x\times N_y})$ to refer to the \our\ algorithm 
that performs an upscaling of factor $s$; coherently with the notation of the 
upscaling operator in section \ref{subsec:super:pnp}.
In contrast we write $U_s (A_{N_x\times N_y})$ to
denote reconstructions obtained by using the super-resolved system matrix 
$A_{{sN_x}\times {sN_y}}$, i.e., the baseline method.

We have performed reconstructions with starting parameters $\mu_0 = 10^{i}$ 
for $i= -16, \dots , -4$ for 100 iterations and computed PSNR and SSIM of all 
iterations. Additionally, we have computed the relative $L^2$-norm 
$\frac{\norm{x^{k+1}_{u} - x^{k}_{u}}_2 }{\norm{x^{k+1}_{u}}_2}$ for each 
phantom, parameter and iteration $k$. The stopping criterion is the following: 
we have considered the iteration $k^*$ and parameter $\mu_0^*$ to be optimal 
if the average PSNR is maximal \emph{and} the average relative $L^2$-norm is 
below $10^{-3}$. The average PSNR curves, the relative average SSIM curves and 
the relative $L^2$-errors are displayed in table \ref{tab:curves:mnist}. 
The optimal parameters as well as the number of iterations selected with 
this validation and the relative average PSNR and SSIM scores are displayed 
in table \ref{tab:mnist}. From table \ref{tab:mnist} we observe that the PSNR 
and the SSIM values obtained with \our\ method are higher than the ones obtained 
with the baseline method that uses a super-resolved system matrix 
($U_2 (A_{45\times 51})$). This result shows quantitatively that the \our\ 
algorithm produces improved reconstruction results compared with more standard 
super-resolution techniques. For a qualitative evaluation of the results, 
we display the final reconstructions as well as the ground truths of 5 phantoms 
in table \ref{tab:pics:mnist}. In particular, we observe that, from a visual 
standpoint, the result with the super-resolved matrix $U_2 (A_{45\times 51})$ 
contained a series of reconstruction artifacts that are not present in the 
reconstructions with the \our\ algorithm.
In section \ref{subsec:datasets} we argue that the MPI-MNIST dataset is  
not properly suitable to test super-resolution because of the way the dataset 
is produced. We recall briefly that the ground truths in the MPI-MNIST dataset 
live natively on a $15\times 17$ grid and are upscaled by nearest neighbors when 
producing the data scans. However, the nearest neighbor interpolation preserves 
the native $15\times 17$ scale of the features of the phantom. Consequently, 
the ground truths thus produced have in principle no fine(r) features living 
on the $45\times 51$ or $90\times 102$ scales on which we reconstruct. 
In particular, it is reasonable to expect that the employment of bilinear 
interpolation results in a smoothing out of the $15\times 17$ features, 
when applied to recover potential features on more finely resolved grids. 
Coherently, it is reasonable to expect that, if there are no higher-level 
features to be extracted, the PSNR and SSIM score of super-resolved 
reconstruction could be lower than reconstruction without super-resolution. 
To verify this fact and corroborate our argument in favor of using the MPI-CHAR 
dataset, we have performed the reconstructions without super-resolution, i.e., 
with $s=1$ and for which the methods $\text{Super}_1 (A_{45\times 51})$ and 
$U_1 (A_{45\times 51})$ coincide. For this non super-resolved reconstruction 
we have obtained the validated parameters $\mu_0=\num{e-4}$, 
$n_{\mathrm{it}}=10$ and PSNR values of $18.18\pm 1.20$, whereas the SSIM 
are $0.800 \pm 0.0356$. Comparing these scores with the one in table 
\ref{tab:mnist}, we observe that neither \our\ algorithm nor the baseline 
method $U_2 (A_{45\times 51})$ achieve PSNR and SSIM score higher than the 
non-super-resolved reconstruction on the $45\times 51$ grid. 
This confirms our hypothesis that this dataset -- which lacks higher resolution features by construction --
is not fully suitable to investigate super-resolution methods.
Given this observation, we have employed the data 
provided with the MPI-MNIST dataset to produce a higher resolution set of 
ground truths to test super-resolution, the MPI-CHAR dataset. In the next 
experiment we perform reconstructions on the MPI-CHAR dataset and obtain 
higher PSNR and SSIM scores than without super-resolution, confirming in hindsight 
that the MPI-MNIST dataset is not fully suitable to test super-resolution.

\subsection{Reconstructions on the MPI-CHAR Dataset}\label{subsec:exp:mpichar}

\def\size{14em}
\newcolumntype{C}{>{\centering\arraybackslash}m{\size}}
\newcolumntype{L}{>{\raggedright\arraybackslash}X}
\begin{table}[!tb]
	\centering
	\setlength{\tabcolsep}{0pt}
	
	\begin{tabularx}{\linewidth}{L*3{C}@{}}
		\toprule
		& PSNR & SSIM & Rel. $\ell^2$-norm \\
		\midrule
		\makecell[l]{\rotatebox[origin=c]{90}{ZS-PnP($A_{15\times17}$)}}& \includegraphics[width=\size]{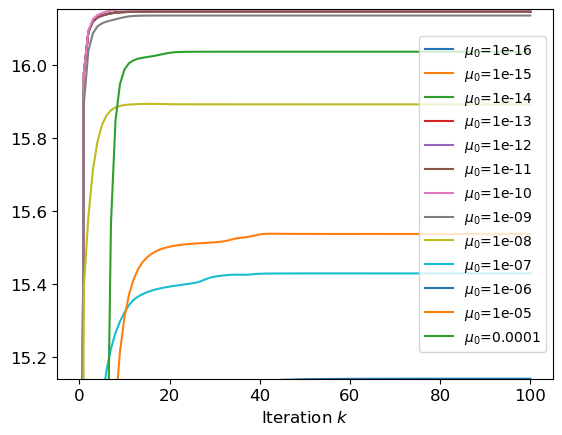} & \includegraphics[width=\size]{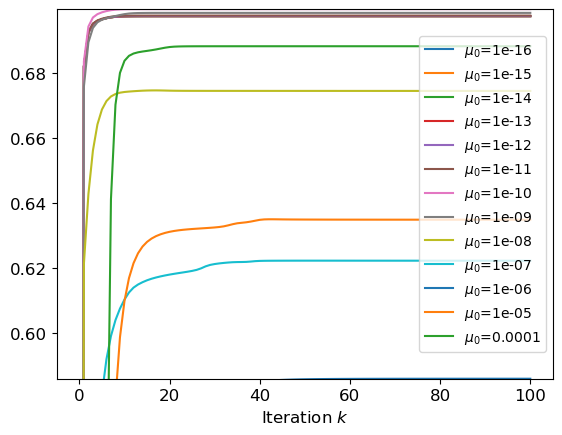} & \includegraphics[width=\size]{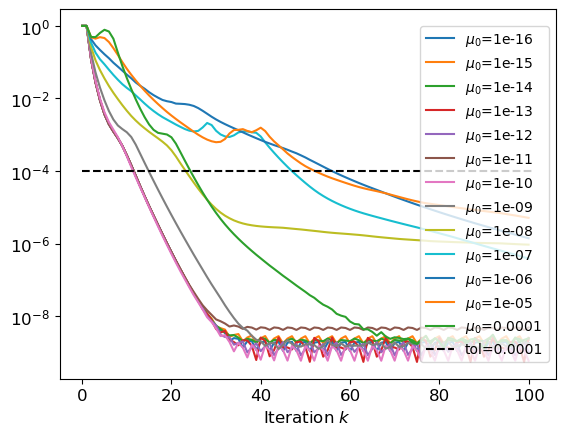}\\
		\midrule
		\makecell[l]{\rotatebox[origin=c]{90}{$\mathrm{Super}_3 (A_{15\times 17} )$}} & \includegraphics[width=\size]{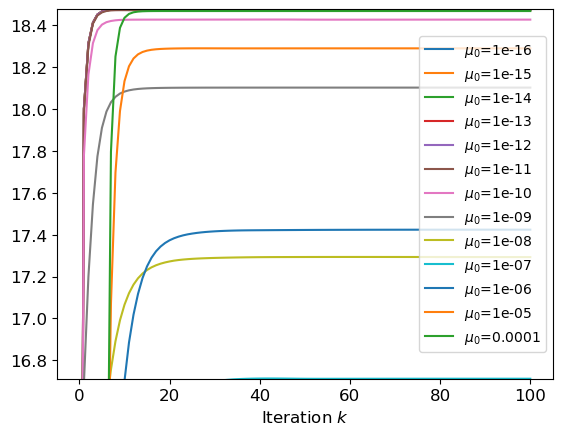} & \includegraphics[width=\size]{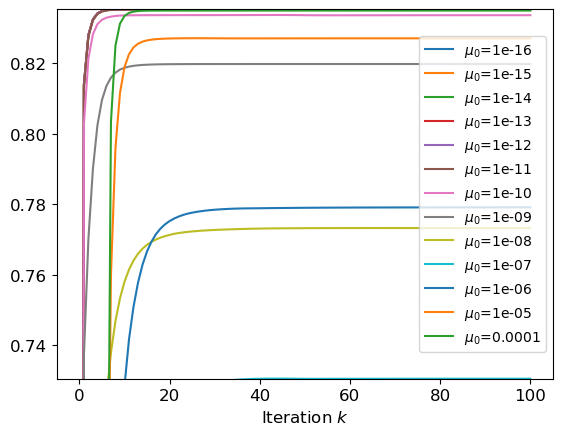} & \includegraphics[width=\size]{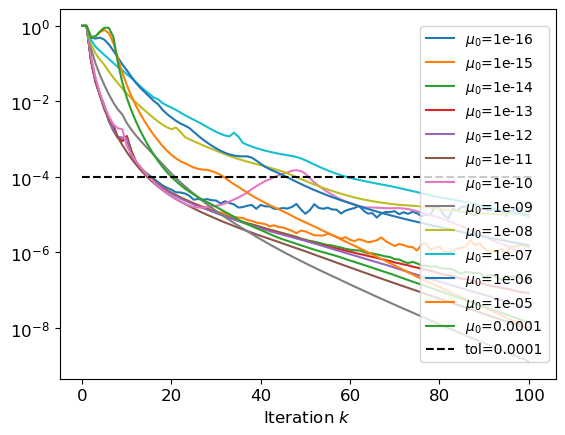} \\
		\midrule
		\makecell[l]{\rotatebox[origin=c]{90}{$\mathrm{Super}_5 (A_{15\times 17} )$}} & \includegraphics[width=\size]{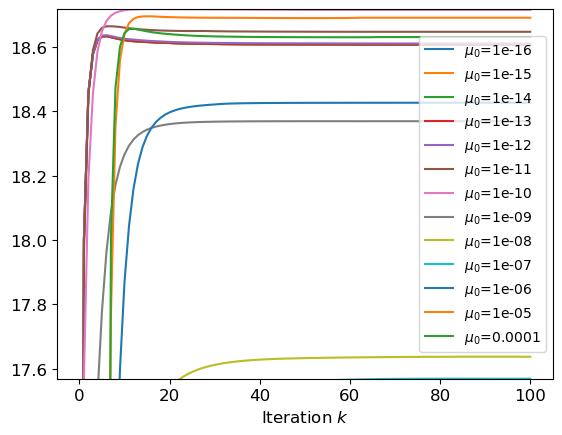} & \includegraphics[width=\size]{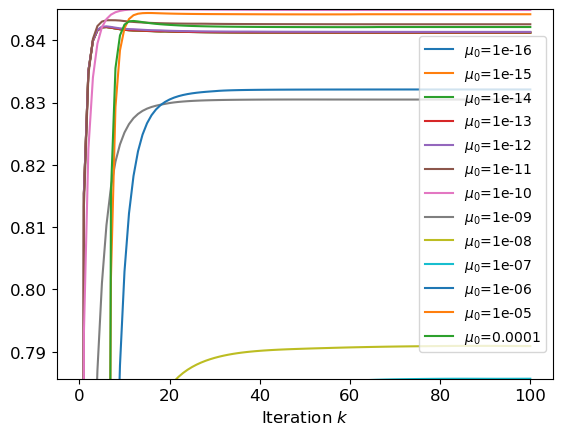} & \includegraphics[width=\size]{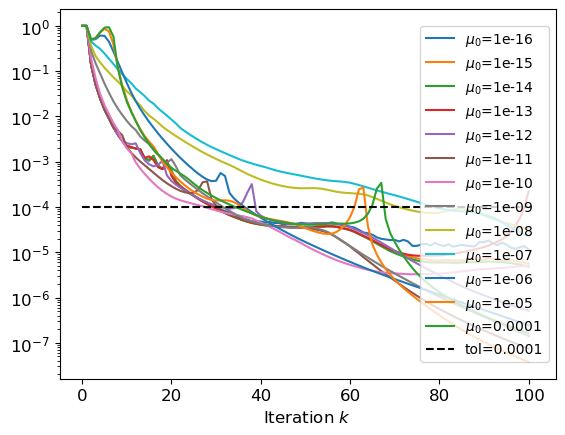} \\
		\bottomrule
	\end{tabularx}
	\caption{Average PSNR, SSIM and relative $L^2$-norm curves over the selected phantoms in the MPI-CHAR 
		dataset for a variety of $\mu_0$ parameters for the three methods.}
	\label{fig:curves:lett}
\end{table}

\begin{table}[!thb]
	\centering
	\begin{tabular}{ccccccc}
		\toprule
		& $s$ & Reco. grid    & $\mu_0$   &   $\nit$ & PSNR ($\uparrow$) & SSIM ($\uparrow$)\\
		\midrule
		ZS-PnP($A_{15\times 17}$)              &  1  & $15\times 17$ & $\num{e-10}$  & 12     & $16.15\pm 2.76$ & $0.6996 \pm 0.0858$\\
		$\mathrm{Super}_3 (A_{15\times 17} )$  &  3  & $45\times 51$ & $\num{e-16}$  & 11      & $18.48 \pm 2.30 $ & $0.8354 \pm 0.0686$\\
		$\mathrm{Super}_5 (A_{15\times 17} )$  &  5  & $75\times 85$ & $\num{e-10}$  & 11      & $18.72 \pm 2.26$ & $0.8449 \pm 0.0671 $\\
		\bottomrule
	\end{tabular}
	\caption{Results of the validation procedure in table~\ref{fig:curves:lett} on the MPI-CHAR dataset. 
		We observe that compared with the low-resolved reconstruction using $A_{15\times 17}$, 
		the employment of \our\ yields higher PSNR and SSIM scores.}
	\label{tab:lett}
\end{table}

\def\size{6em}
\newcolumntype{C}{>{\centering\arraybackslash}m{6em}}
\newcolumntype{L}{>{\raggedright\arraybackslash}m{8em}}  
\begin{table}[!thb]
	\centering
	\setlength{\tabcolsep}{0pt}
	\begin{tabular}{L*6{C}@{}}
		\toprule
		GT $75\times 85$ & \includegraphics[width=\size]{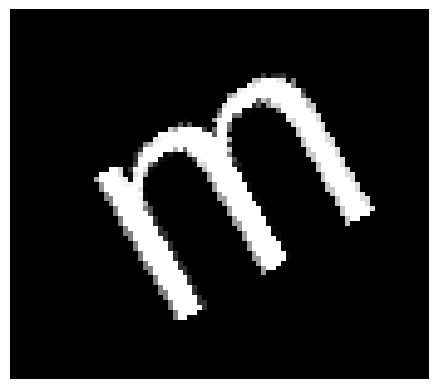} & \includegraphics[width=\size]{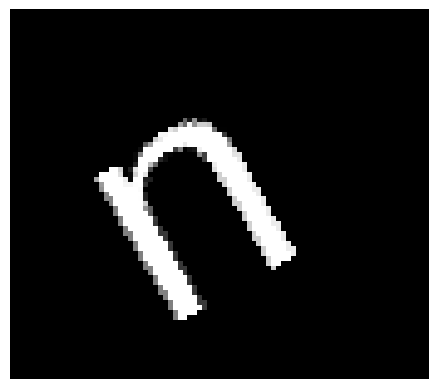} & \includegraphics[width=\size]{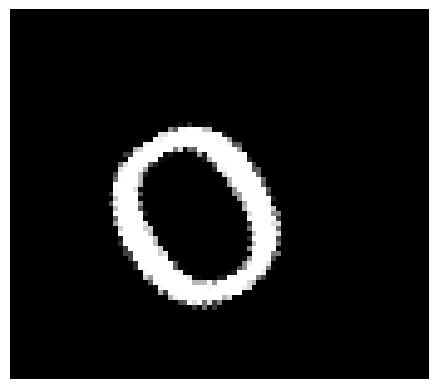} & \includegraphics[width=\size]{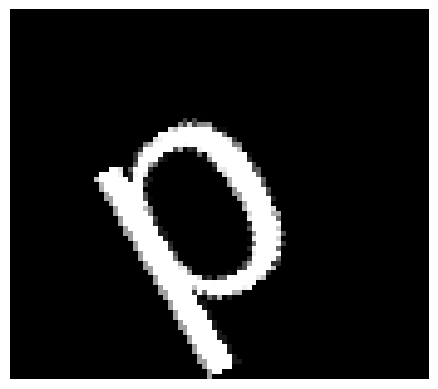} & \includegraphics[width=\size]{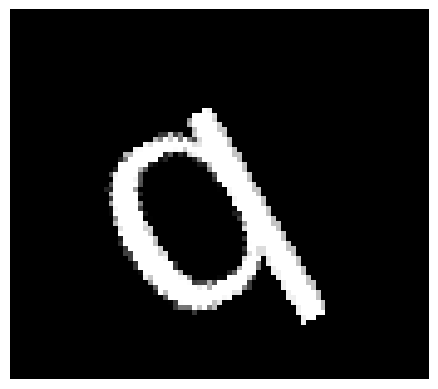}\\
		\midrule
		ZS-PnP$(A_{15\times 17})$ & \includegraphics[width=\size]{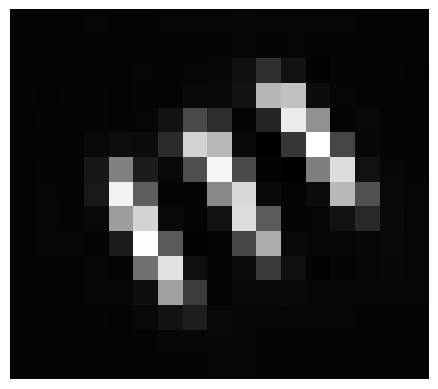} & \includegraphics[width=\size]{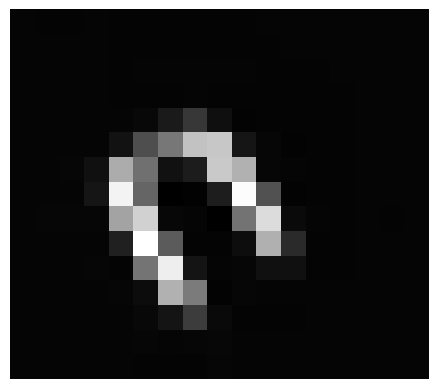} & \includegraphics[width=\size]{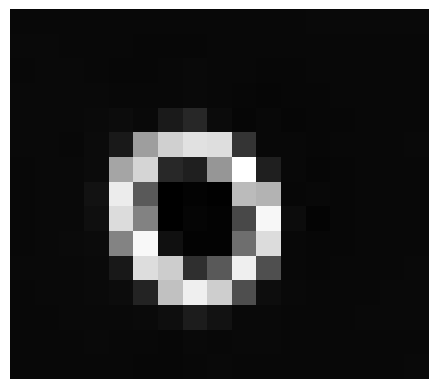} & \includegraphics[width=\size]{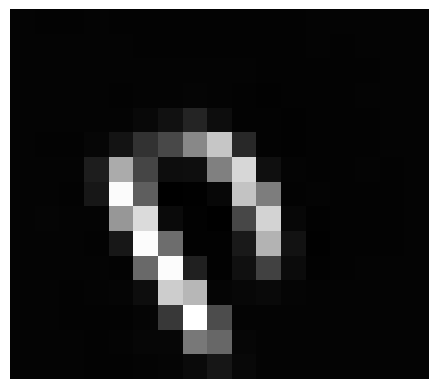} & \includegraphics[width=\size]{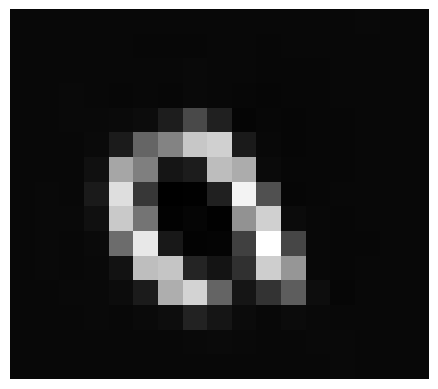}\\
		$\mathrm{Super}_3 (A_{15\times 17})$& \includegraphics[width=\size]{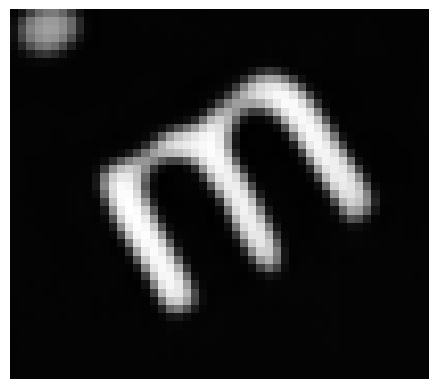} & \includegraphics[width=\size]{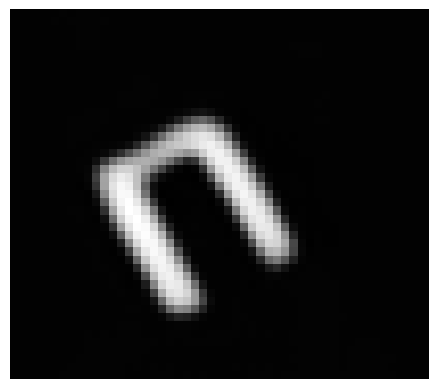} & \includegraphics[width=\size]{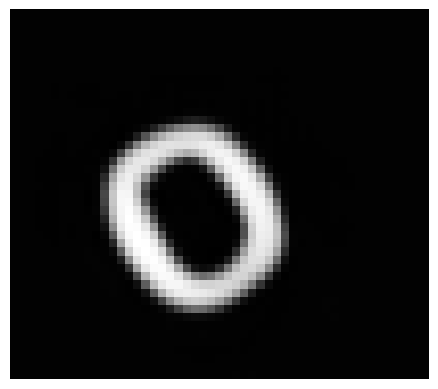} & \includegraphics[width=\size]{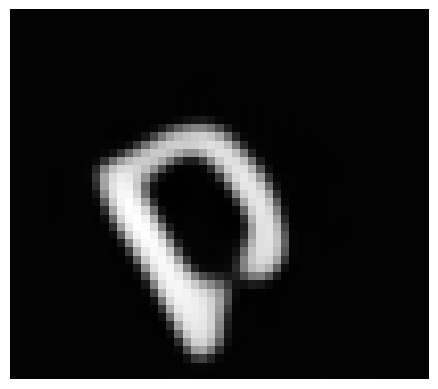} & \includegraphics[width=\size]{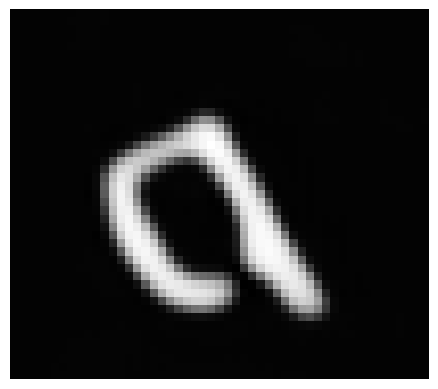}\\
		$\mathrm{Super}_5 (A_{15\times 17} )$& \includegraphics[width=\size]{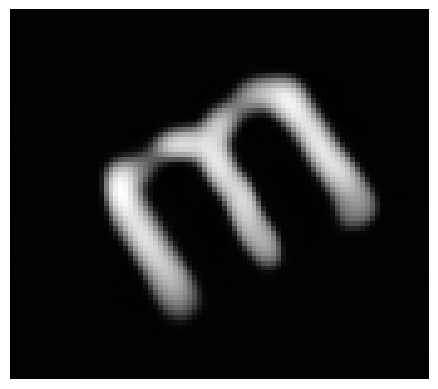} & \includegraphics[width=\size]{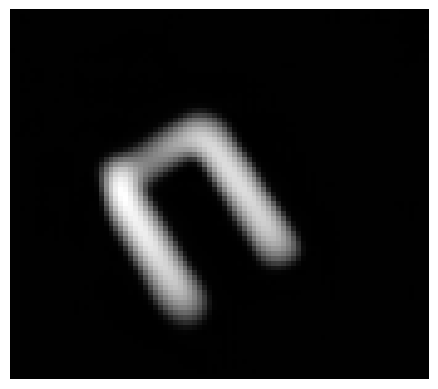} & \includegraphics[width=\size]{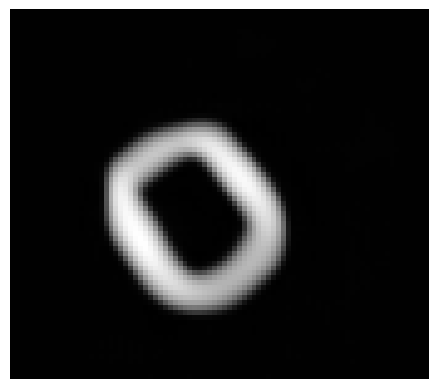} & \includegraphics[width=\size]{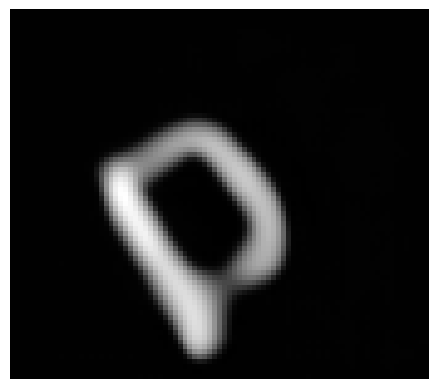} & \includegraphics[width=\size]{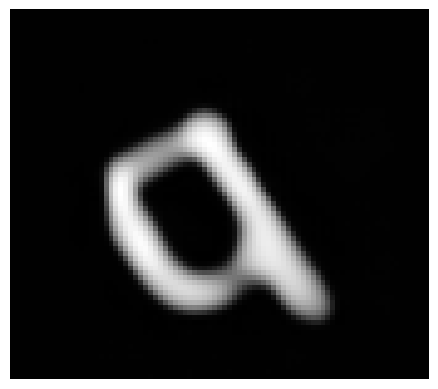}\\
		\bottomrule
	\end{tabular}
	\caption{A selection of final reconstruction results on the MPI-CHAR dataset. 
		In the first (top) row we display the ground truths on the native $75\times 85$ scale. 
		In the other rows, we display reconstructions obtained using the $A_{15\times 17}$.
		In the second row, no super-resolution has been introduced, where in the third and 
		fourth row we display reconstruction with \our\ using a scaling factor $s$ of 3 and 5, 
		respectively.}
	\label{fig:examples:lett}
\end{table}

In the previous experiment we have shown that, compared with the baseline method 
(super-resolving the system matrix), the \our\ algorithm is capable to yield reconstructions 
with higher PSNR and SSIM values as well as with less reconstruction 
artifacts, proving the advantages of using \our . We have also observed that the 
MPI-MNIST dataset contains ground truths 
that natively contain features that live on a $15\times 17$ grid and that consequently, 
is not suitable to properly test the results of super-resolution. In this experiment 
we show that the \our\ is capable of yielding improved reconstruction results from 
coarse system matrix in a simulated scenario. These improvements are demonstrated 
both quantitatively and qualitatively.
The dataset utilized has been obtained in the following way: a set of 94 ground truths 
of size $75\times 85$ is obtained using the letters from on an openly available font 
dataset described in \ref{subsec:datasets}; from the MPI-MNIST 
dataset~\cite{iske2025mpimnist,iske2025learneddiscrepancy} we have considered the 
clean system matrix $A_{75\times 85}$ and real noise samples $\eta$ provided in 
the same dataset; the simulated scans are then obtained for each ground 
truth $x\in\mathbb{R}^{75\cdot 85}$ by
\begin{equation}
	y = A_{75\times 85}x + \eta  .
\end{equation}
To test the super-resolution reconstruction we have considered the coarse 
system matrix $A_{15\times 17}$ for the reconstruction. First, we simply reconstruct 
with $s=1$, i.e., with the ZeroShot-PnP algorithm (we denote the results coherently 
with section \ref{subsec:exp:mpimnist} as ZS-PnP). The outputs of size $15\times 17$ 
are upscaled with zero-order upscaling to $75\times 85$ to compute the PSNR. Subsequently, 
we test the \our\ algorithm with $s=3$ and $s=5$, and compare it upon upscaling with the 
ground truths to compute the PSNR. The reconstruction with the  \our\ algorithm are denoted 
with $\text{Super}_s$ for the upscaling factor $s$. We have plotted the average PSNR, 
SSIM and $L^2$-norm curves for each of the starting parameter $\mu_0 = 10^{i}$ with 
$i = -16, \cdots , -4$ in table \ref{fig:curves:lett}. Coherently with the previous experiment, 
we have chosen the optimal parameters that maximize the PSNR value before the relative $L^2$-norm 
reaches a tolerance of $10^{-4}$. The corresponding validated parameters and final average PSNR 
and SSIM values are displayed in table \ref{tab:lett}. From the results in table \ref{tab:lett} 
we observe that, when the underlying ground truth has features that natively lives on a high-resolution 
scale ($75\times 85$), the super-resolution effect of \our\ algorithm yields higher PSNR and SSIM. 
This shows quantitatively that the proposed \our\ algorithm is beneficial in retrieving finer 
features of the underlying phantoms. The improvement in the reconstruction quality is also 
qualitatively supported by the reconstructed examples displayed in table \ref{fig:examples:lett}. 
From table \ref{fig:examples:lett} we can make two observations. The first observation is that, 
as expected, when the ground truth natively lives on a higher resolution grid ($75\times 85$), 
then reconstructions with a lower-resolution system matrix ($A_{15\times 17}$) corresponds to a 
loss in the reconstructed features (cf. row one and two in table~\ref{tab:lett}). This observation 
supports in retrospect the observation that the MPI-MNIST is not properly suitable to test super-resolution 
methods. Additional quantitative support to this fact are the higher PSNR and SSIM values of the 
super-resolved reconstructions when compared to the low-resolution reconstructions on the 
$15\times 17$ grid (cf. the values of the second and third row in table~\ref{tab:lett} with the first row). 
The second observation is that the reconstructions obtained with the \our\ algorithm contain higher resolution
features of the ground truth even when using the low-resolution system matrix $A_{15\times 17}$. 
This second observation supports, in a simulated scenario, the usage of \our\ for super-resolution 
in MPI. In order to show that the super-resolution capabilities of \our\ are transferable to real 
data, we show next reconstructions on openly available datasets containing real MPI scans.

\subsection{Reconstructions on the 2D OpenMPI Dataset}\label{subsec:exp:openmpi}

\def\size{6em} 
\newcolumntype{C}{>{\centering\arraybackslash}m{8em}} \newcolumntype{L}{>{\raggedright\arraybackslash}m{8em}} 
\begin{table}[!th] 
	\centering 
	\setlength{\tabcolsep}{0pt} 
	\begin{tabular}{L*3{C}@{}} 
		\toprule 
		& Shape & Resolution & Concentration \\ 
		\midrule 
		ZS-PnP$(A_{37\times 37})$ & \includegraphics[width=\size]{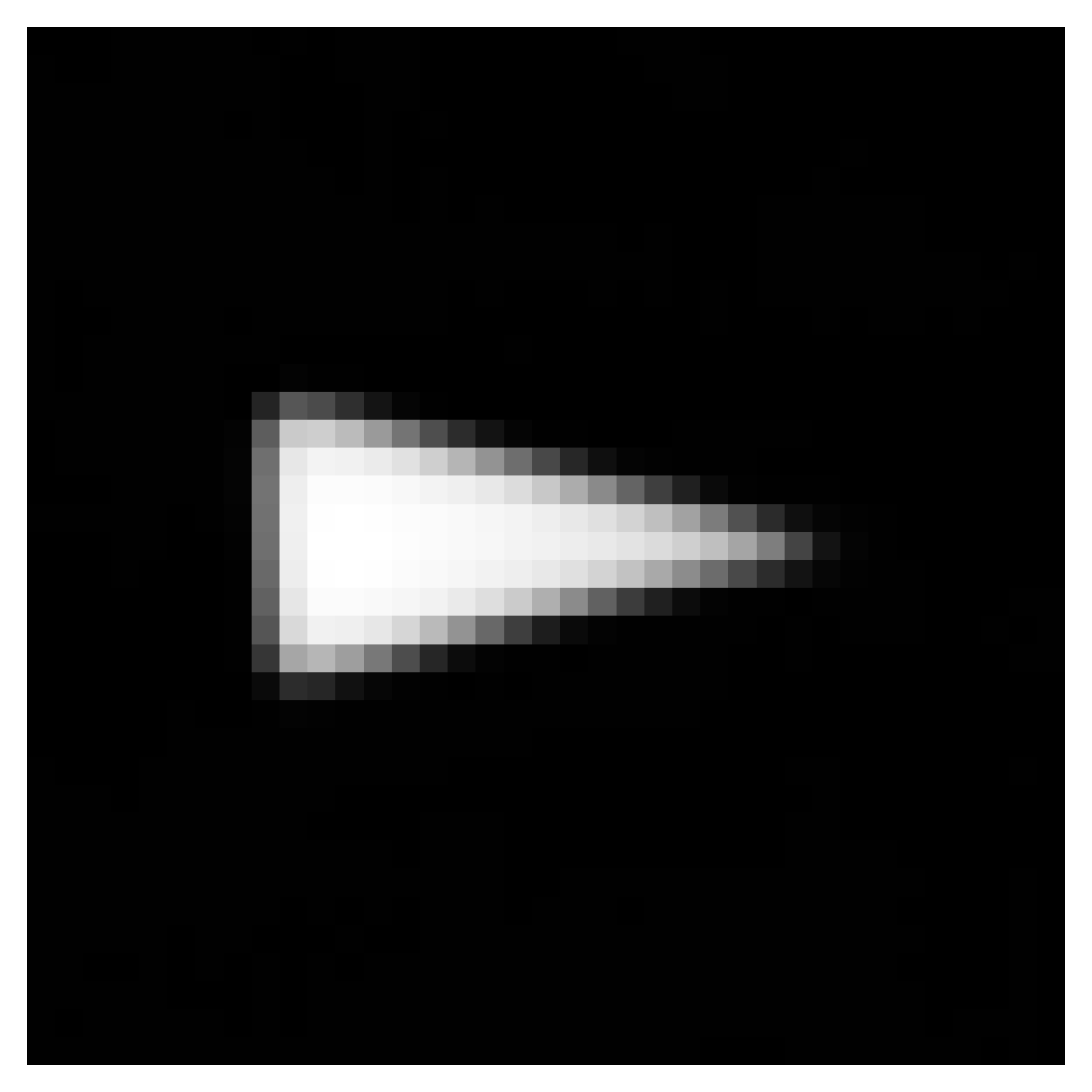} & \includegraphics[width=\size]{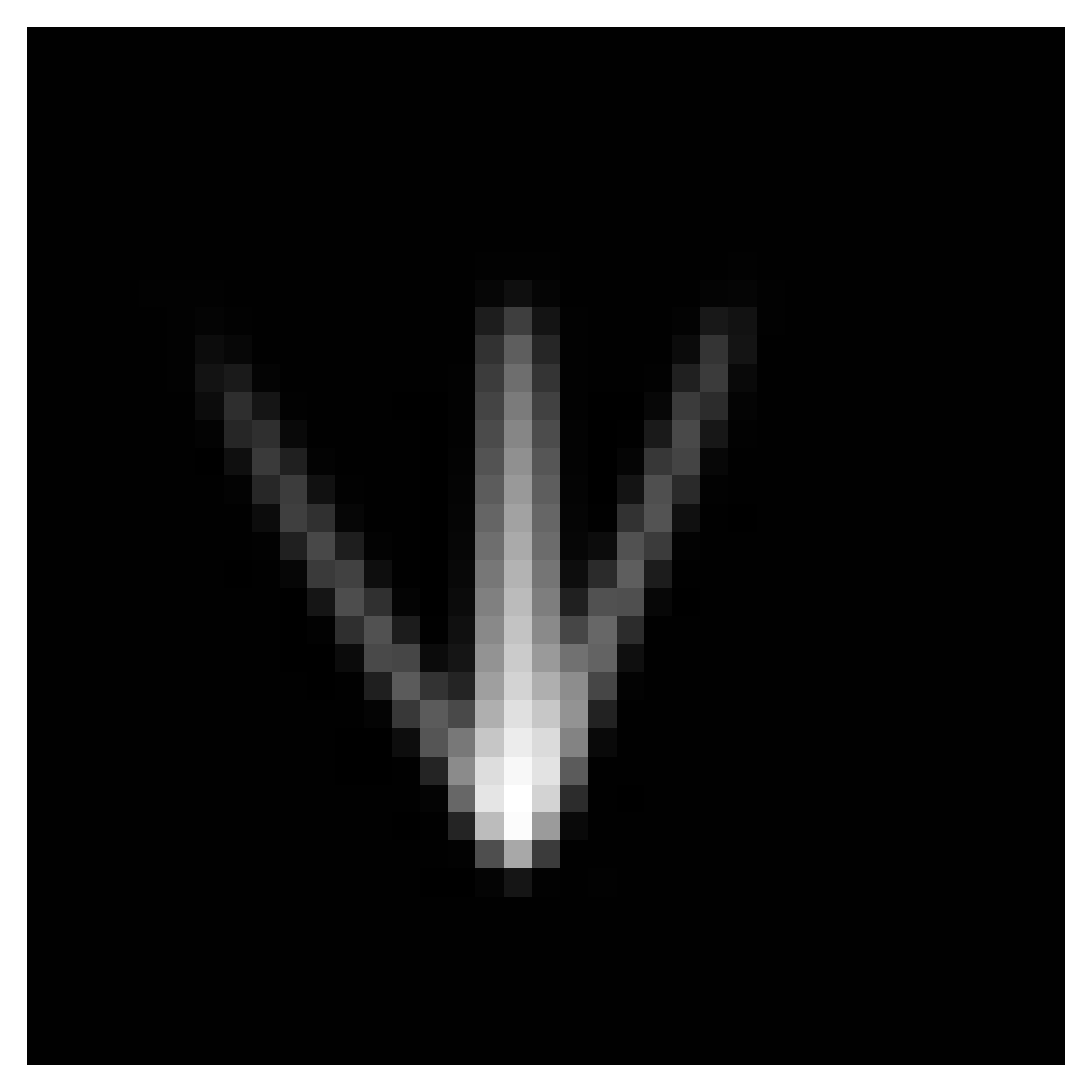} & \includegraphics[width=\size]{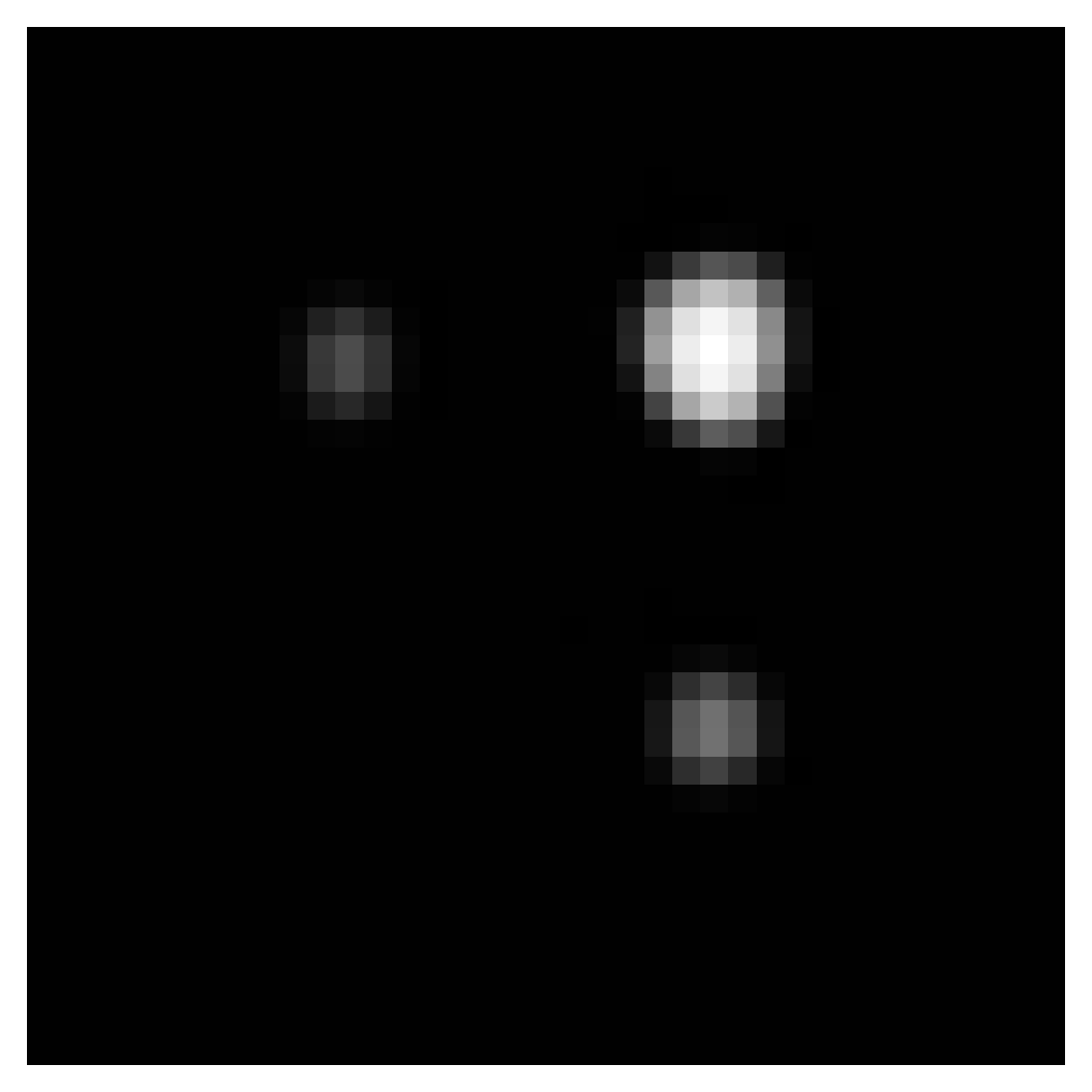} \\
		ZS-PnP$(A_{19\times 19})$ & \includegraphics[width=\size]{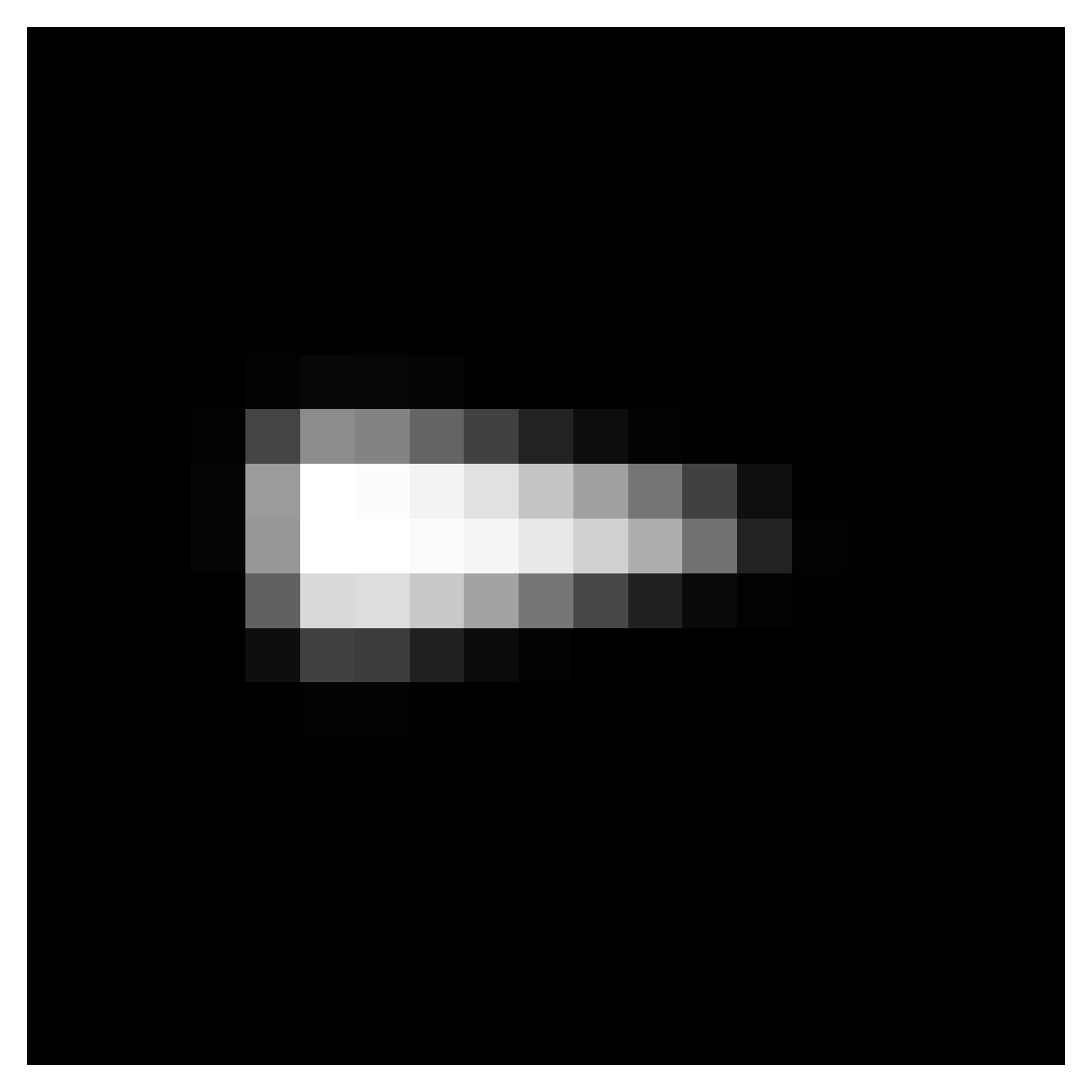} & \includegraphics[width=\size]{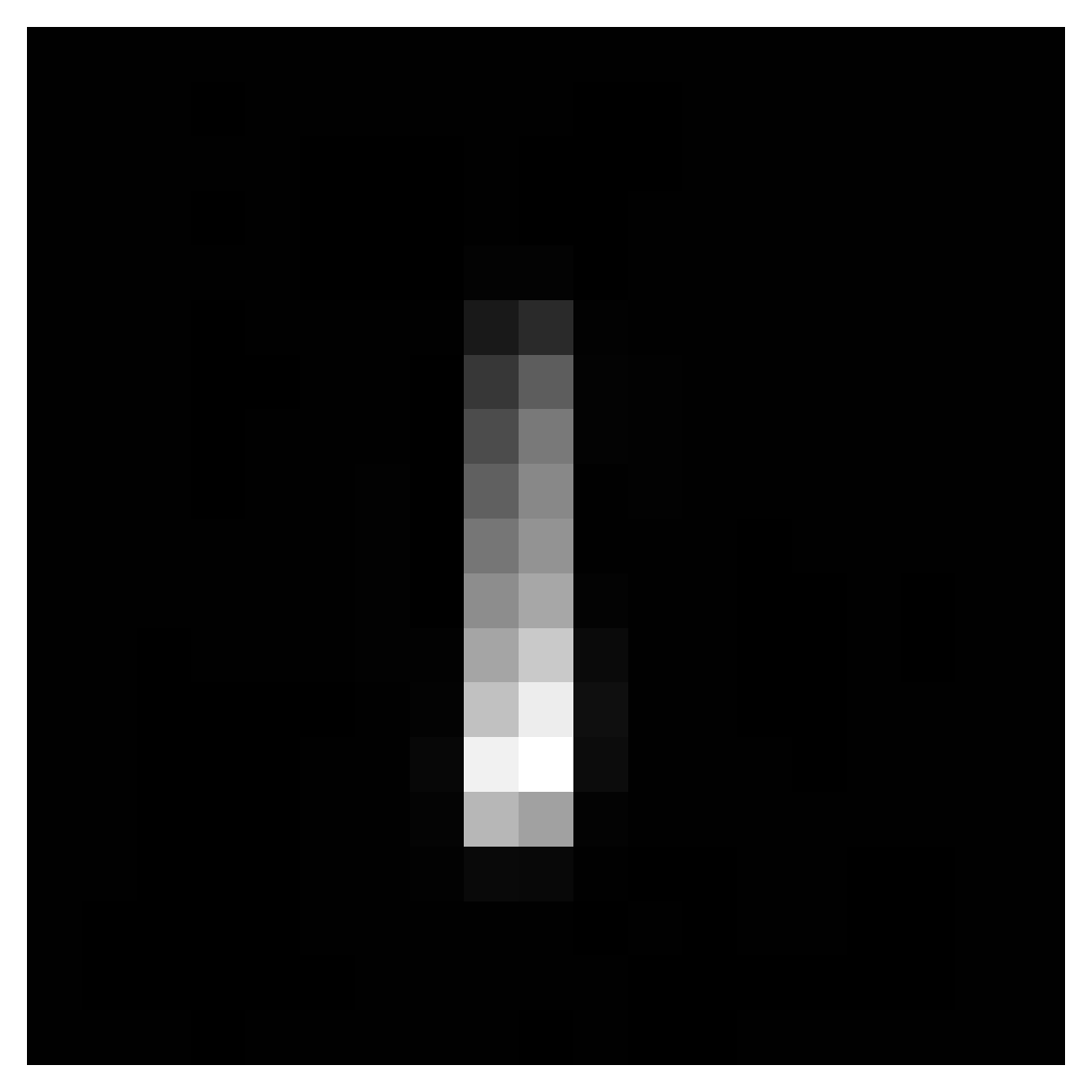} & \includegraphics[width=\size]{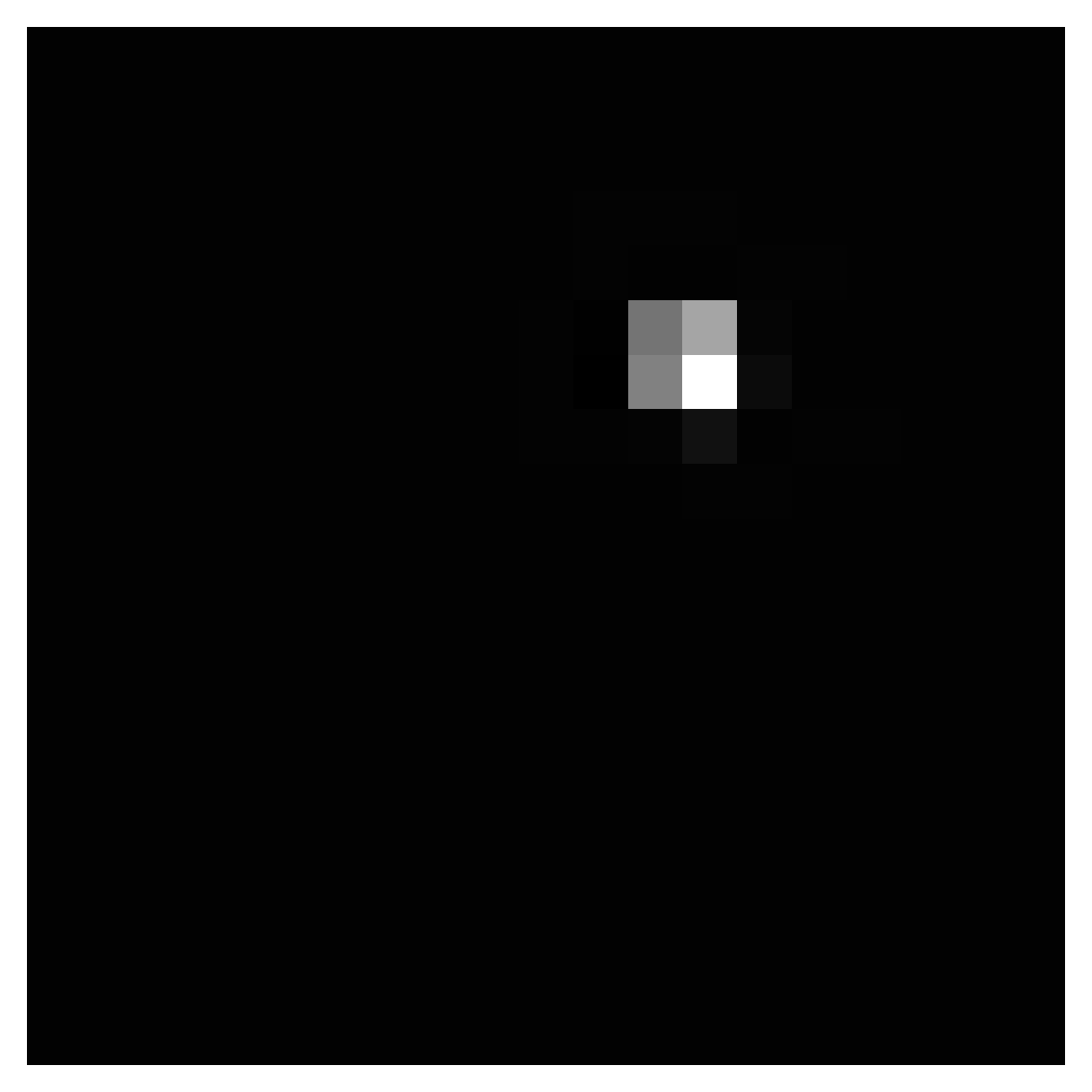} \\
		$\mathrm{Super}_2 (A_{19\times 19} )$ & \includegraphics[width=\size]{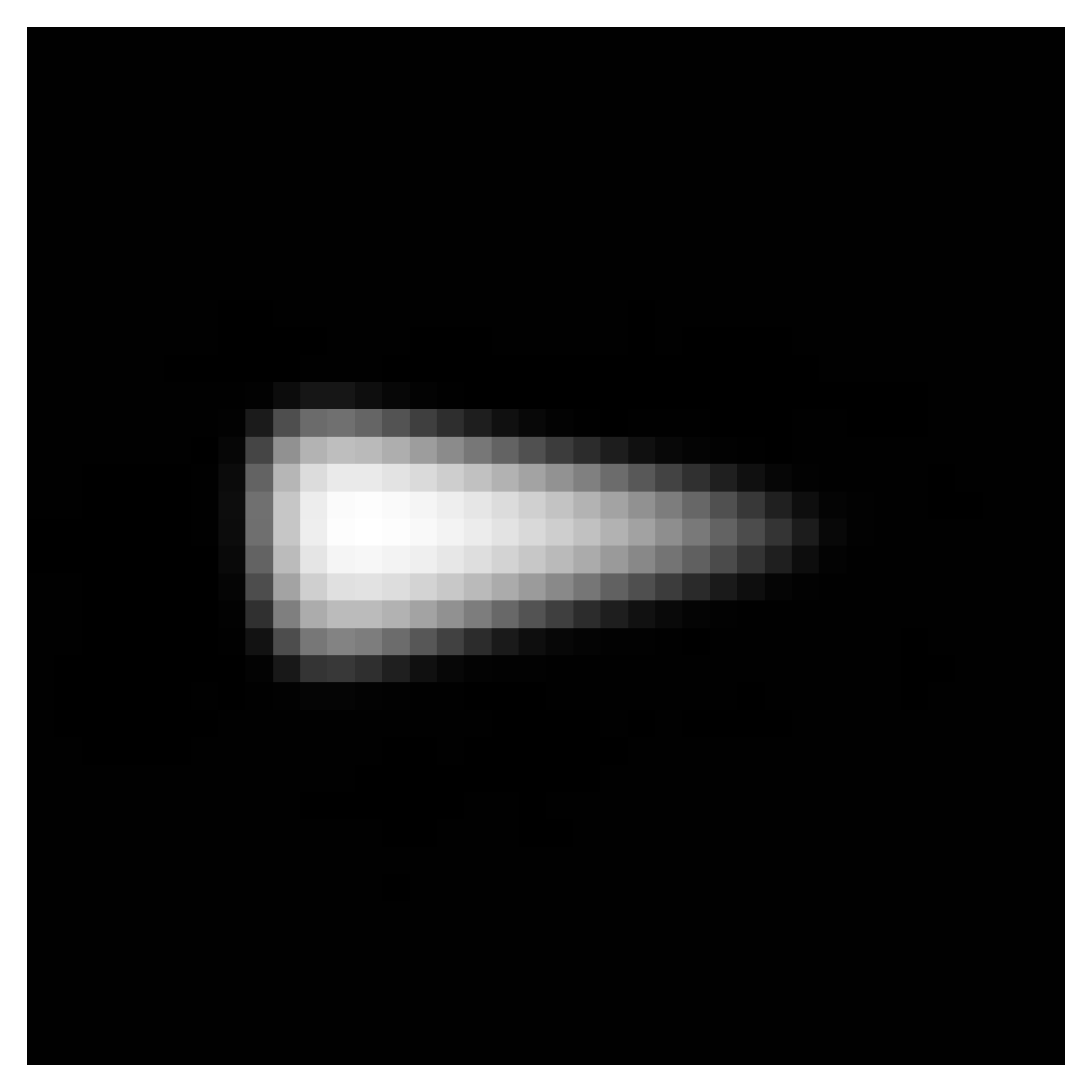} & \includegraphics[width=\size]{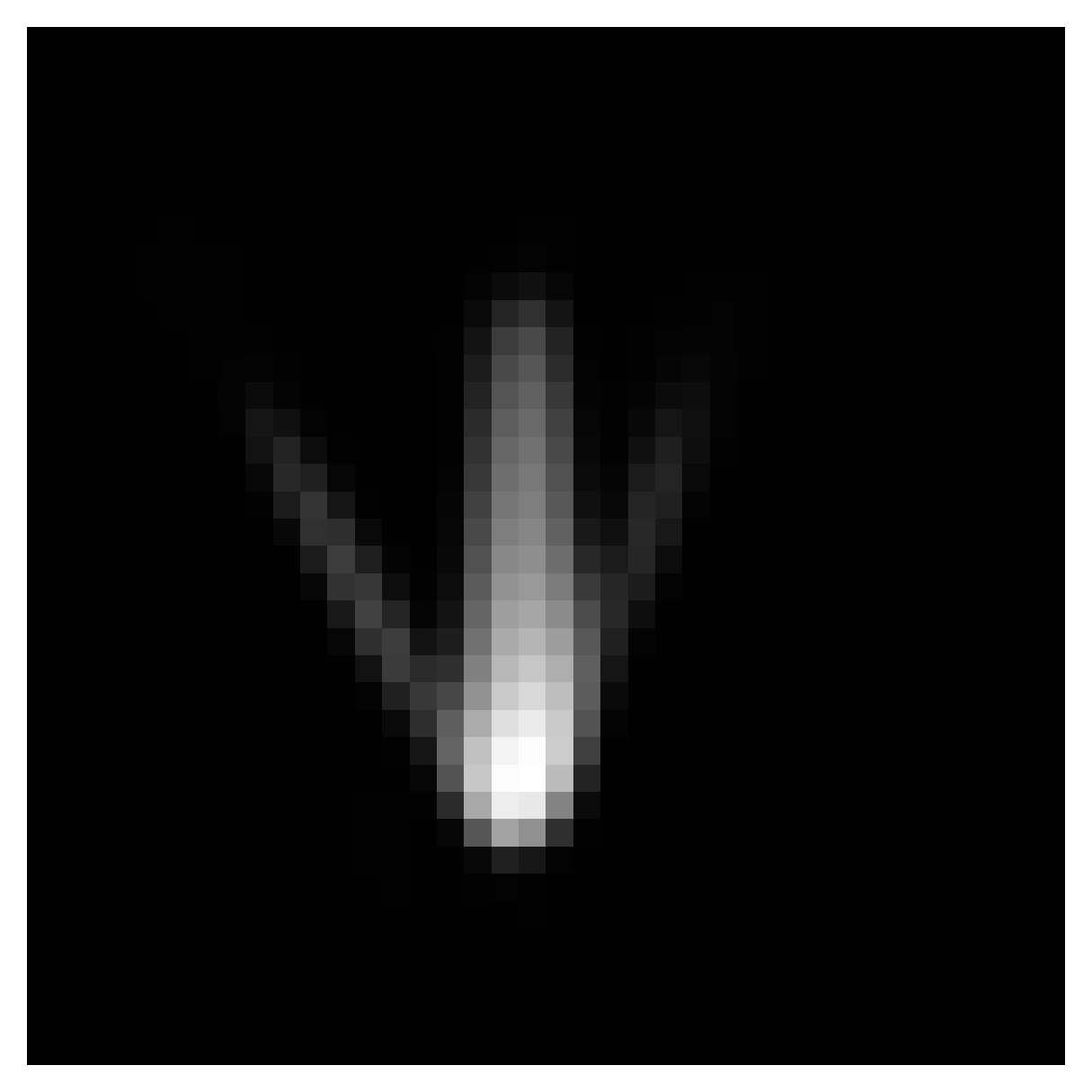} & \includegraphics[width=\size]{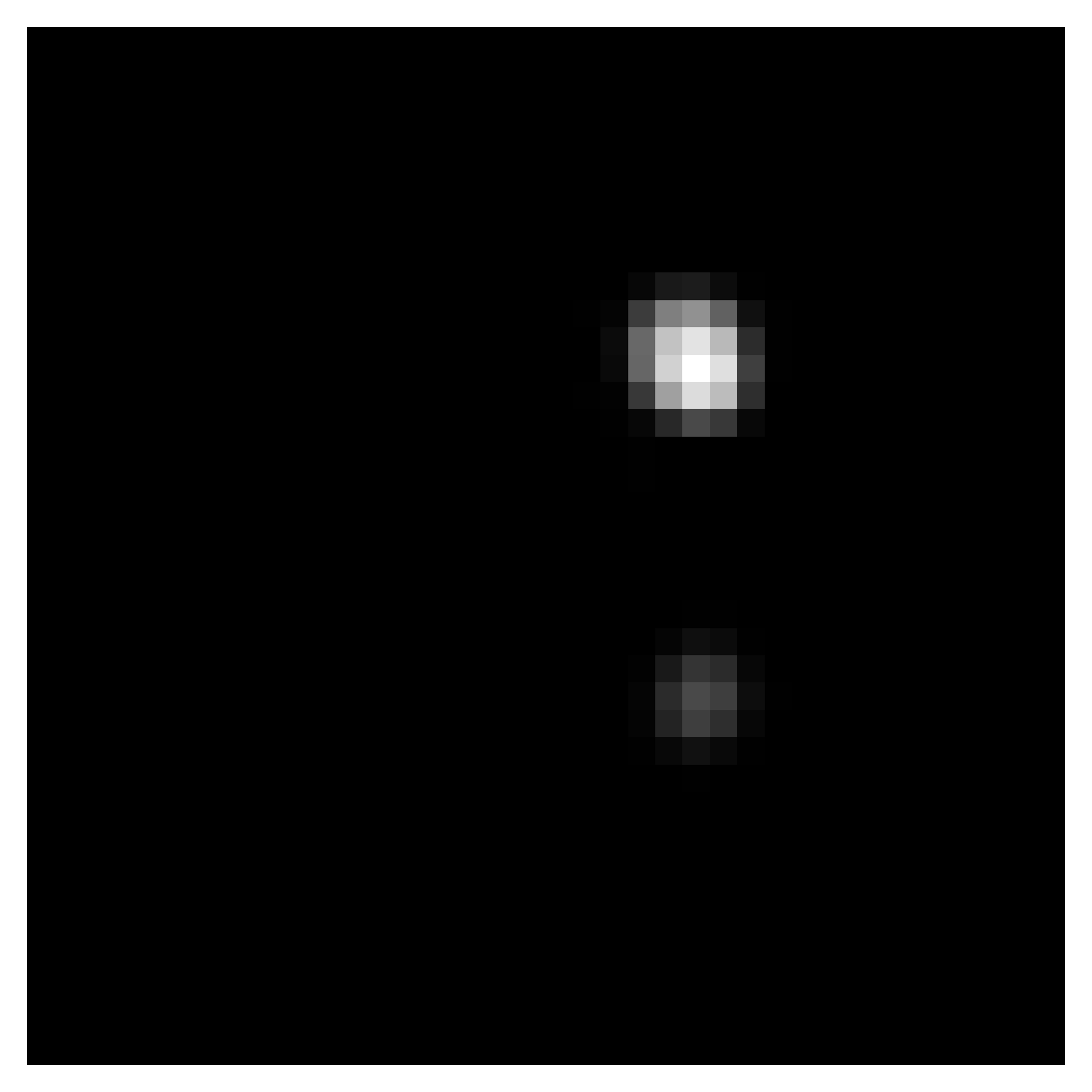} \\
		$\mathrm{Super}_3 (A_{19\times 19} )$ & \includegraphics[width=\size]{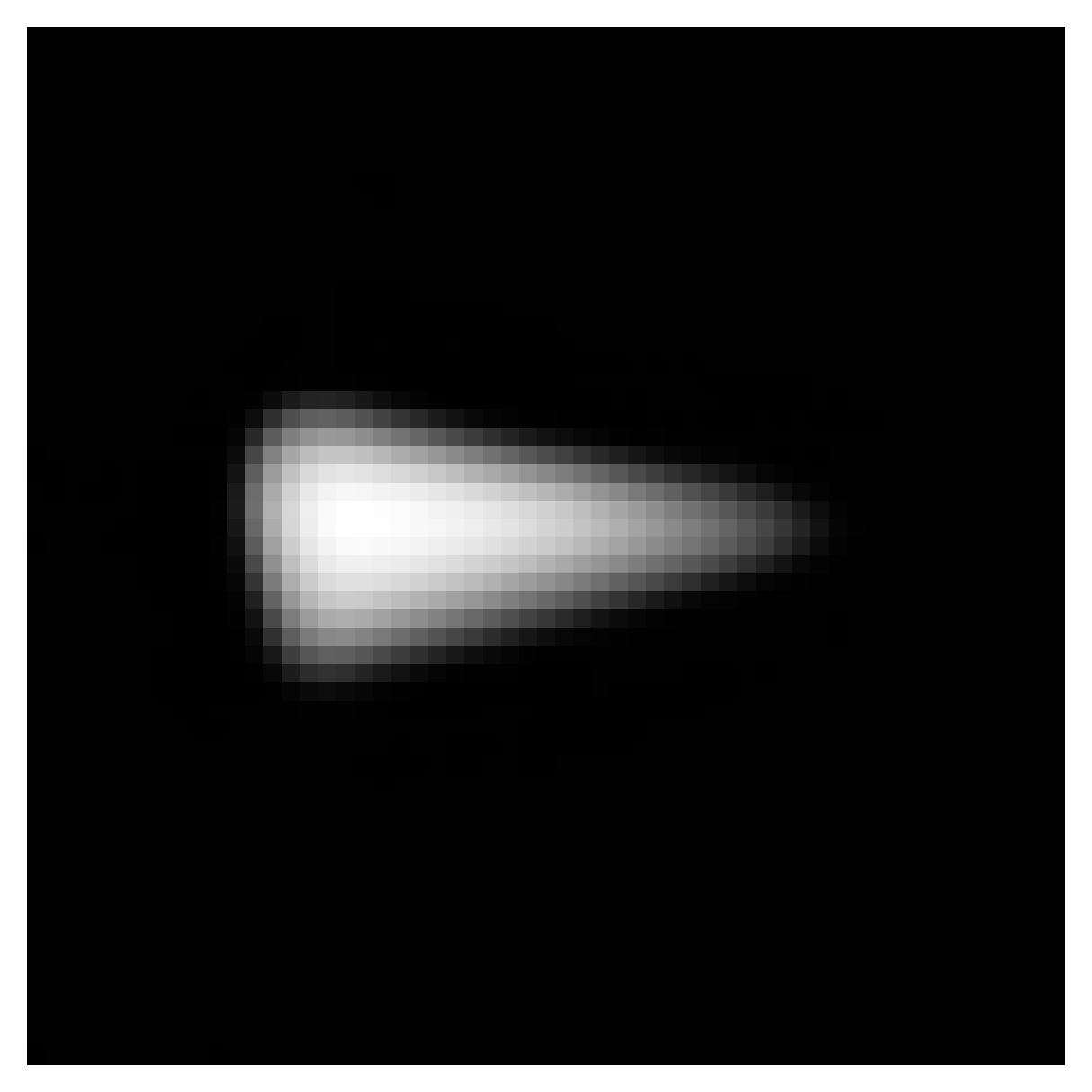} & \includegraphics[width=\size]{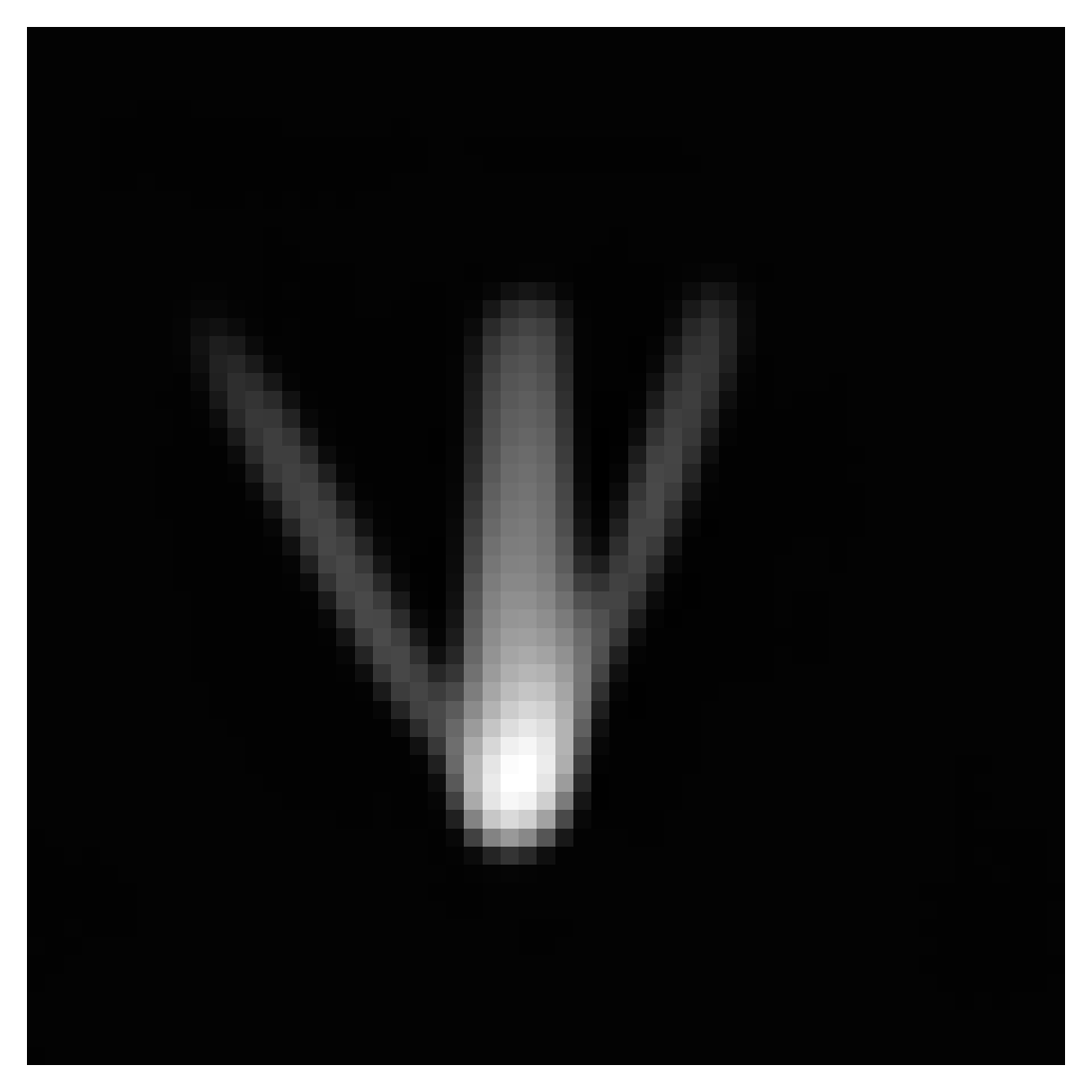} & \includegraphics[width=\size]{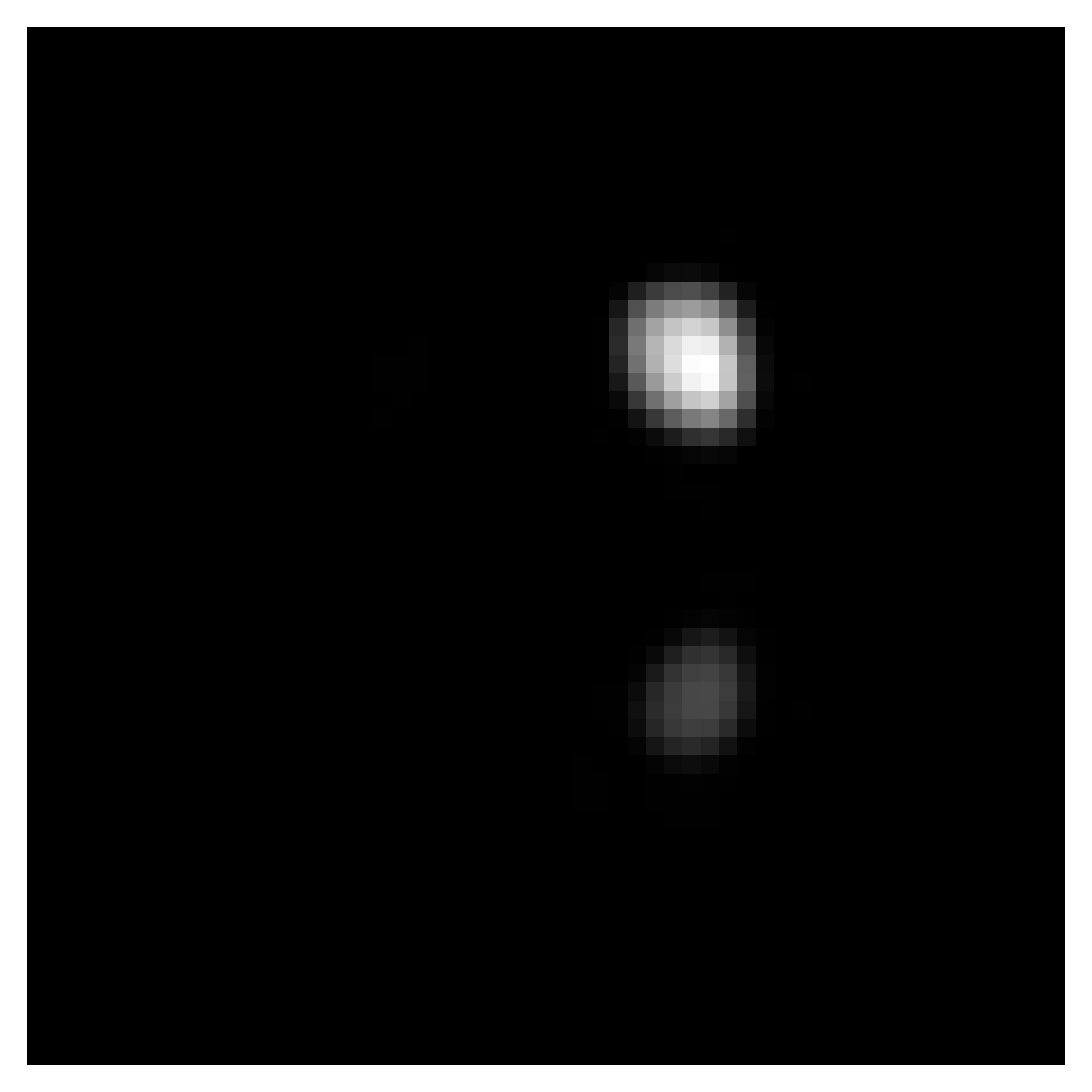} \\
		$\mathrm{Super}_4 (A_{19\times 19})$ & \includegraphics[width=\size]{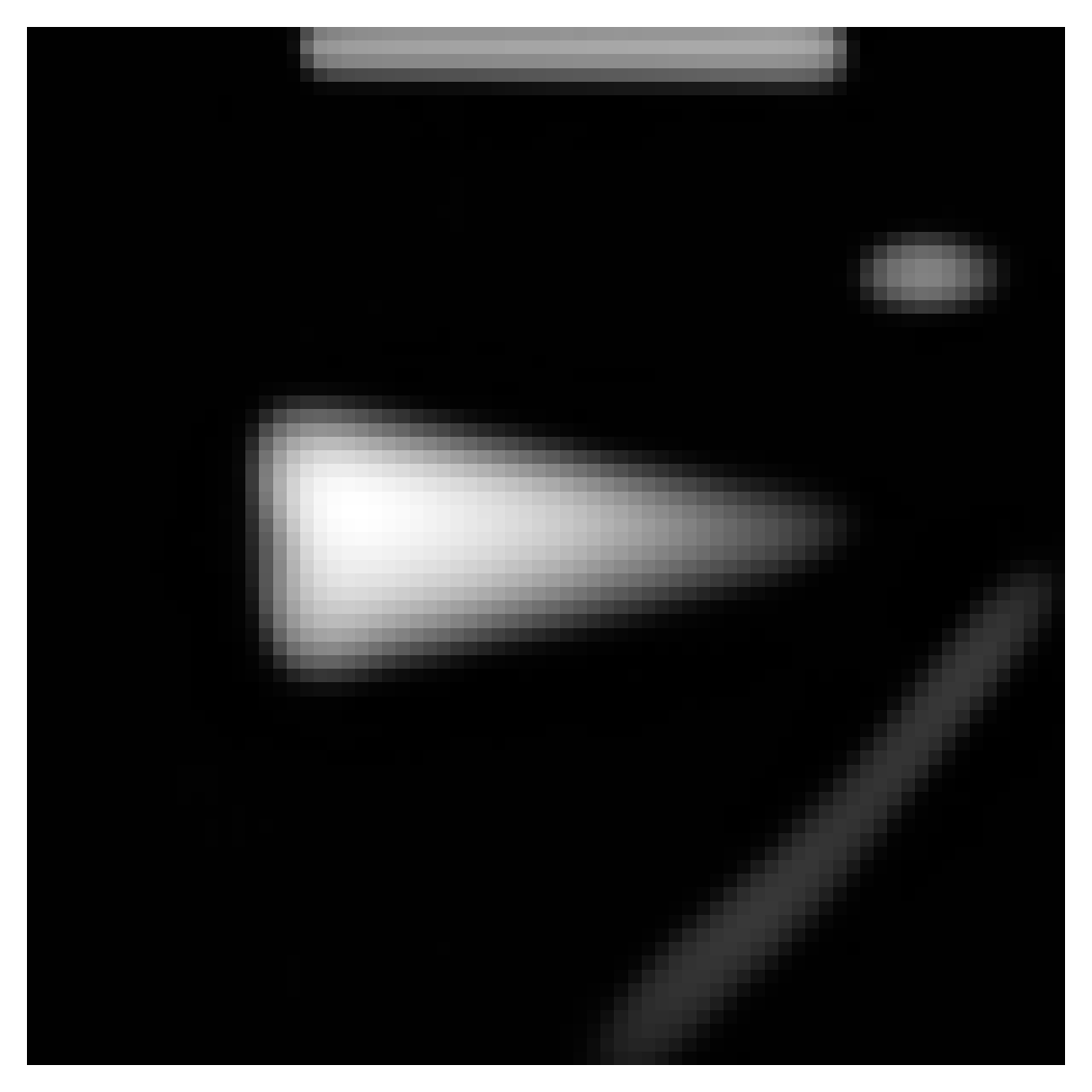} & \includegraphics[width=\size]{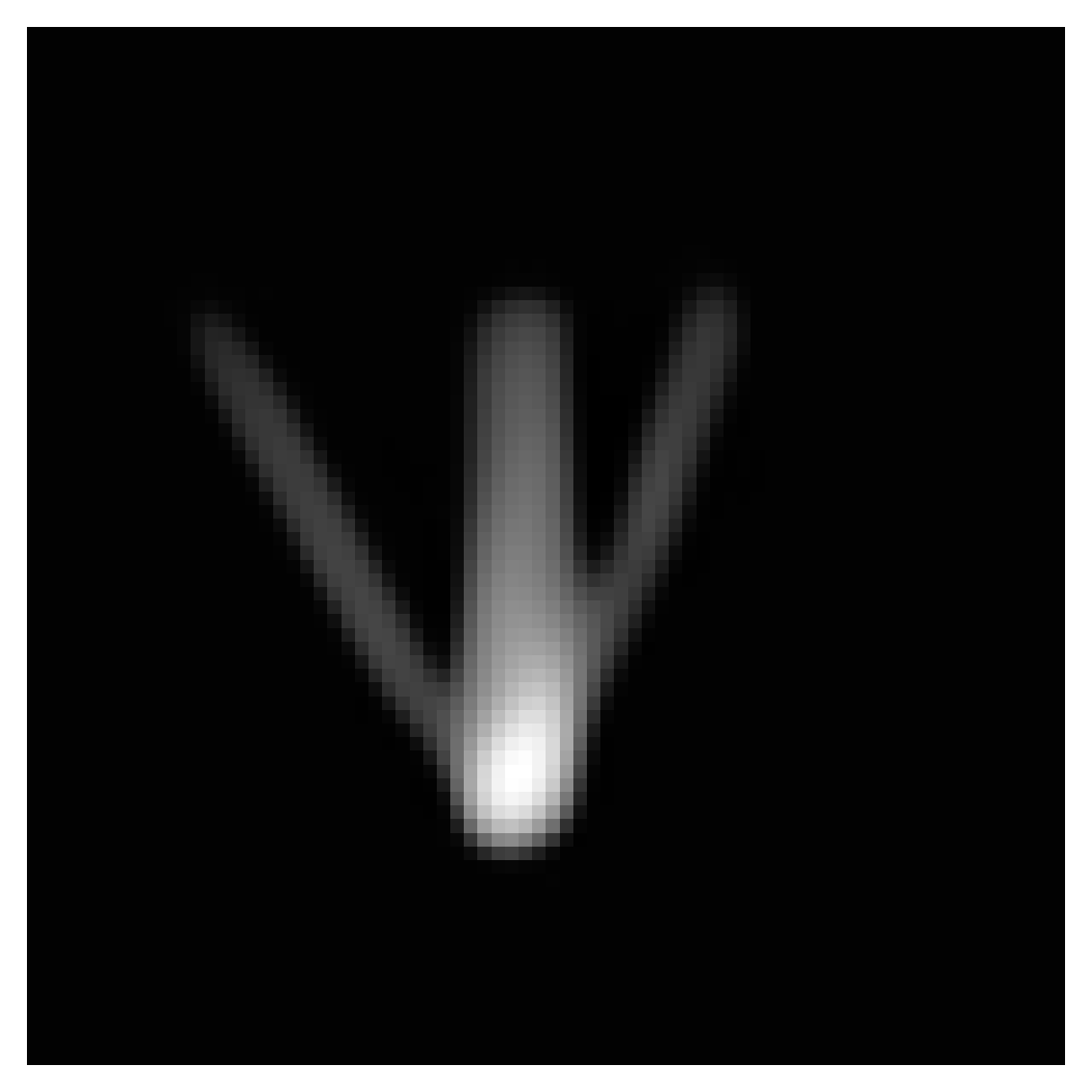} & \includegraphics[width=\size]{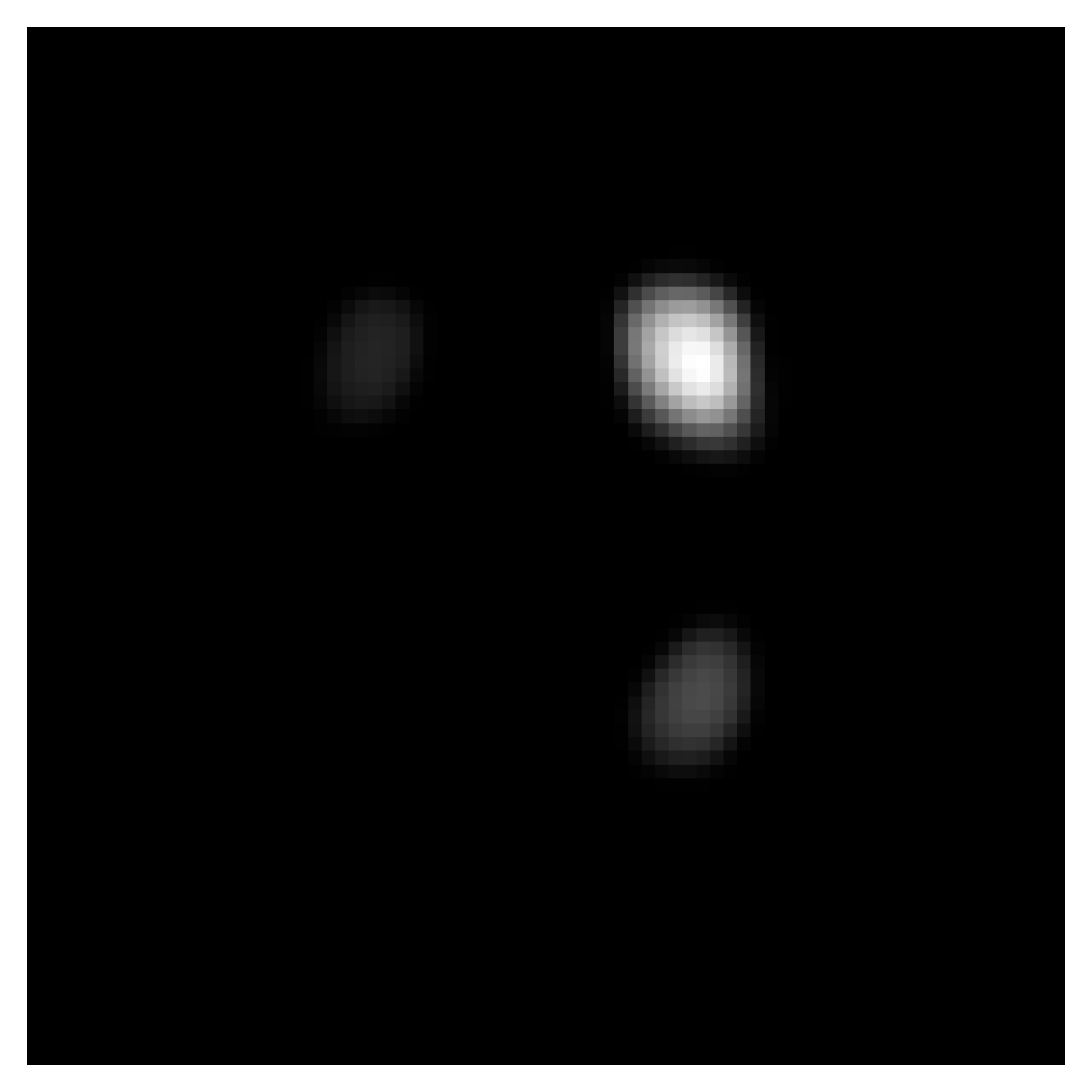} \\
		$\mathrm{Super}_5 (19\times 19 )$ & \includegraphics[width=\size]{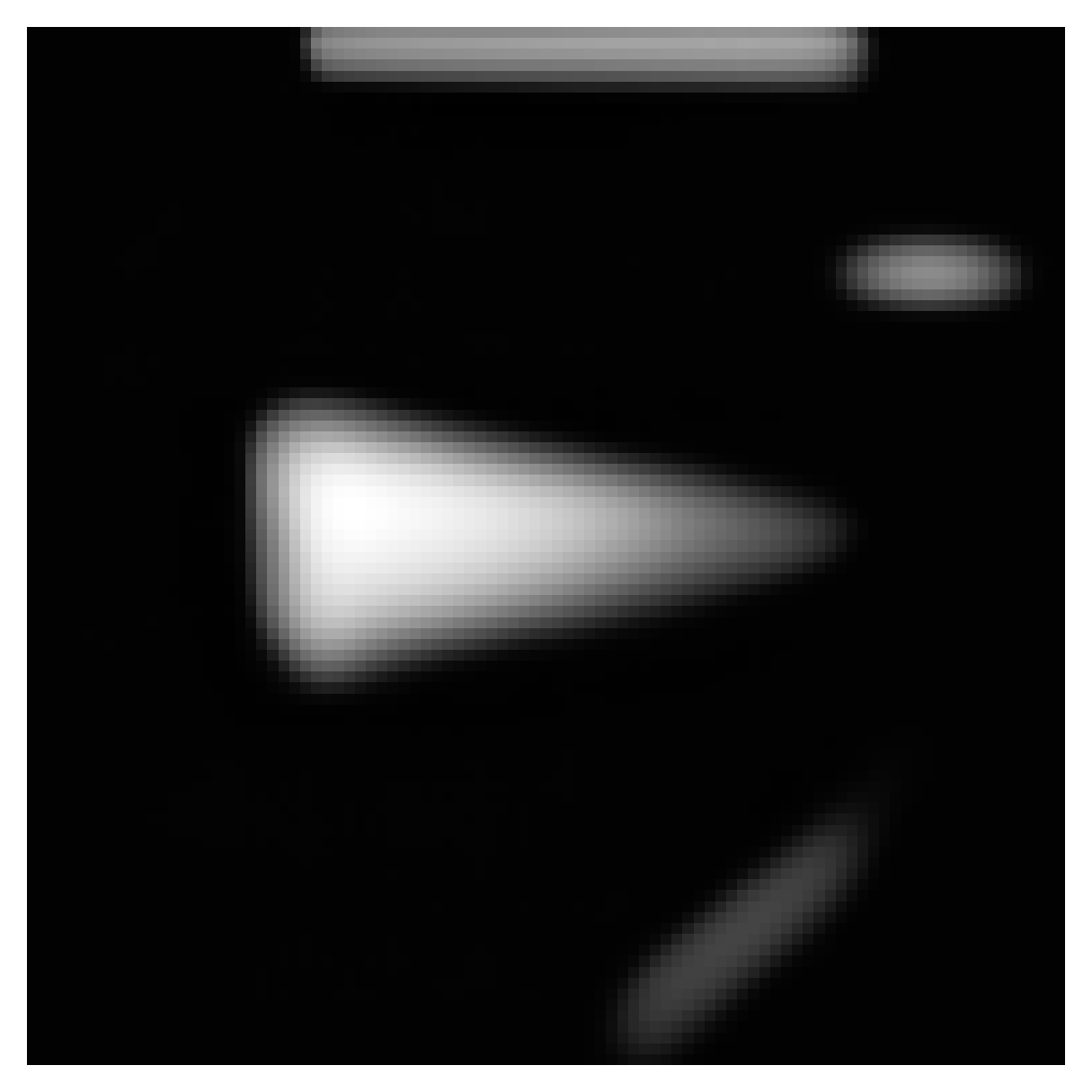} & \includegraphics[width=\size]{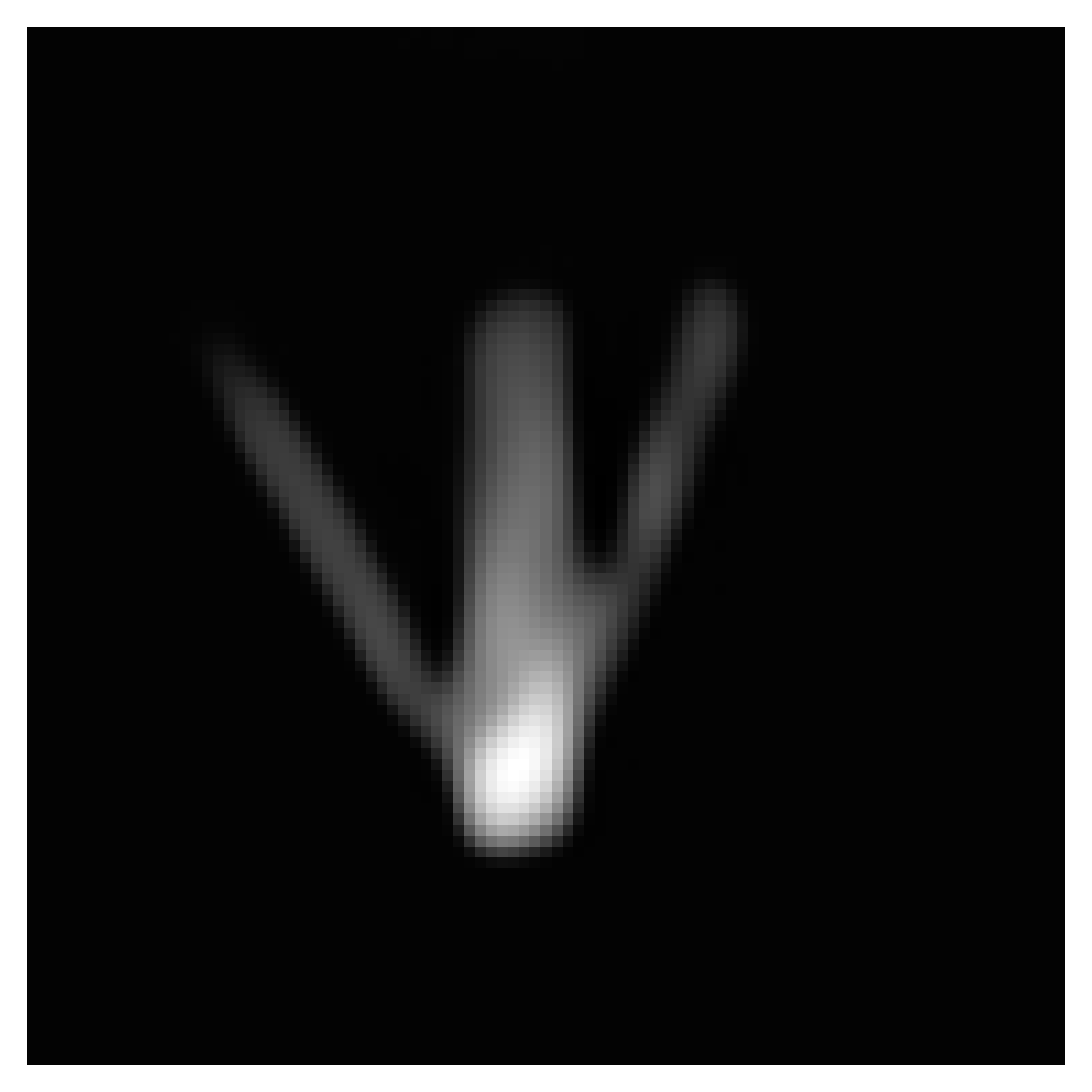} & \includegraphics[width=\size]{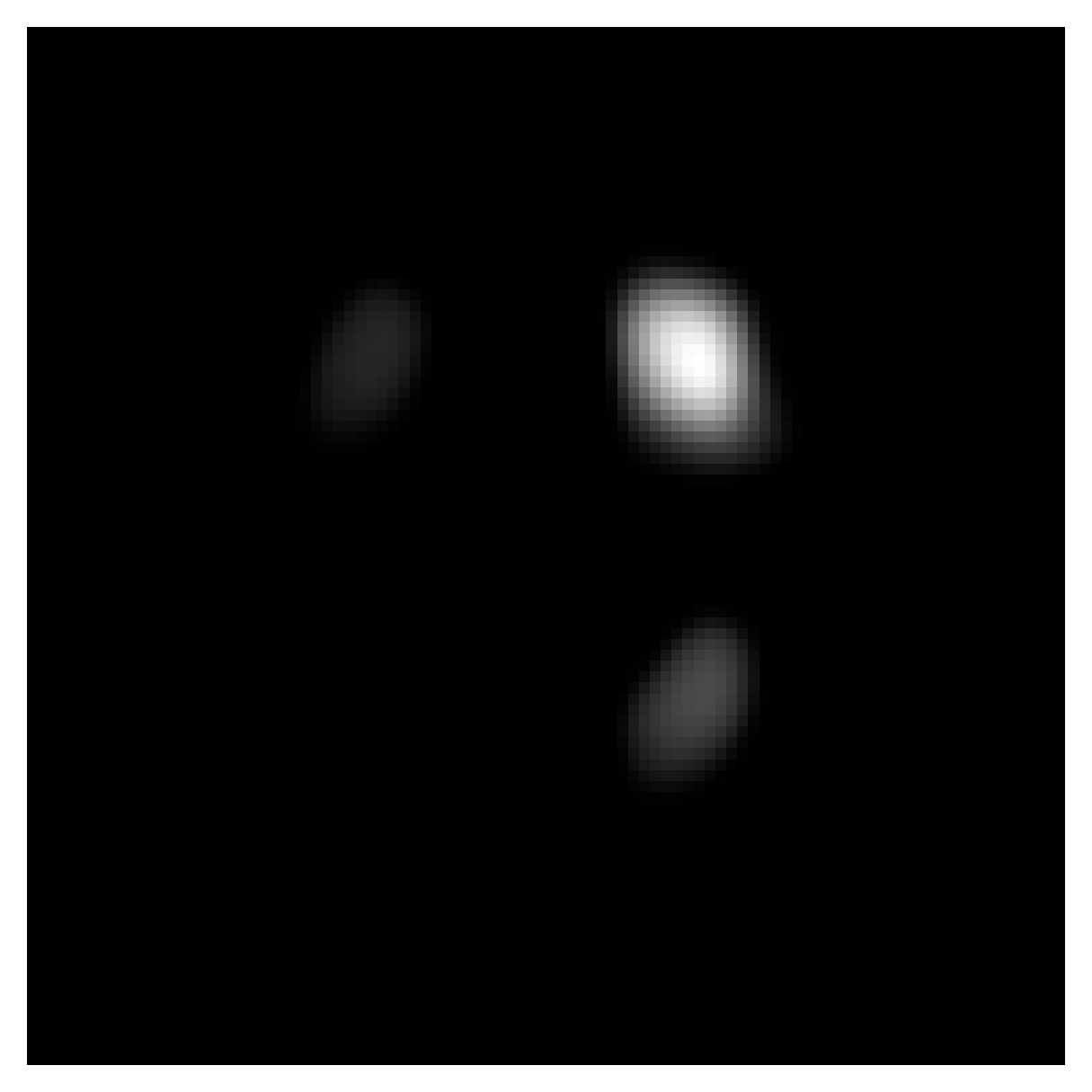} \\
	\end{tabular} 
	\caption{Reconstructions on the 2D OpenMPI dataset. In the top and second row, 
		we have reconstructions performed with ZS-PnP using the two matrices $A_{37\times 37}$ and $A_{19\times 19}$, 
		provided within the dataset. The reconstructions displayed in rows 3-6 are obtained with \our\ 
		and $A_{19\times 19}$ with a super-resolution factor $s$ ranging from 2 to 5. We observe that 
		using \our\ helps reconstruct fine-resolution features of the phantoms starting from the same system matrix $A_{19\times 19}$ (cf. the rows 2-6 with 1 and 2).} 
	\label{tab:reco:openmpi} 
\end{table}

In this section we perform reconstructions on the openly available OpenMPI dataset. 
The OpenMPI dataset contains real data scans of three phantoms: the shape, resolution 
and concentration phantoms. In addition to the the scan data, two different system matrices 
are available: a system matrix $A_{19\times 19}$ calibrated on a $19\times 19$ grid and 
a higher resolution system matrix $A_{37\times 37}$ calibrated on a $37\times 37$ grid. 
The presence of the higher-resolution system matrix $A_{37\times 37}$ is important because 
in the real-data scenario ground truths are not available. We therefore use $A_{37\times 37}$ 
to reconstruct the phantom from the scan data and obtain an approximation of the underlying 
real phantom on a $37\times 37$ grid. These reconstructions are used as a comparison basis 
for the reconstructions obtained with the \our\ algorithm. In fact, in this experiment we 
compare the reconstructions on the $37\times 37$ grid with reconstructions that use 
$A_{19\times 19}$ but produce super-resolved reconstruction using \our . We produce 
reconstruction results for scaling factors $s = 2 ,3,4,5$.
Concerning the preprocessing steps described in section \ref{subsec:data:preproc}, 
all reconstructions have been performed discarding the frequencies below the $80\ \si{\kilo\hertz}$, 
as they are notoriously unreliable \cite{maass2024equilibriumanysotropy}. Additionally, we use the 
SNR estimation available with the dataset and cut all frequencies whose SNR is below 1 for the 
reconstructions with $A_{37\times 37}$. No SNR thresholding has been applied to the 
reconstructions that use $A_{19\times 19}$. Finally, a low rank approximation is performed 
using rSVD and a target rank K as described in section \ref{subsec:data:preproc}. This rank 
has been set to $K=\frac{N_x \cdot N_y}{2}$ when the system matrix $A_{N_x \times N_y}$ is employed. 
For all reconstructions, the number of iterations has been set to $n_{\mathrm{it}}=10$. The 
reconstruction parameter $\mu_0$ has been set to $10^{7}$ for all reconstructions with 
$A_{37\times 37}$. For the reconstructions with $A_{19\times 19}$ we have set $\mu_0$ to 
be $10^{8}$ for the shape and the resolution phantoms, and $10^{10}$ for the concentration phantom. 
These parameters have been selected by visual inspection of the final results. A certain degree of 
dependency of $\mu_0$ on the specific phantom when working with real data has also been observed in
\cite{gapyak2025ell1pnp}. The final reconstruction are displayed in table~\ref{tab:reco:openmpi}. 
As mentioned, we use the reconstructions obtained with $A_{37\times 37}$ as a comparison 
reconstruction in which the real features of the underlying phantoms are reconstructed and 
visible (top row). We observe that, when using $A_{19\times 19}$, finely-resolved features of 
the phantoms are not being reconstructed (e.g. the side branches of the resolution phantom or 
the missing diagonal dots in the concentration phantom in table~\ref{tab:reco:openmpi}). 
However, when using \our\ with $s \geq 2$ we observe that the features of the phantoms that 
were missing in the $19\times 19$ reconstruction are now visible. We remark once more that 
both the reconstruction on the $19\times 19$ grid and the reconstructions using \our\ 
algorithm both use $A_{19\times 19}$, i.e., the reconstructed feature with \our\ are due 
to \our\ itself and not due to an increase in the amount of input information. These results 
underpin the usefulness of \our\ to reconstruct super-resolved images starting from lower-resolution 
system matrices in real MPI scenarios. We observe that although increasing $s$ from 2 to 4 can 
help reconstructing the features of the concentration phantom, it can also introduce reconstruction 
artifacts as in the case of the shape phantom.

\subsection{Reconstructions on the EMWA Dataset}\label{subsec:exp:emwa}

\def\size{6em}
\newcolumntype{C}{>{\centering\arraybackslash}m{6em}}
\newcolumntype{L}{>{\raggedright\arraybackslash}m{8em}}  
\begin{table}[!thb]
	\centering
	\setlength{\tabcolsep}{0pt}
	\begin{tabular}{L*6{C}@{}}
		\toprule
		GT  & \includegraphics[width=\size]{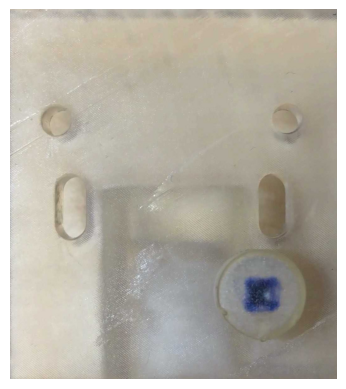} & \includegraphics[width=\size]{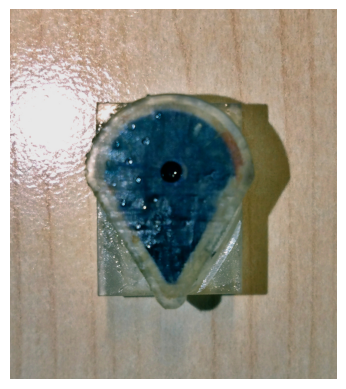} & \includegraphics[width=\size]{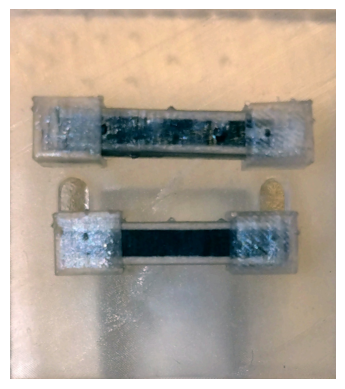} & \includegraphics[width=\size]{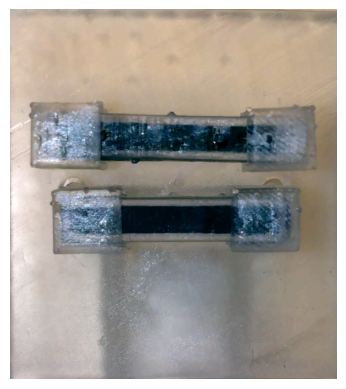} &
		\includegraphics[width=\size]{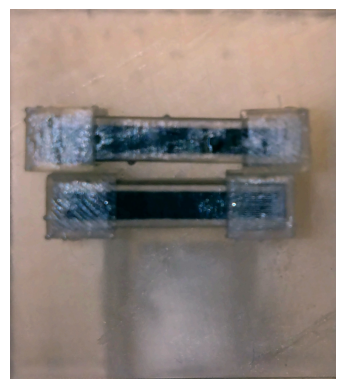} & \includegraphics[width=\size]{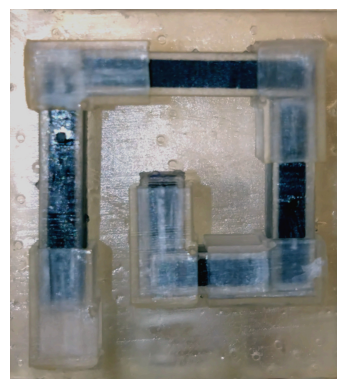}\\
		ZS-PnP($A_{17\times 15})$ & \includegraphics[width=\size]{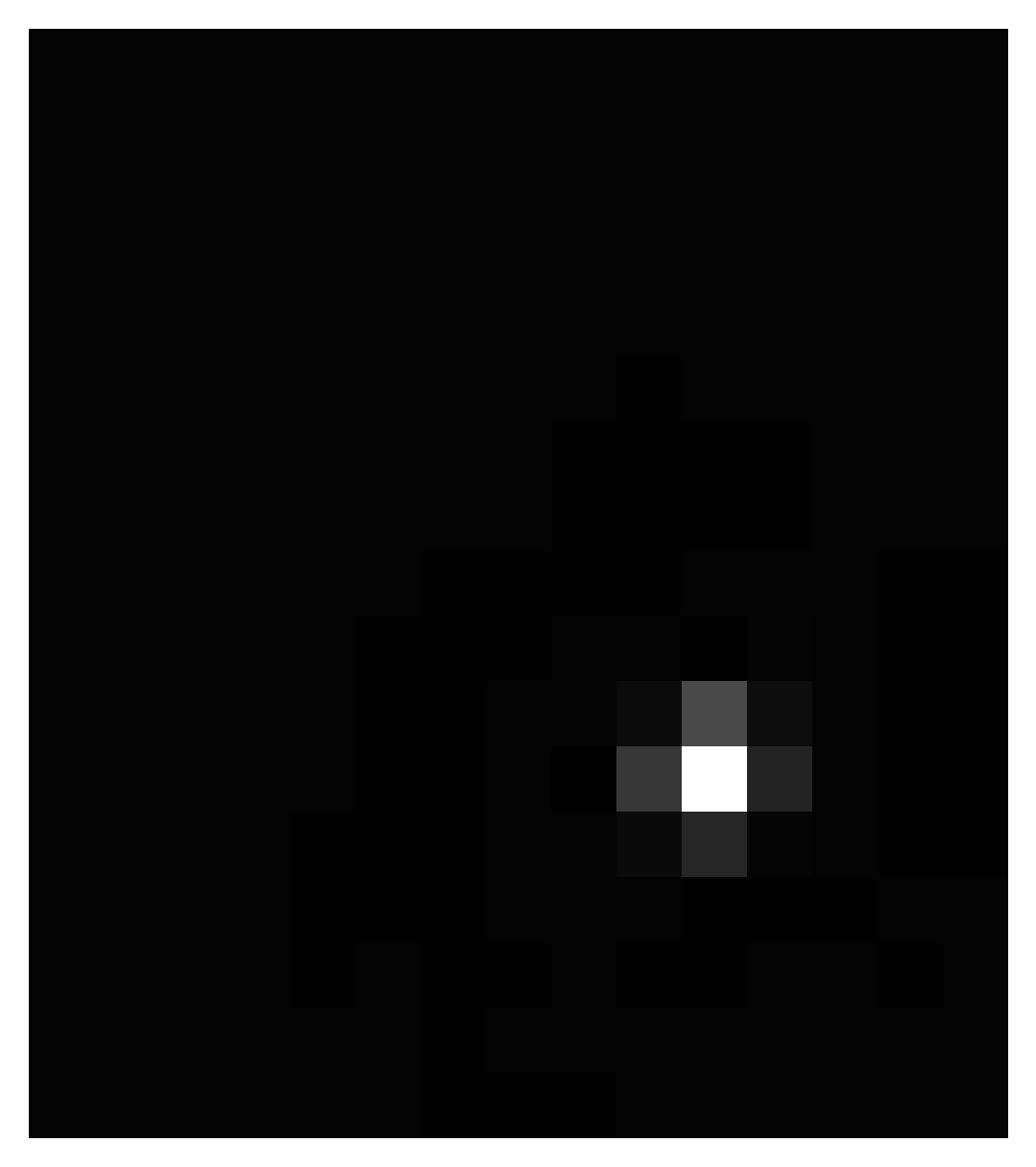} & \includegraphics[width=\size]{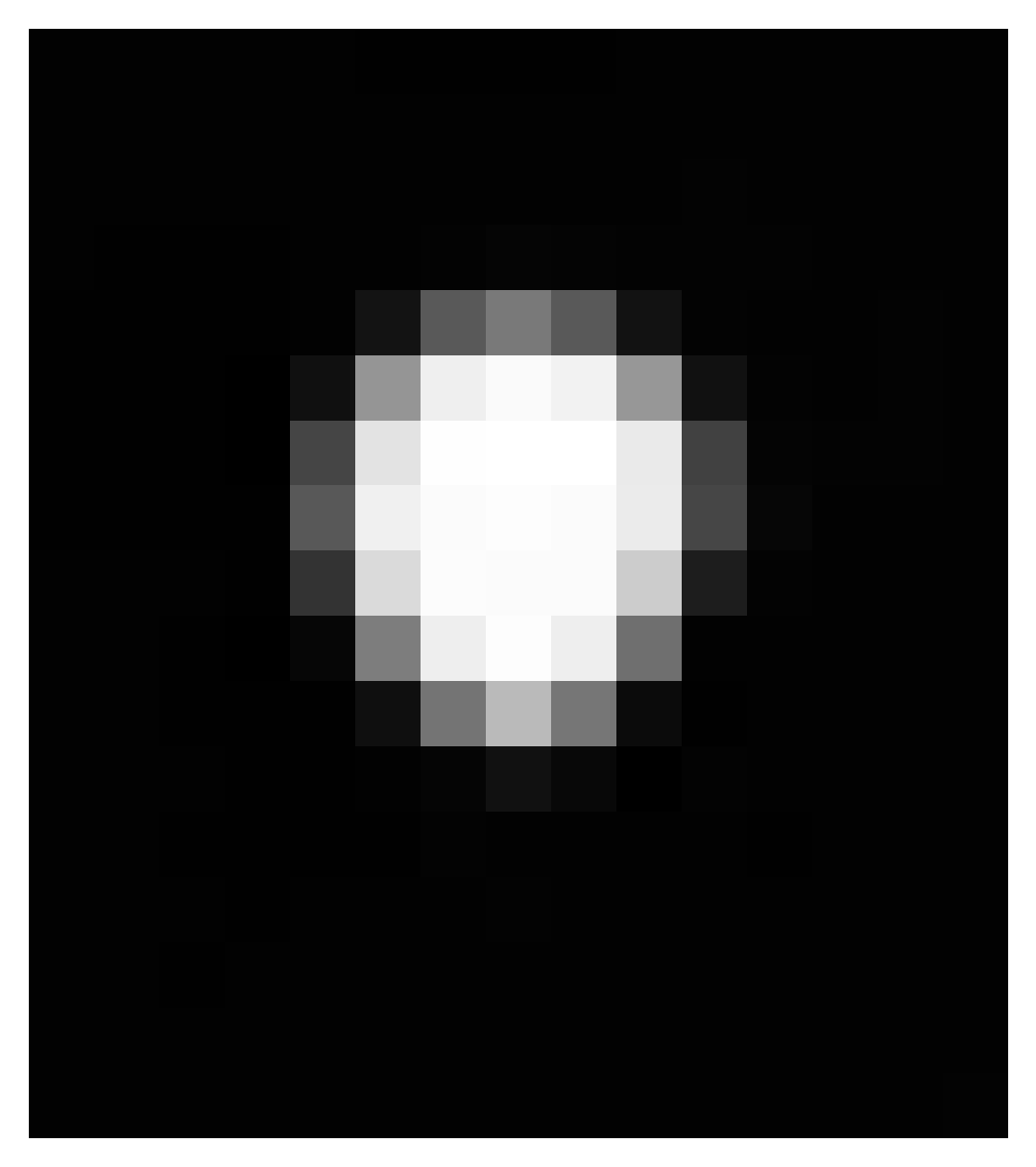} & \includegraphics[width=\size]{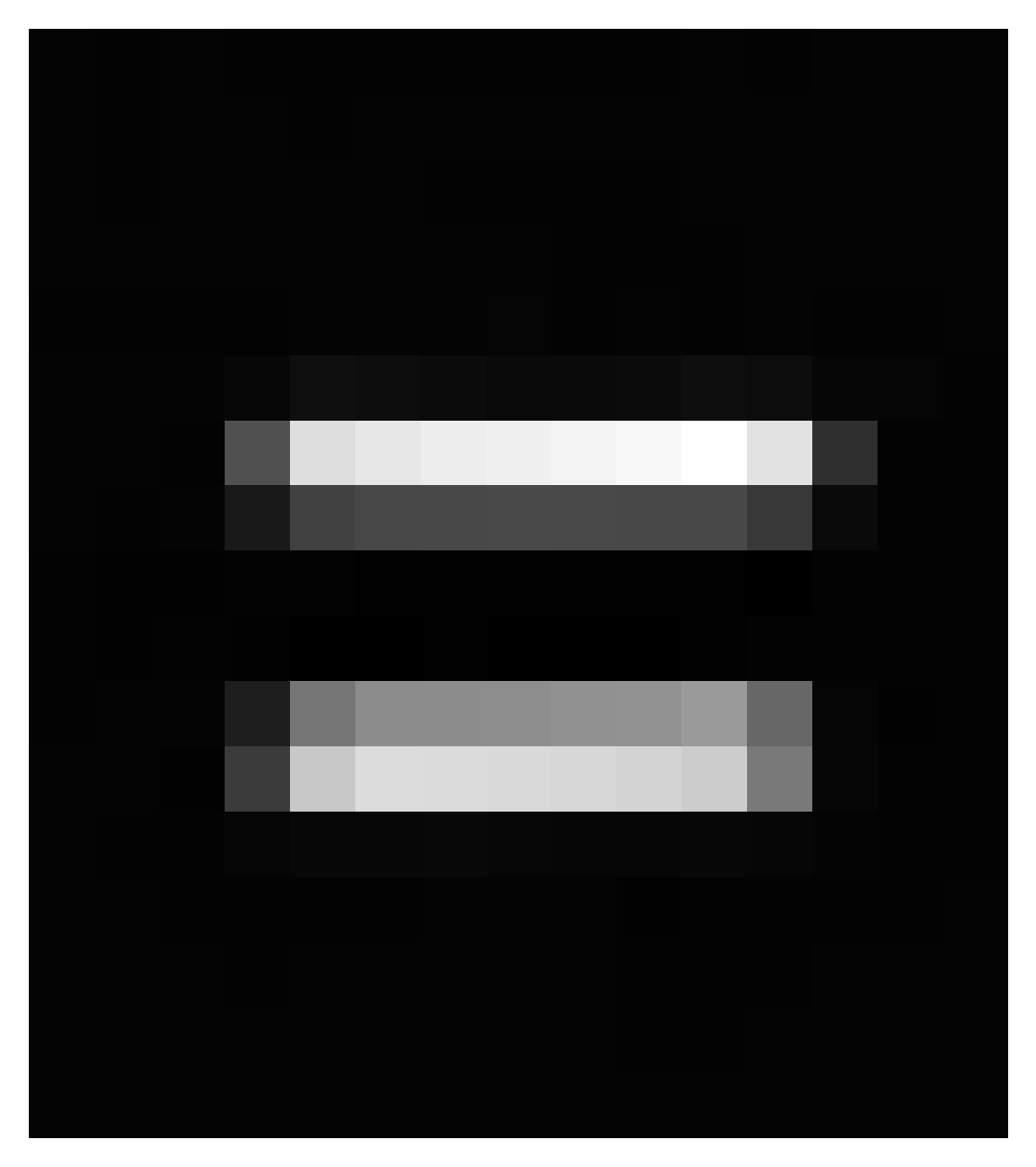} & \includegraphics[width=\size]{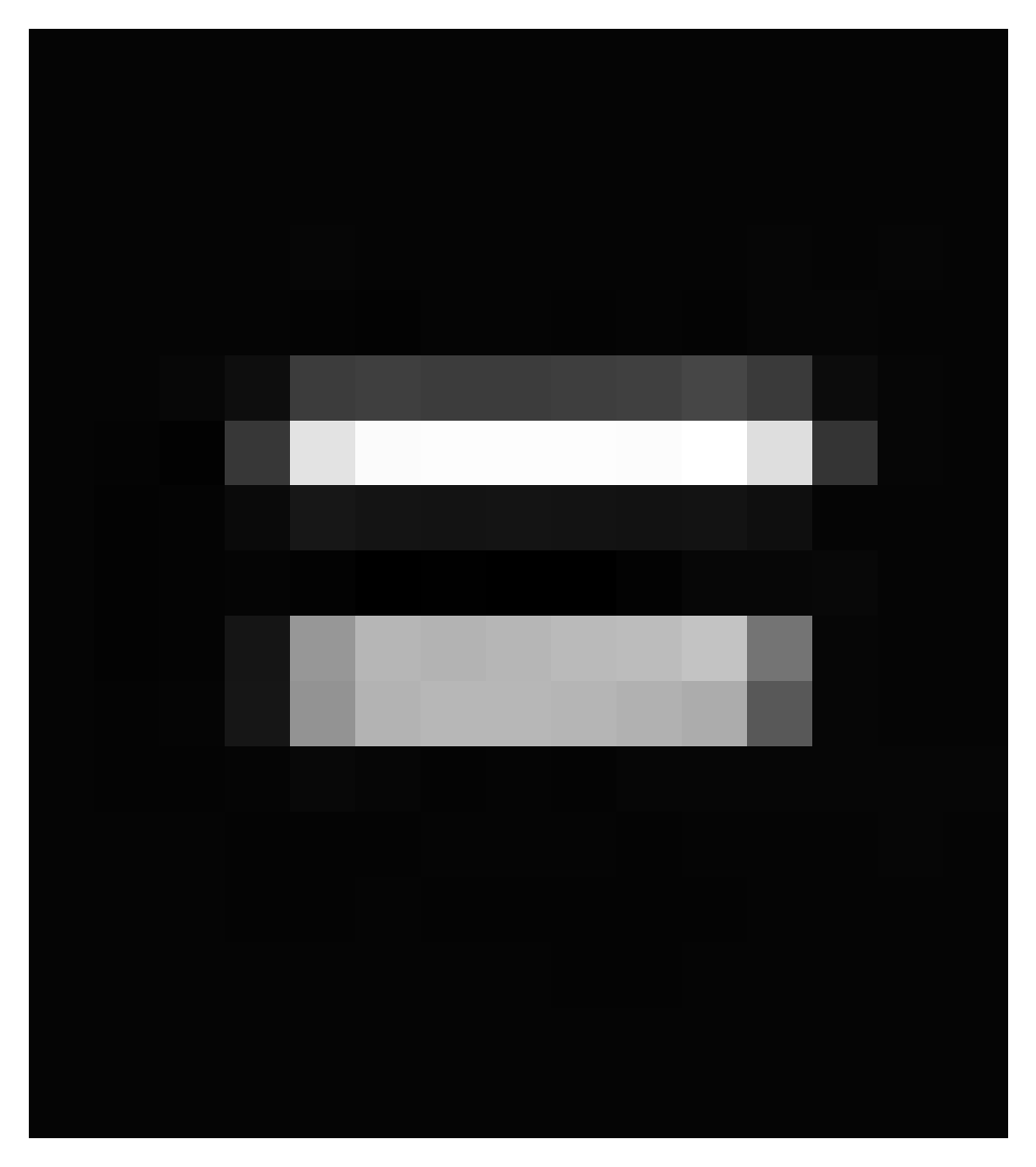} &
		\includegraphics[width=\size]{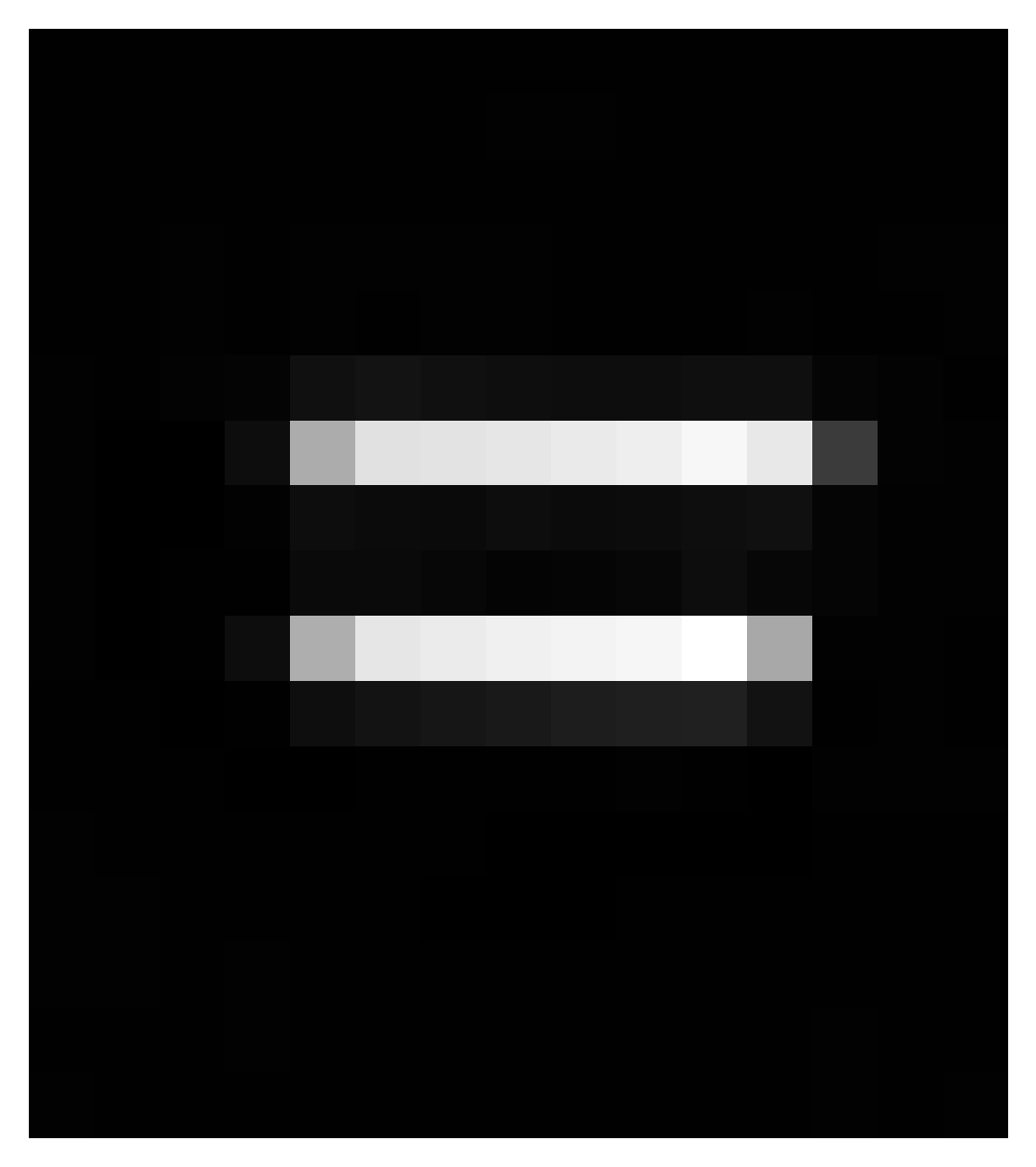} & \includegraphics[width=\size]{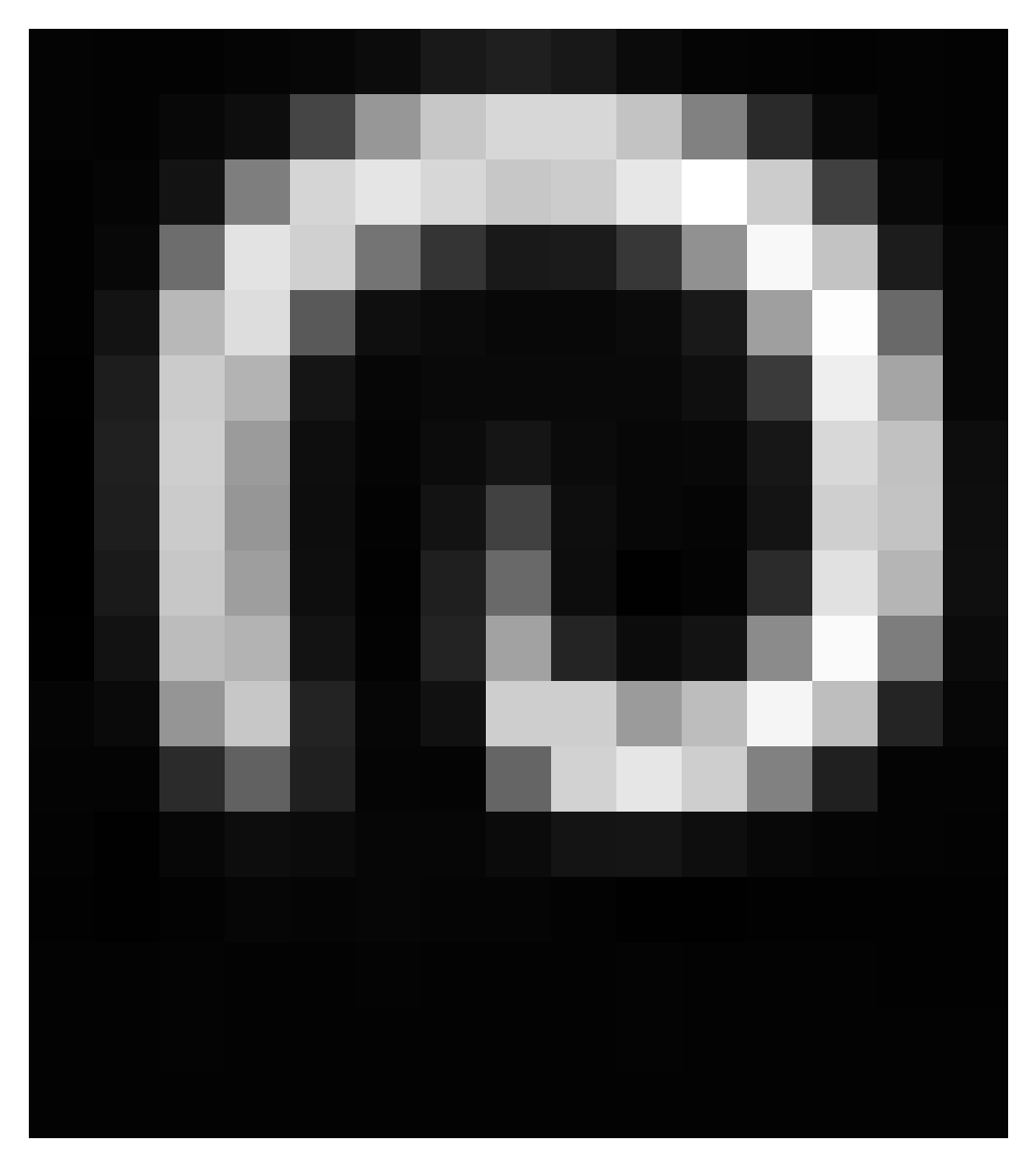}\\
		$\mathrm{Super}_2 (A_{17\times 15} )$ & \includegraphics[width=\size]{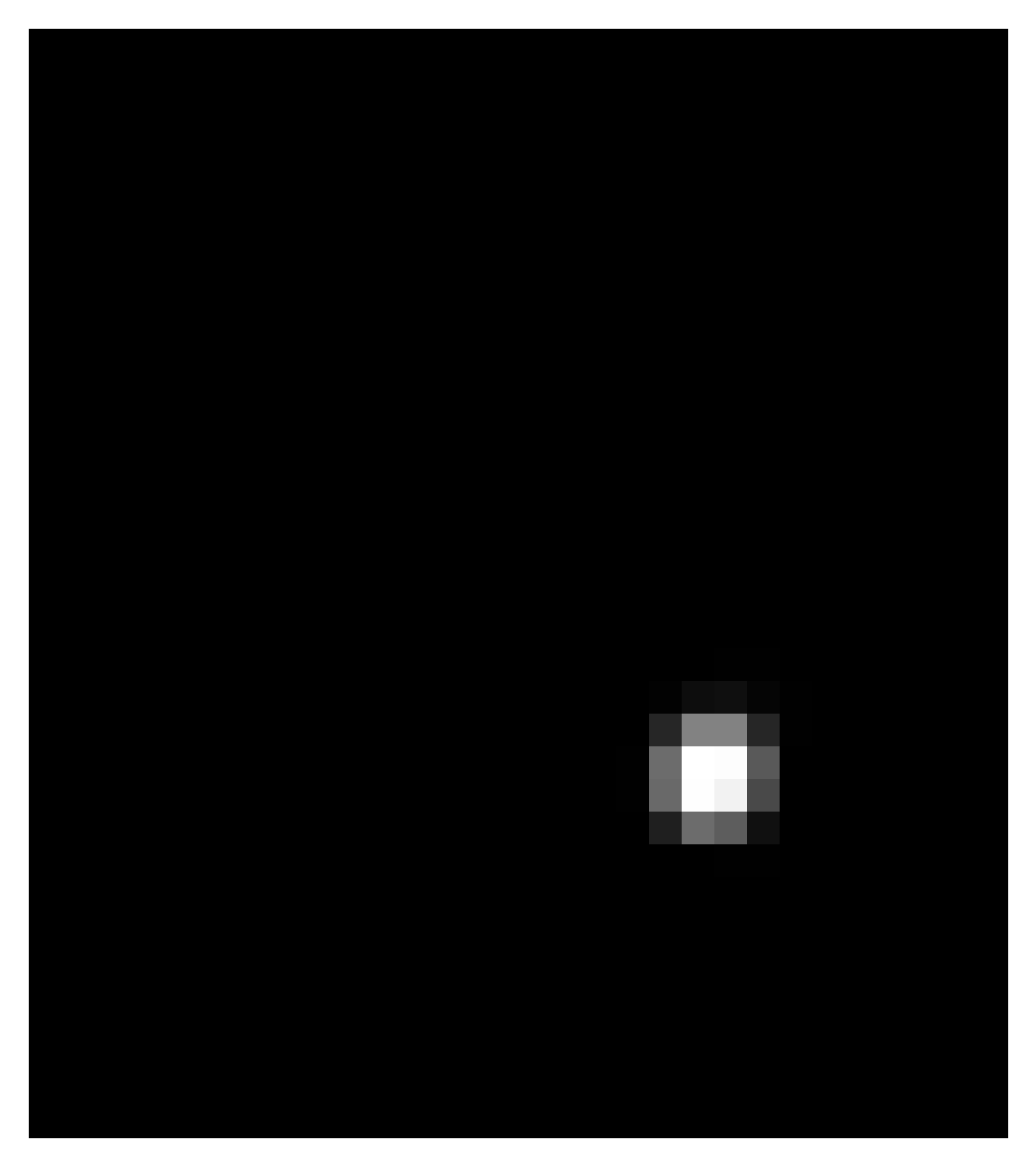} & \includegraphics[width=\size]{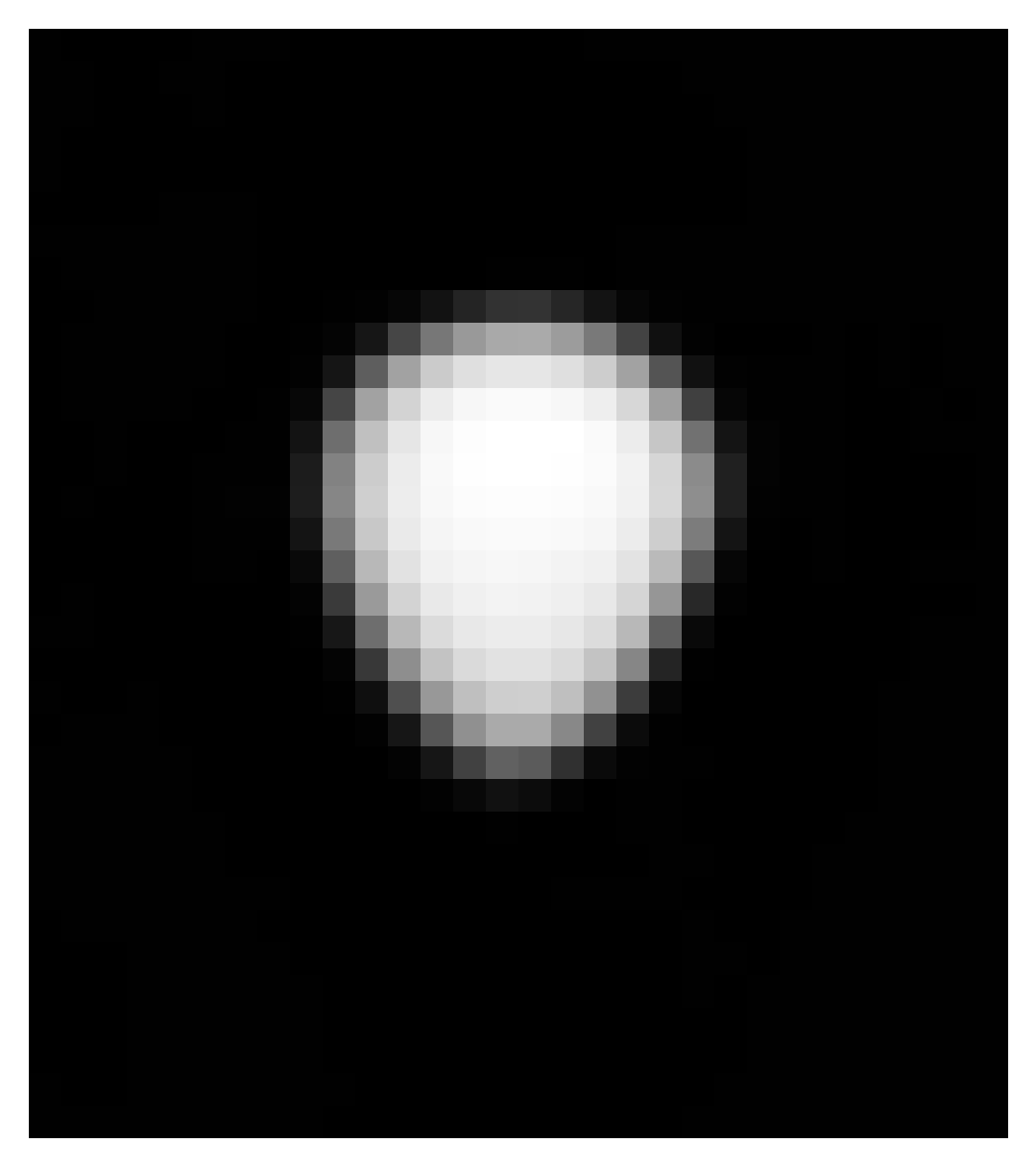} & \includegraphics[width=\size]{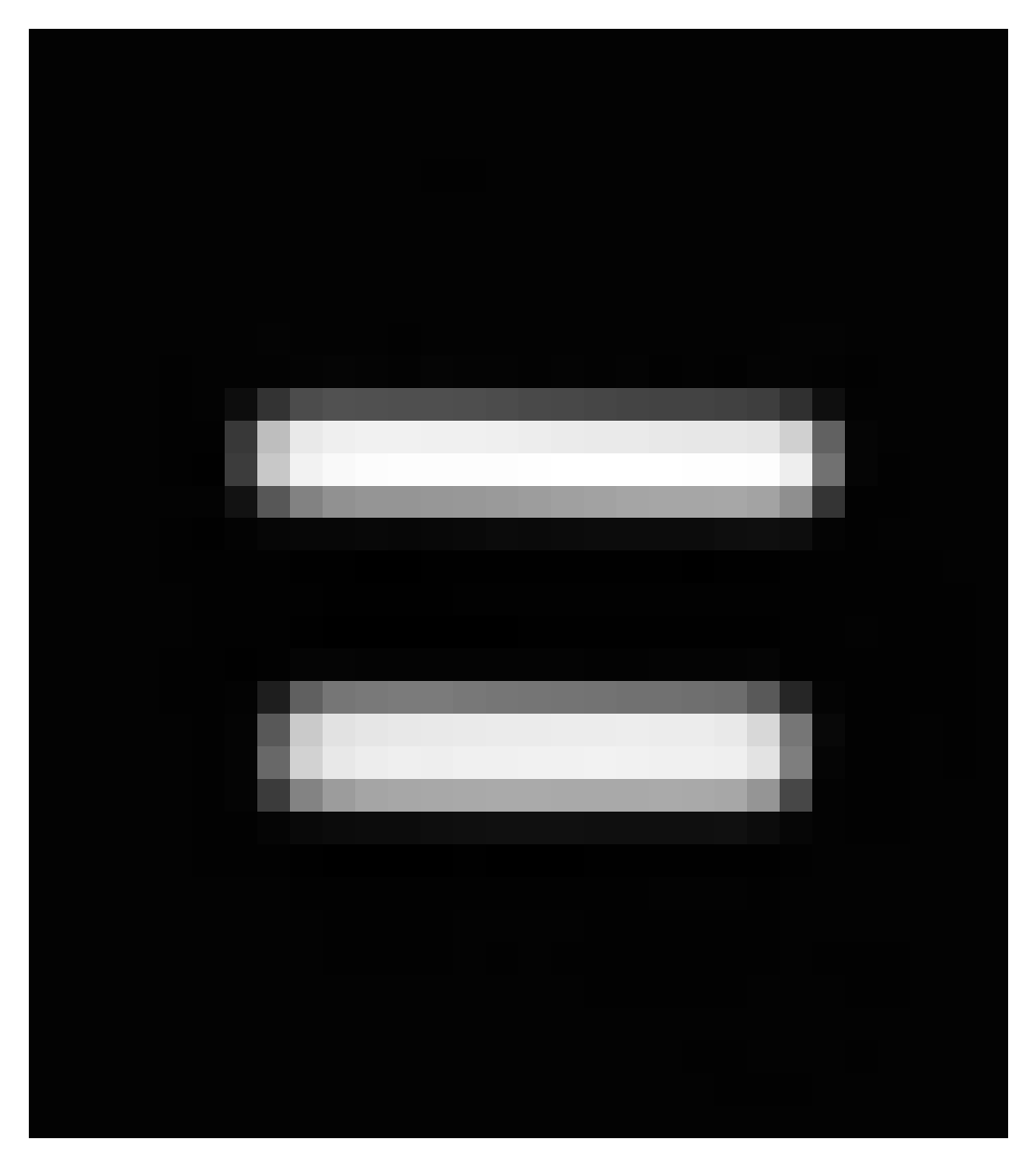} & \includegraphics[width=\size]{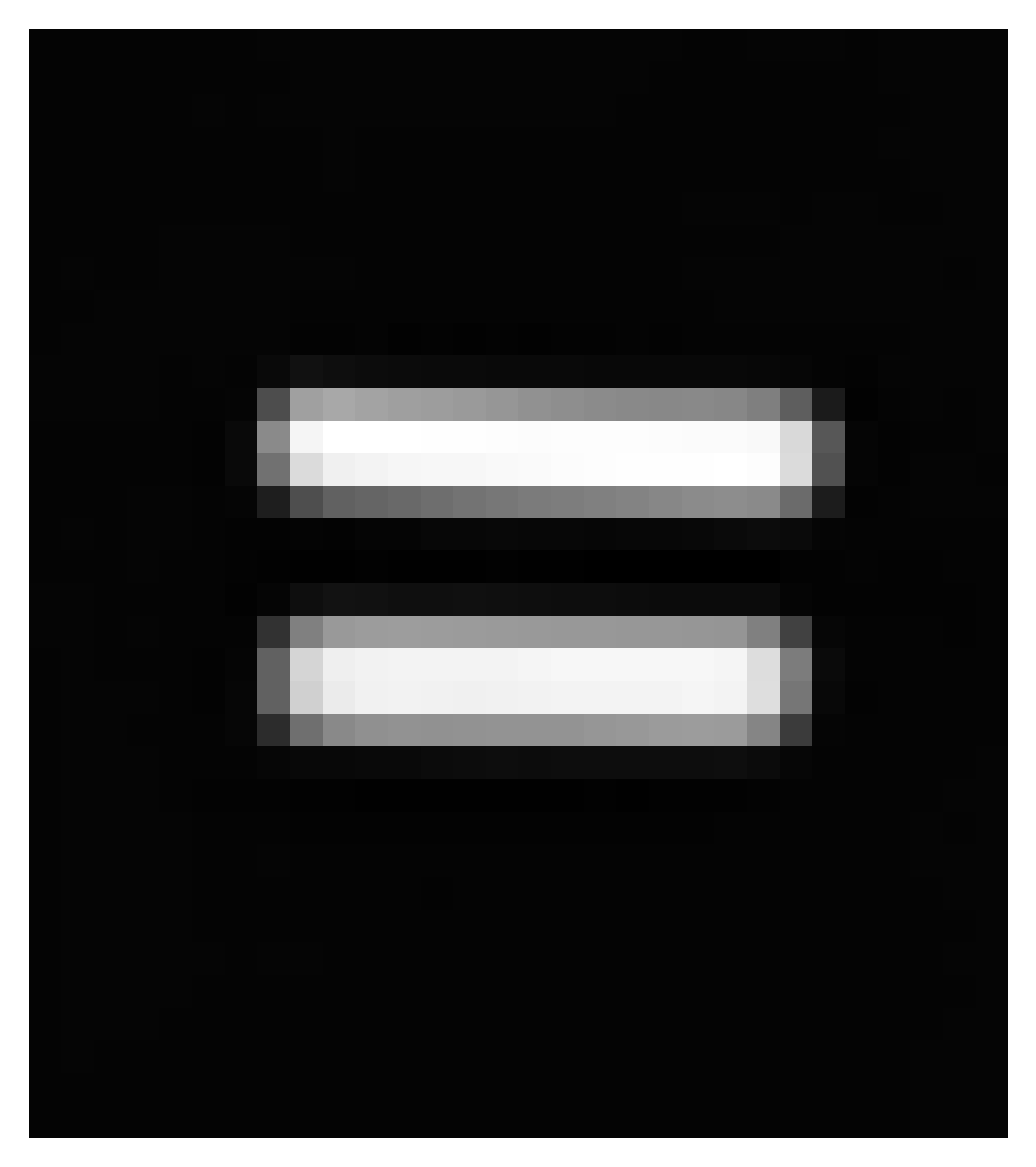} &
		\includegraphics[width=\size]{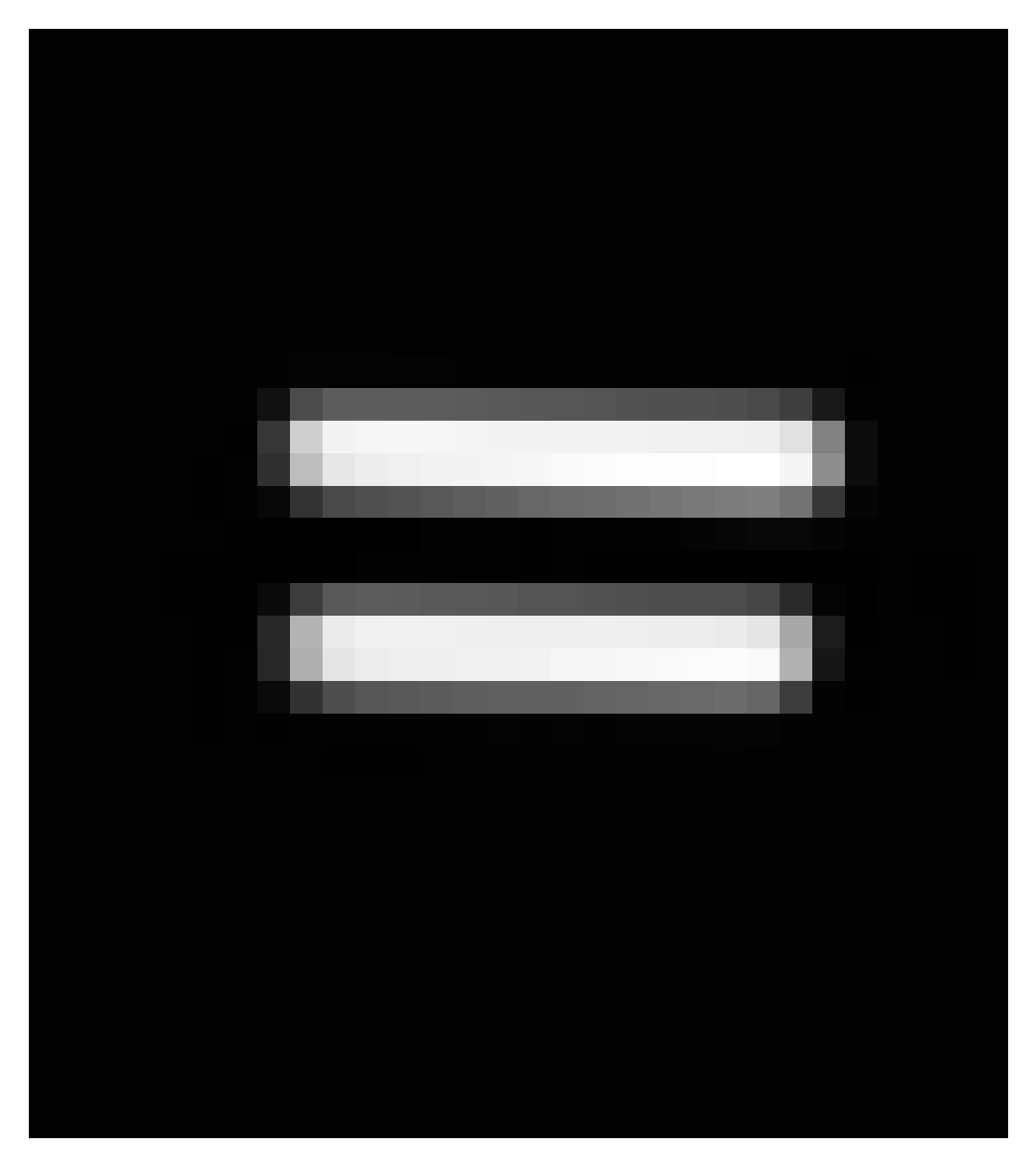} & \includegraphics[width=\size]{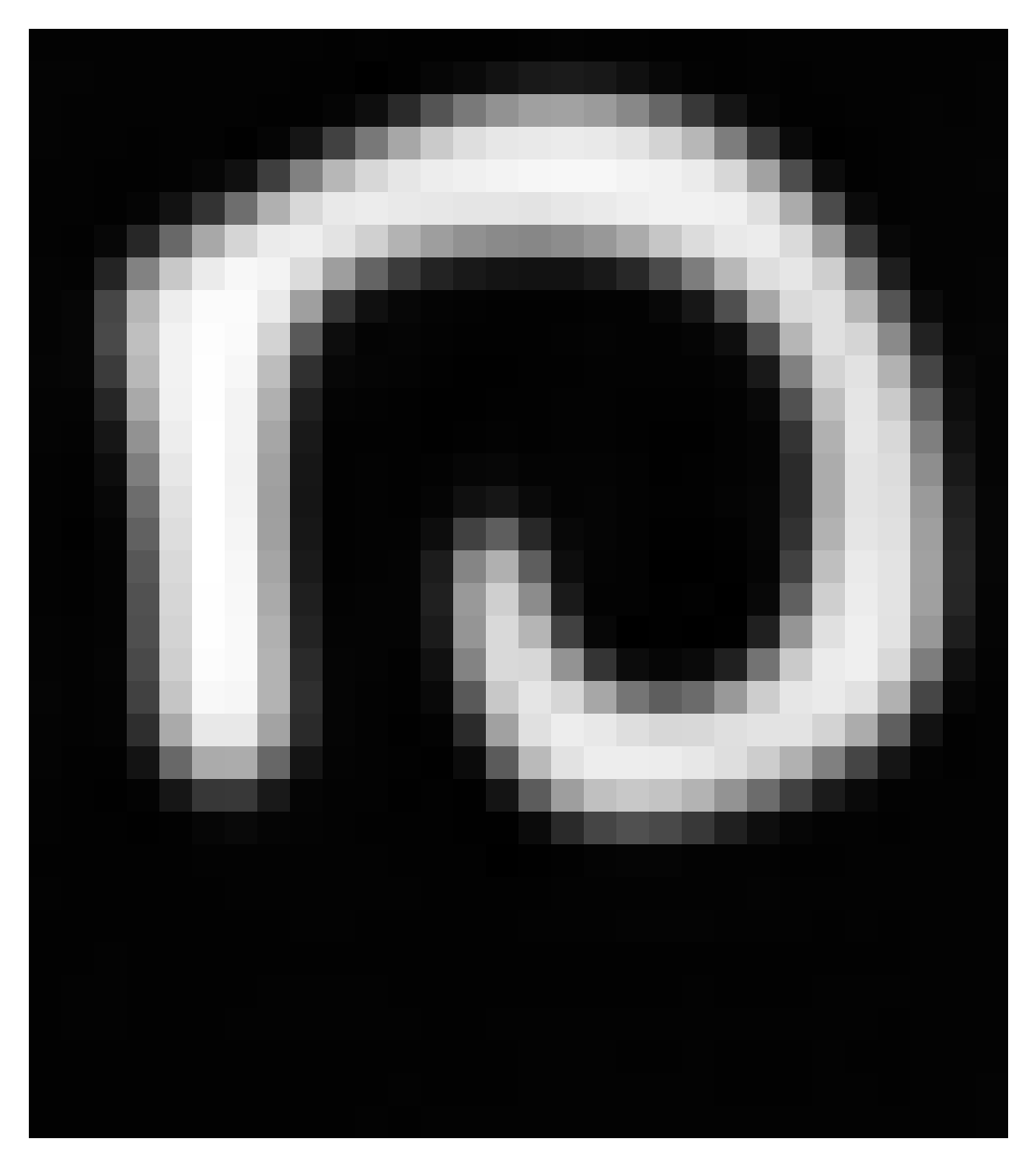}\\
		$\mathrm{Super}_3 (A_{17\times 15} )$ & \includegraphics[width=\size]{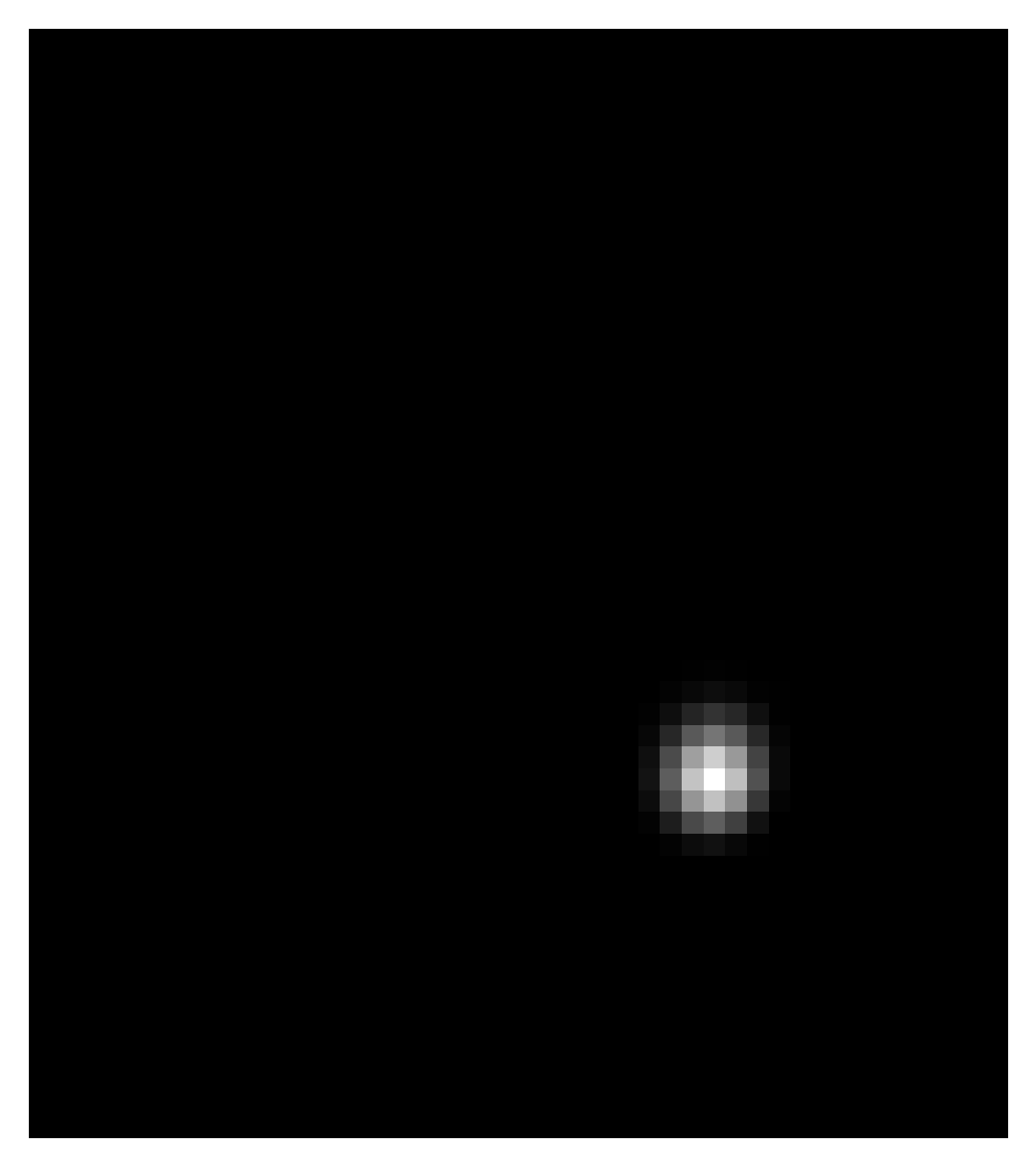} & \includegraphics[width=\size]{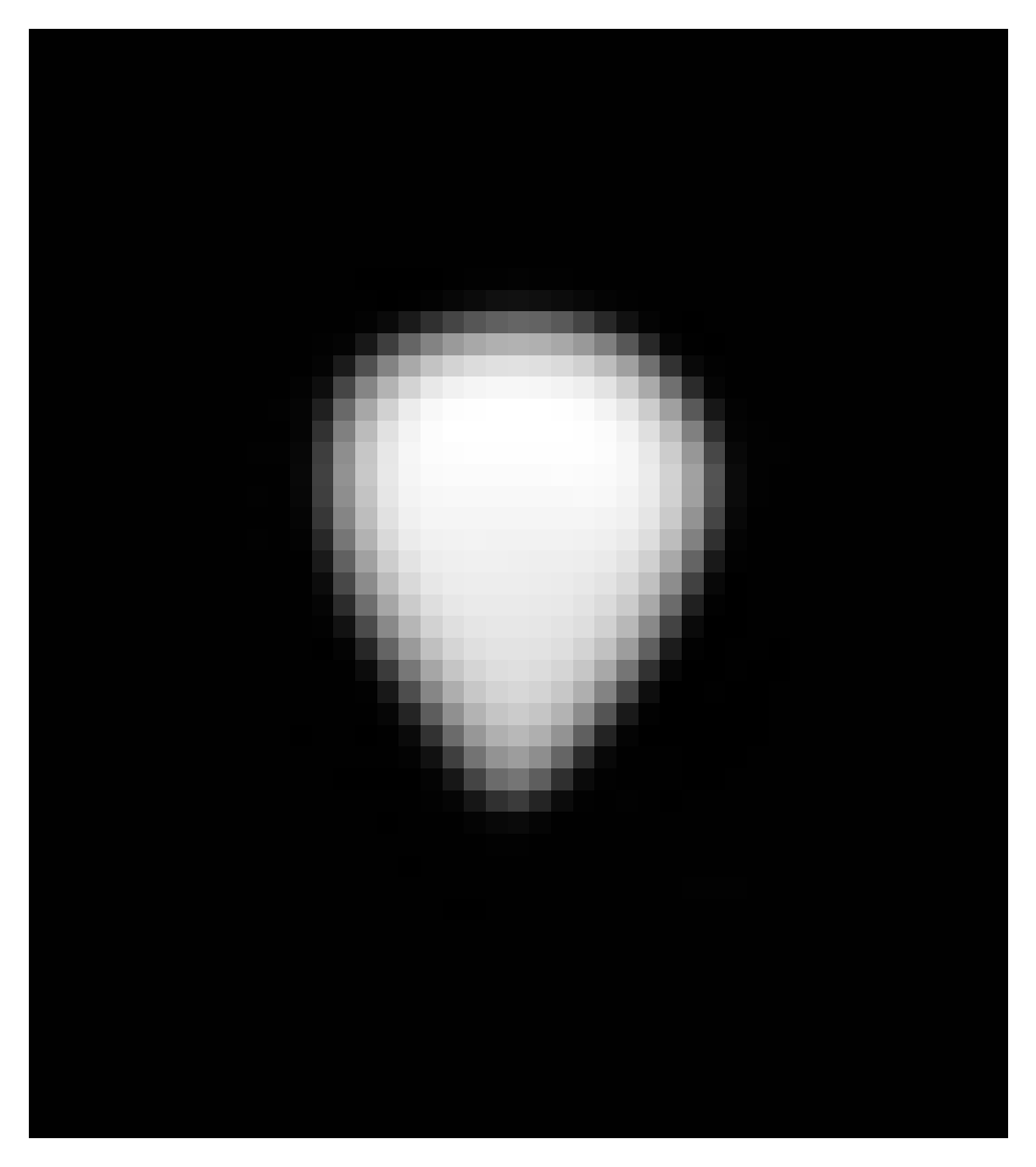} & \includegraphics[width=\size]{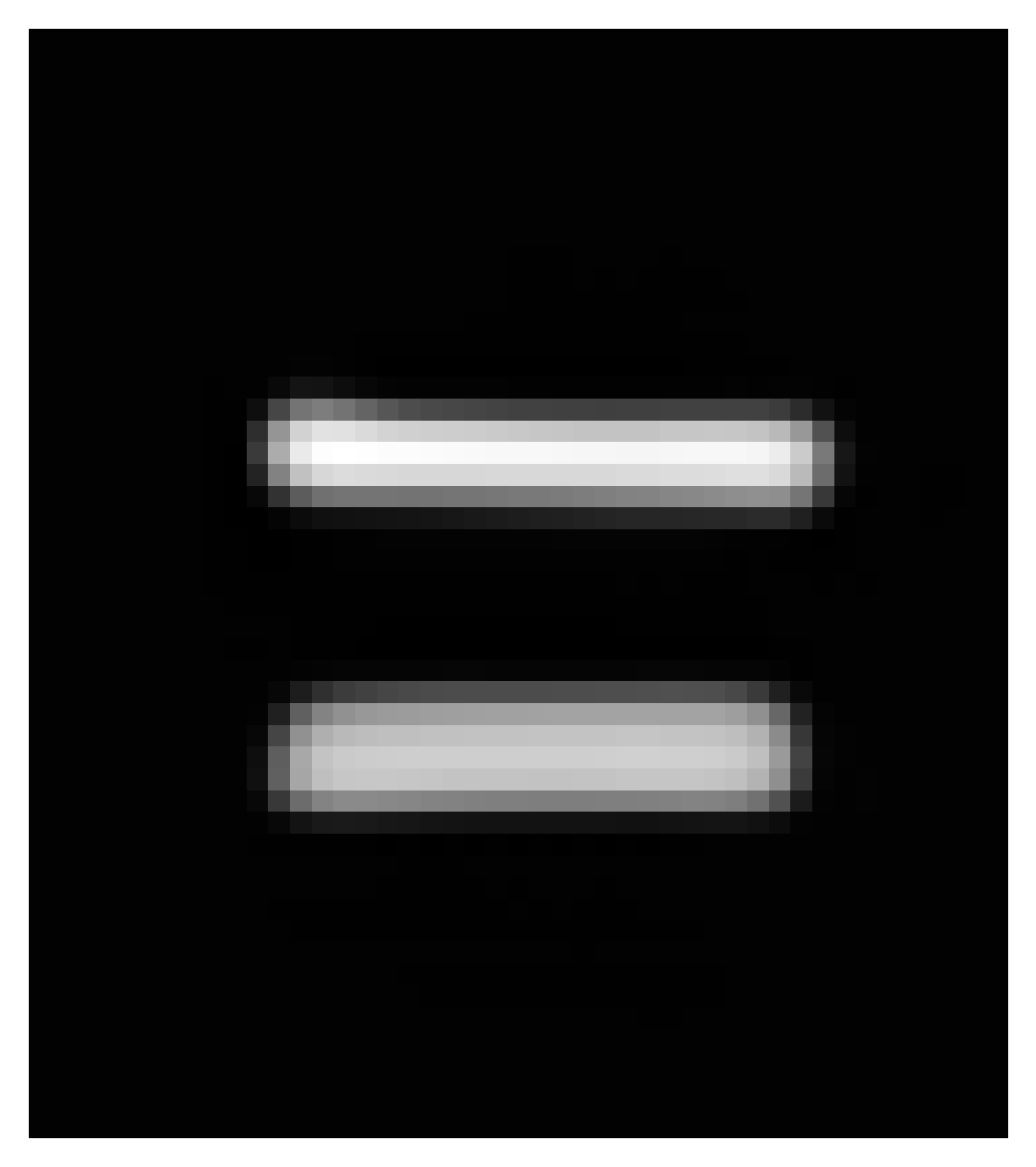} & \includegraphics[width=\size]{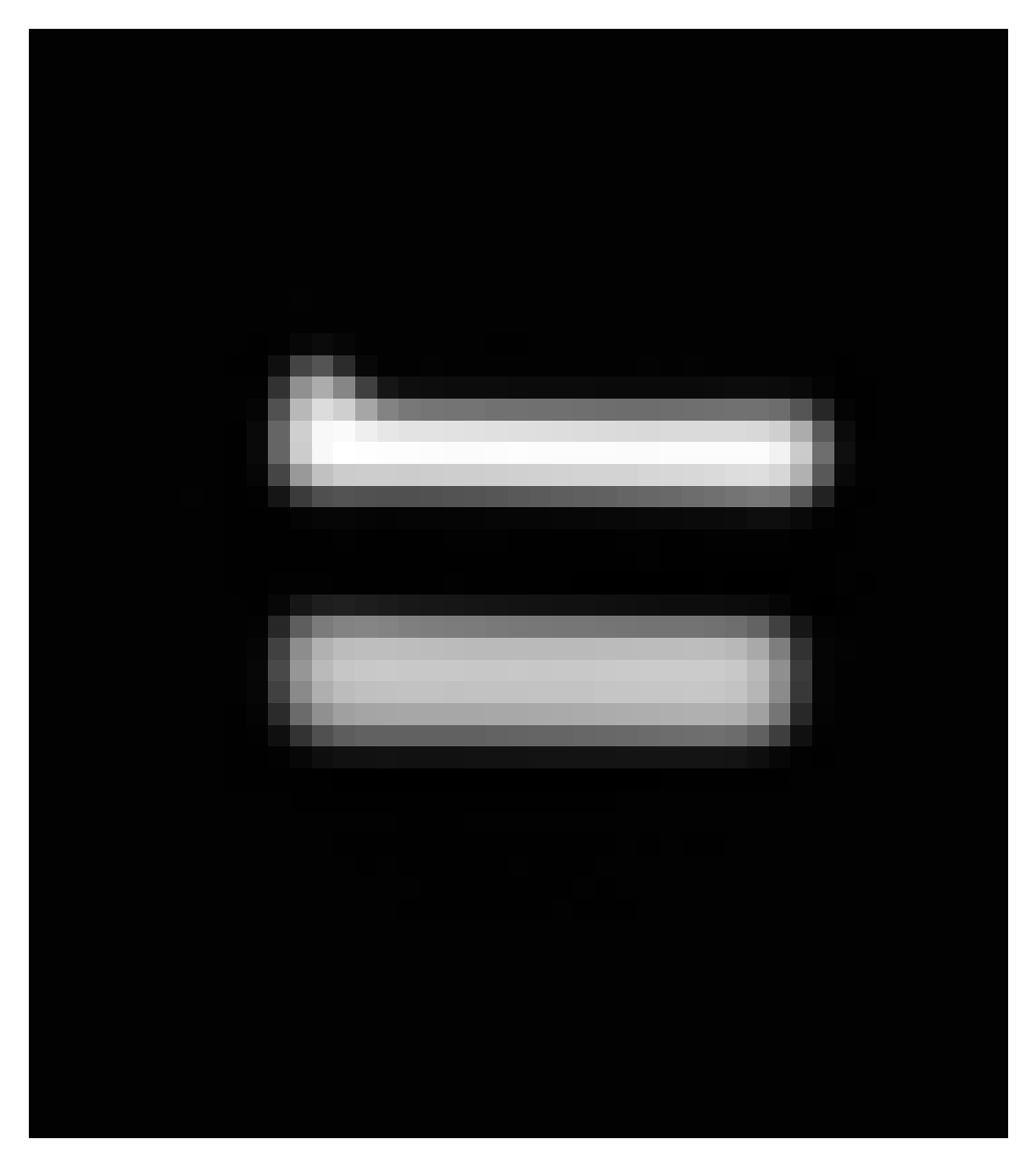} &
		\includegraphics[width=\size]{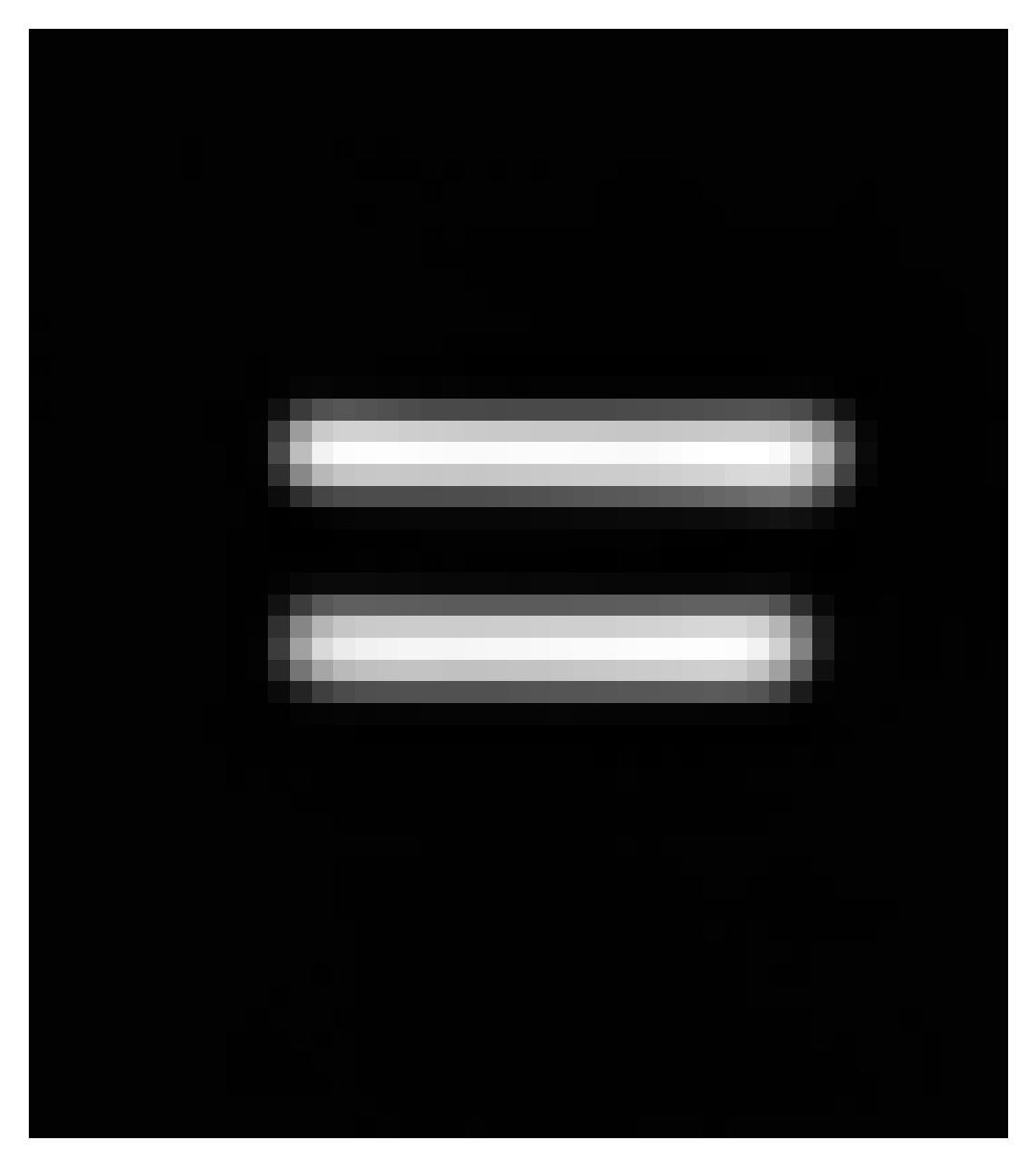} & \includegraphics[width=\size]{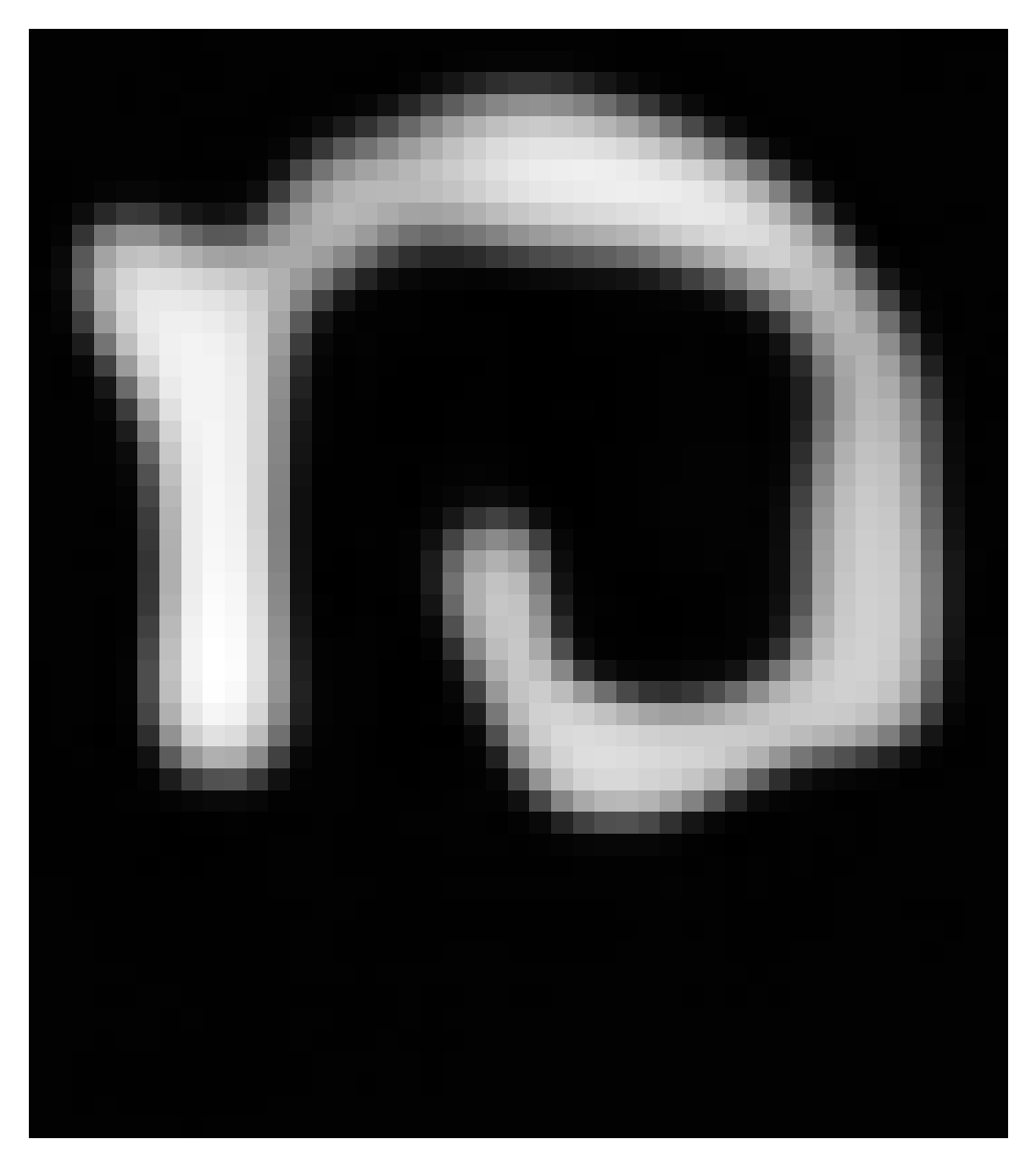}\\
		$\mathrm{Super}_5 (A_{17\times 15} )$ & \includegraphics[width=\size]{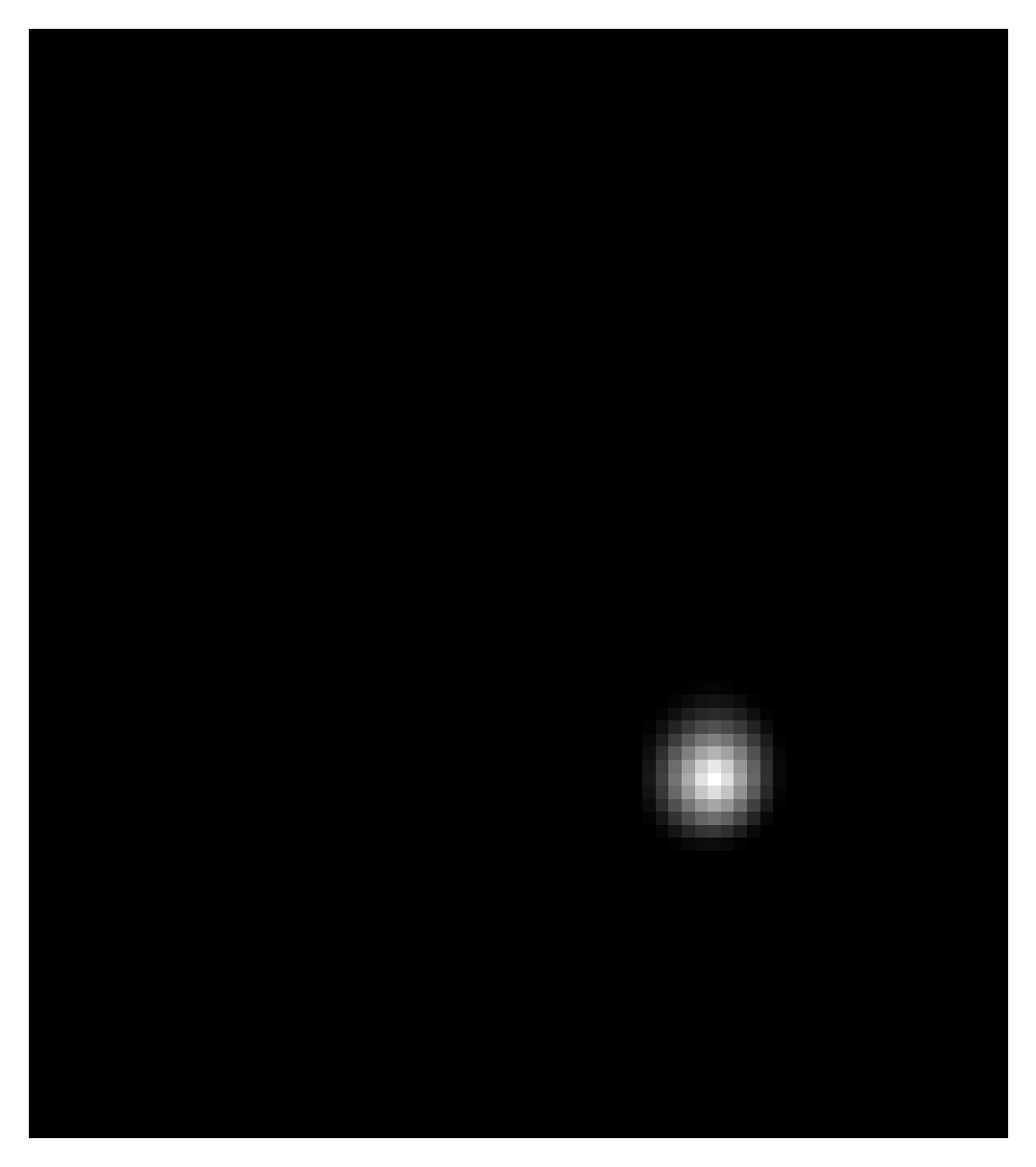} & \includegraphics[width=\size]{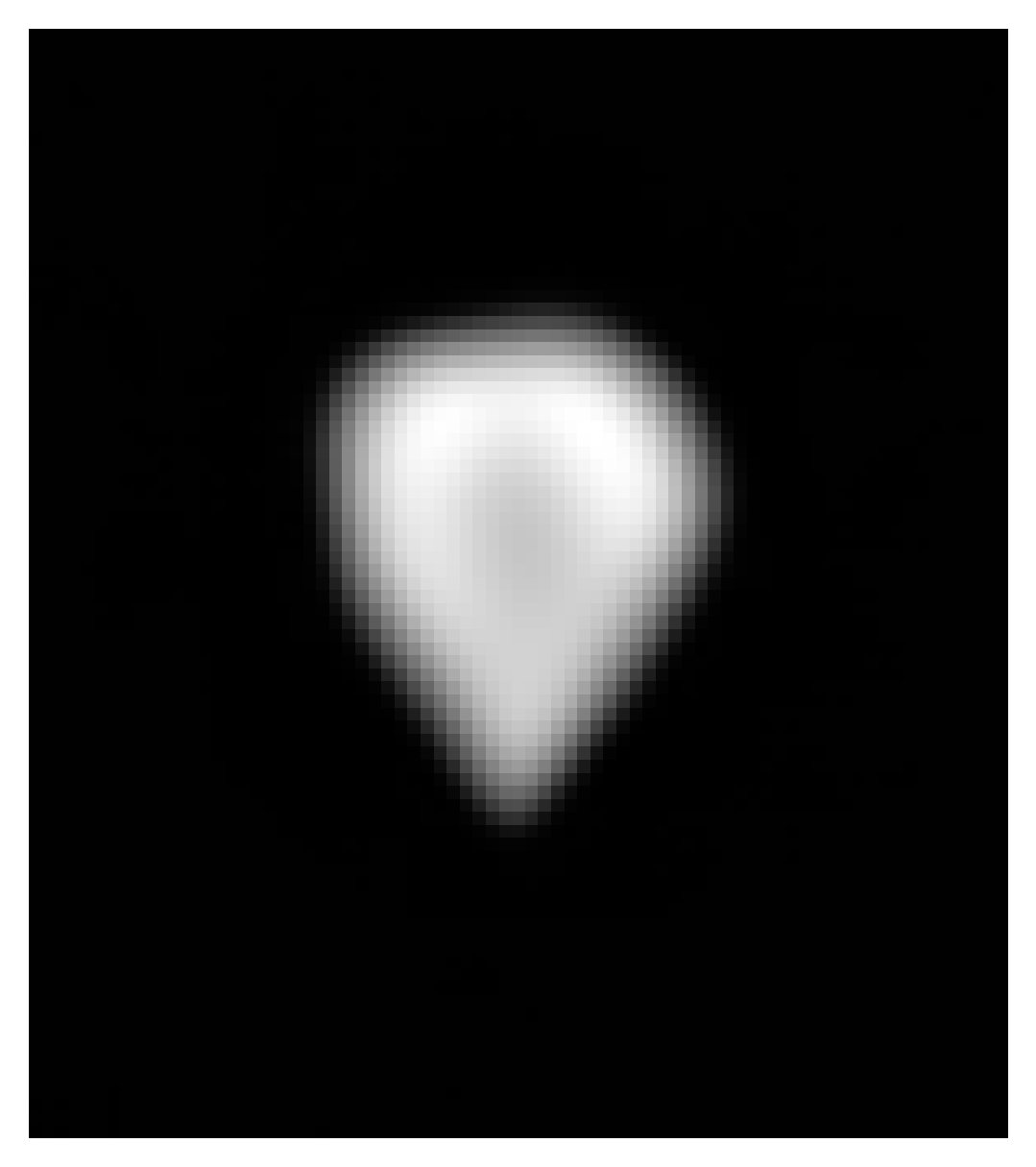} & \includegraphics[width=\size]{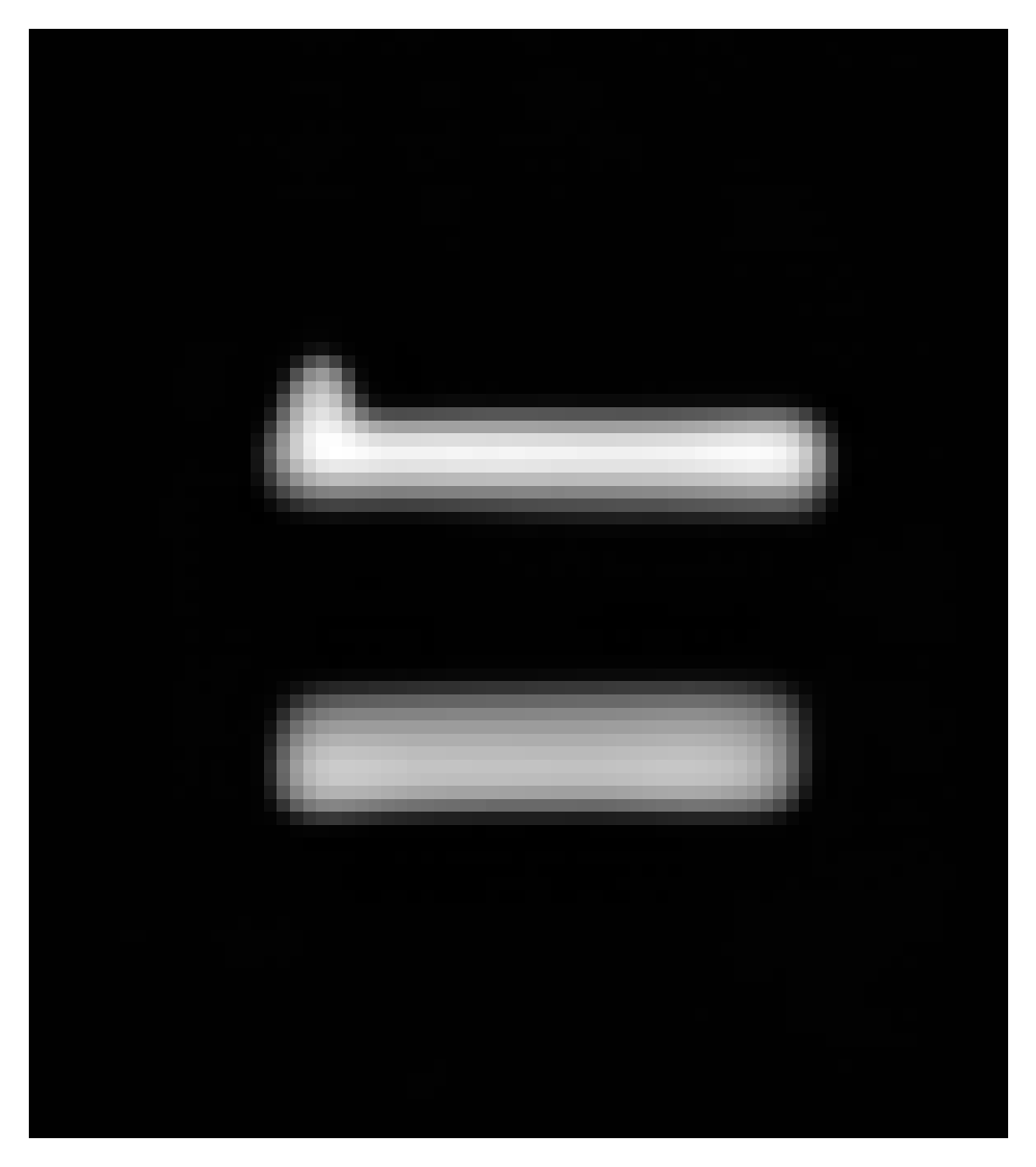} & \includegraphics[width=\size]{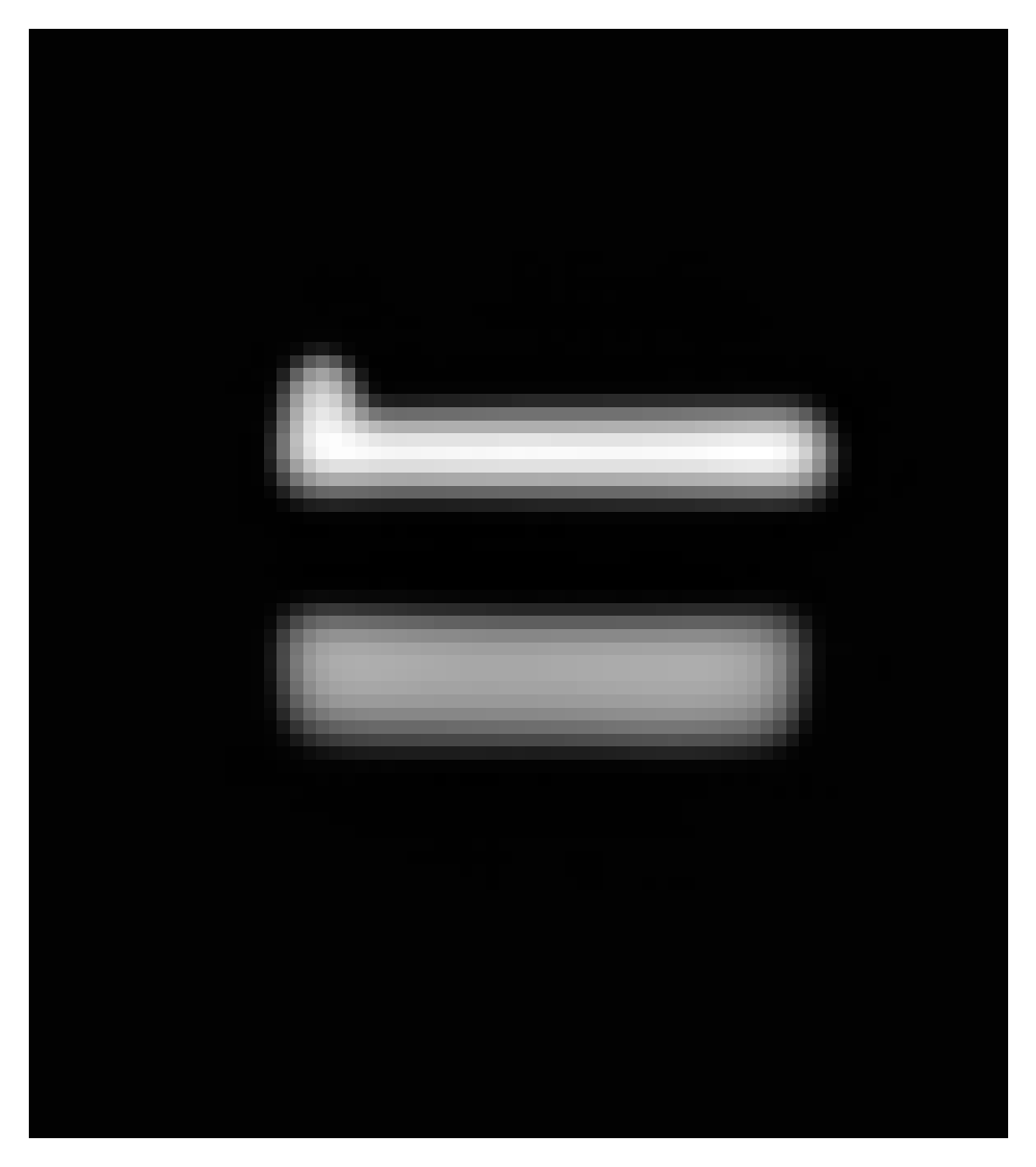} &
		\includegraphics[width=\size]{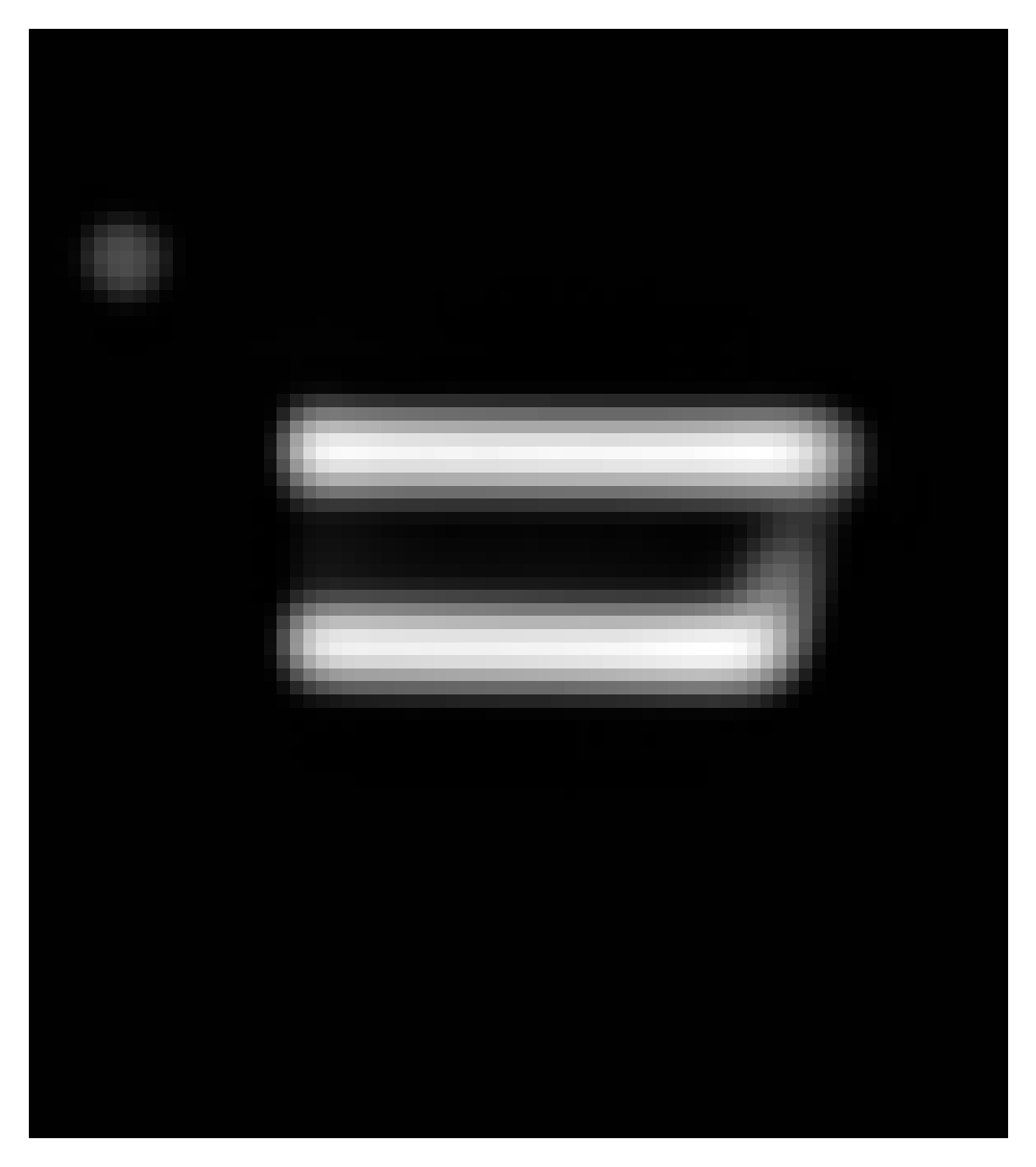} & \includegraphics[width=\size]{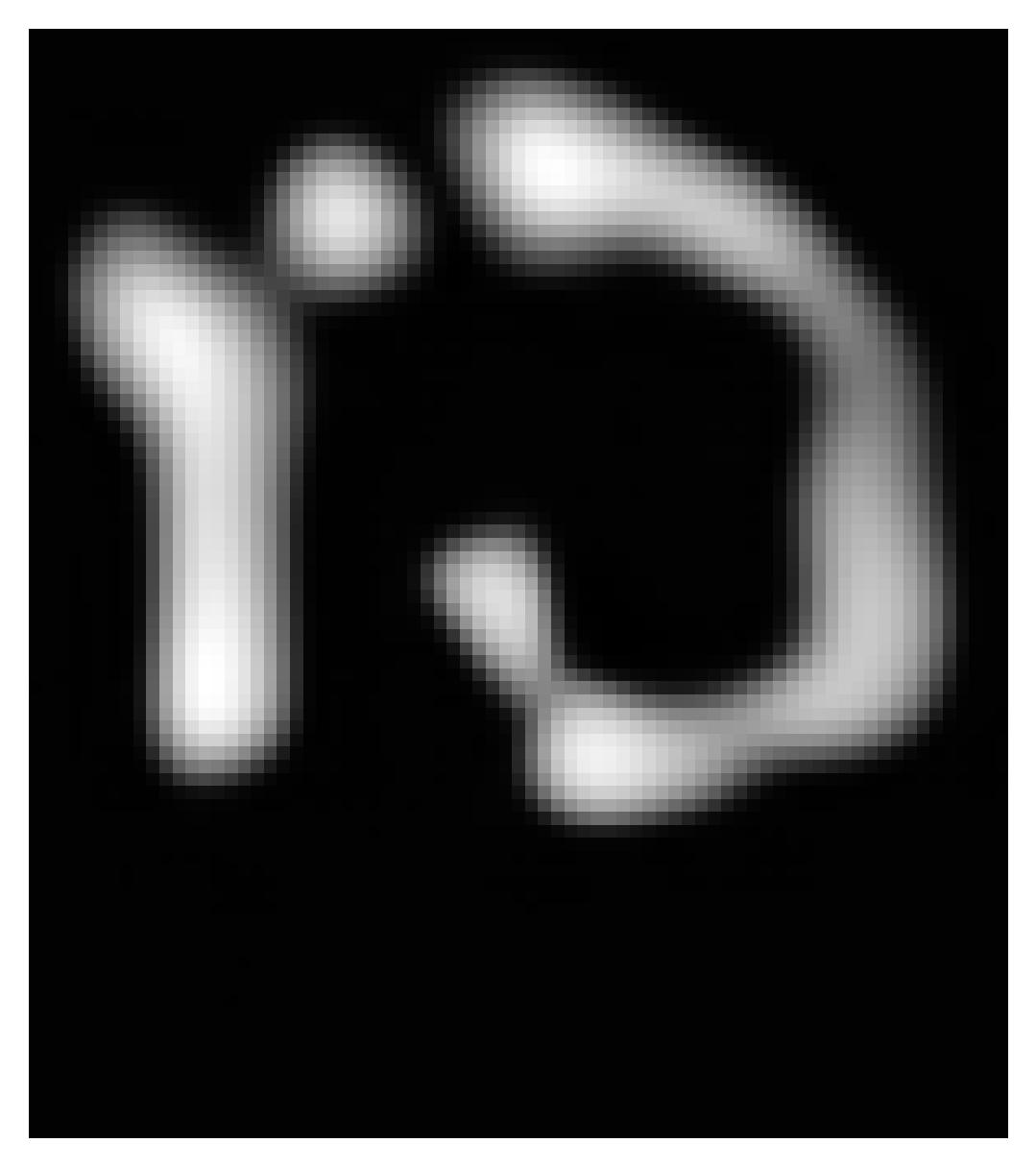}\\
		\bottomrule
	\end{tabular}
	\caption{Reconstructions on the EMWA dataset using the provided system matrix 
		calibrated on a $15\times 17$ grid using the ZS-PnP algorithm as well as \our\ with a scale 
		factor $s$ of 2,3 and 5.}
	\label{tab:emwa}
\end{table}

In the previous experiment we have shown reconstructions on the OpenMPI dataset, which contains 3 phantoms. 
In order to further show the super-resolution capabilities of \our\ across multiple instances of real data, 
we additionally perform reconstructions on the EMWA dataset. This dataset contains scans of 6 phantoms and 
a system matrix calibrated on a $15\times 17$ grid. The dataset does neither contain ground truths nor a 
higher-resolution system matrix, but pictures of the phantoms. Consequently, we only qualitatively 
evaluate the super-resolution capabilities of \our . 
The data has been processed by cutting frequencies below $80\ \si{\kilo\hertz}$, 
by performing SNR thresholding with a threshold of 1 and low-rank approximation with rSVD with target 
rank $K=\frac{15\cdot 17 -1}{2}$. Reconstructions have been performed with $\mu_0 = 10^9$ for all phantoms 
and resolution factors $s$. Concerning $s$, as done for the OpenMPI dataset, we perform reconstructions 
for $s =2,3,5$. The final reconstructions and the provided pictures with the phantoms are displayed in 
table~\ref{tab:emwa}. We observe that, compared with 
the non-super-resolved reconstruction on the $15\times 17$ grid, 
\our\ algorithm can help in reconstructing more sharply certain 
features of the phantoms (e.g. the tip of the ice-cream phantom and a clearer structure 
of the snail phantom). In this experiment we also observe the presence of reconstruction artifacts when 
the super-resolution factor $s\geq 3$. 

\section{Conclusions}\label{sec:conclusion}

In this paper, we have derived a plug-and-play-type method for super-resolved 
MPI reconstruction. More specifically, we have introduced \our , which leverages the denoising capabilities of a zero-shot denoiser: the \ddp. 
The incorporation of super-resolution was derived using a splitting scheme 
that minimizes a specifically defined cost function.
We have provided mathematical motivation for the proposed algorithm.
Further, we have discussed the choice of parameters, and we have provided 
a pseudocode of the algorithm.

We have shown the proposed method's potential by application to real and synthetic data. 
More precisely, we have first applied the derived method to the synthetic 
MPI-MNIST dataset, and derived quantitative and qualitative results. 
We observed that the dataset lacks finer level details. This observation led us to consider 
a self-designed synthetic dataset which is inspired by 
the MPI-MNIST dataset but provides higher-level details. 
On this dataset, we derived quantitative and qualitative results as well. 
In particular, we could observe the algorithm's capabilities concerning 
the reconstruction of higher-level details.
Finally, we have applied the algorithm to real data from the 
MPIData:~EquilibriumModelWithAnisotropy, and the 2D-OpenMPIData datasets.
In summary, our experiments on real and synthetic data, have shown the 
proposed method's potential for super-resolved MPI reconstruction.  

Topics of future research include the incorporation of more recent zero-shot denoisers (based on the transformer architecture). Additionally, examining the effect of other splitting and interpolation schemes, constitute a further research direction.

\section*{Acknowledgements}
This work was supported by the Hessian Ministry of Higher Education, Research, Science and the Arts within the Framework of the ``Programm zum Aufbau eines akademischen Mittelbaus an hessischen Hochschulen" and by the German Science Fonds DFG under grant INST 168/4-1.

\bibliographystyle{ieeetr}
\bibliography{literature}

\end{document}